\documentclass{article}
\usepackage{authblk} 
\usepackage[margin=1in]{geometry}

\usepackage[utf8]{inputenc} 
\usepackage[T1]{fontenc}    
\usepackage{lmodern} 
\usepackage{color}
\usepackage[dvipsnames]{xcolor}
\usepackage[colorlinks=true,
    linkcolor=RoyalBlue,
    citecolor=ForestGreen,
    urlcolor=Magenta]{hyperref}
\usepackage{url}            
\usepackage{booktabs}       
\usepackage{amsfonts}       
\usepackage{nicefrac}       
\usepackage{microtype}      
\usepackage{lipsum}		    
\usepackage{graphicx}
\usepackage{natbib}
\usepackage{doi}

\usepackage{enumitem}
\setlist[description]{font=\normalfont} 
\usepackage{subcaption}
\usepackage[usestackEOL]{stackengine}
\usepackage{changepage}
\usepackage{algorithm}
\usepackage{algpseudocode}
\usepackage{eqparbox}
\usepackage{multirow}
\usepackage{colortbl}
\usepackage{array}
\usepackage{csquotes}
\usepackage{kbordermatrix}

\algnewcommand{\LineComment}[1]{\Statex \hfill {\color{gray} $\triangleright$ #1}}

\usepackage{tabularx} 
\usepackage{booktabs} 
\usepackage{mathtools}

\DeclarePairedDelimiter\floor{\lfloor}{\rfloor}
\DeclarePairedDelimiterX\setc[2]{\{}{\}}{\,#1 \;\delimsize\vert\; #2\,}
\DeclarePairedDelimiter\abs{\lvert}{\rvert}%

\newtheorem{mydef}{Definition}

\newcommand{\figuresautorefname}{Figures~}
\newcommand{\algorithmautorefname}{Algorithm~}

\newcommand{\definitionautorefname}{Definition~}
\newcommand{\lineautorefname}{Line~}
\newcommand{\linesautorefname}{Lines~}

\renewcommand{\tableautorefname}{Tables~}
\renewcommand{\equationautorefname}{Equation~}
\renewcommand{\figureautorefname}{Figure~}
\renewcommand{\tableautorefname}{Table~}
\renewcommand{\sectionautorefname}{Section~}

\providecommand{\keywords}[1]{\textbf{Keywords:} #1}

\title{Topological Structure Description for Artcode Detection Using the Shape of Orientation Histogram\thanks{This work is an extension of an ACM MM'17 workshop paper \citep{liming2017recognizing}, which was completed in late 2017 and early 2018 during the first author's doctoral studies at the University of Nottingham.}}

\author[1]{Liming Xu}
\author[2]{Dave Towey\thanks{Corresponding author.}}
\author[3]{Andrew P.~French}
\author[3]{Steve Benford}

\affil[1]{Department of Engineering, University of Cambridge, Cambridge, United Kingdom\\}
\affil[2]{School of Computer Science, University of Nottingham Ningbo China, Ningbo, China}
\affil[3]{School of Computer Science, University of Nottingham, Nottingham, United Kingdom}

\date{} 

\begin{document}
\maketitle

\renewcommand\thefootnote{}\footnotetext{Contact: lx249@cam.ac.uk, dave.towey@nottingham.edu.cn, \{andrew.p.french, steve.benford@nottingha.ac.uk\}}

\begin{abstract}
    The increasing ubiquity of smartphones and resurgence of VR/AR techniques, it is expected that our everyday environment may soon be decorating with objects connecting with virtual elements.
    Alerting to the presence of these objects is therefore the first step for motivating follow-up further inspection and triggering digital material attached to the objects.
    This work studies a special kind of these objects --- Artcodes --- a human-meaningful and machine-readable decorative markers that camouflage themselves with freeform appearance by encoding information into their topology.
    We formulate this problem of recongising the presence of Artcodes as Artcode proposal detection, a distinct computer vision task that classifies topologically similar but geometrically and semantically different objects as a same class.  
    To deal with this problem, we propose a new feature descriptor, called the shape of orientation histogram, to describe the generic topological structure of an Artcode.
    We collect datasets and conduct comprehensive experiments to evaluate the performance of the Artcode detection proposer built upon this new feature vector. 
    Our experimental results show the feasibility of the proposed feature vector for representing topological structures and the effectiveness of the system for detecting Artcode proposals. 
    Although this work is an initial attempt to develop a feature-based system for detecting topological objects like Artcodes, it would open up new interaction opportunities and spark potential applications of topological object detection.
\end{abstract}


\section{Introduction}
\label{sec:introduction}
We are surrounded by beautiful, decorative patterns. 
Decorative patterns are ubiquitous within our everyday life: almost every object is embellished with carefully designed patterns that enhance their beauty, meaning, and value \citep{meese2013codes}.
The increasing ubiquity of smartphones and resurgence of VR/AR techniques, it is expecting that our everyday environment will containing objects augmented by virtual elements in the near future.  
Interacting with such augmented objects may become routine: watching product relevant contents by scanning packages, reading a guitar's life story by scanning its surface decorative patterns  \citep{benford2015carolan}, playing music through scanning a bespoke pattern, etc. 
These surface patterns are visual markers that serve as {\it entry points} for accessing additional digital material \citep{xu2022connecting}. 
These visual marker are not all inherently ``visible'' to people: some are simply visible due to their special such as QR codes \citep{wave2015information}, and some are ``invisible'' (or ``hidden''), masquerading as often encountered pictures, drawings or posters. 
The scan {\it affordance} of these ``invisible'' visual markers are not naturally established: the cues that a surface pattern is scannable are unknown to humans.  
Explicit instructions or additional marker discovery processes may need to bridge the ``affordance gap''. 
Explicit instructions, such as a reminder for scanning, may affect aesthetics in certain scenarios.
This paper aims to discover these kinds of invisible visual markers computer vision technologies, hinting at their existence to users. 
Specifically, this work deals with this ``hidden'' visual markers discovery (or invisible visual markers {\it proposal}) problem using a feature-based machine learning-driven approach, as exemplified by a special case of visual markers --- Artcodes.

\begin{figure}
    \centering
    \begin{subfigure}[t]{0.35\textwidth}
        \includegraphics[width=\textwidth]{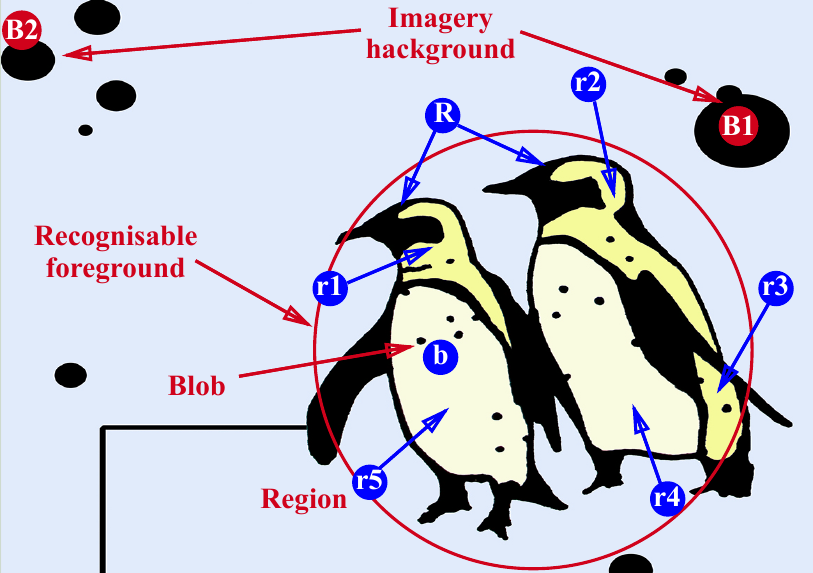}
        \caption{Artcode components} 
        \label{fig:artcodeIllustration-a}       
    \end{subfigure}
    \begin{subfigure}[t]{0.35\textwidth}
        \includegraphics[width=\textwidth]{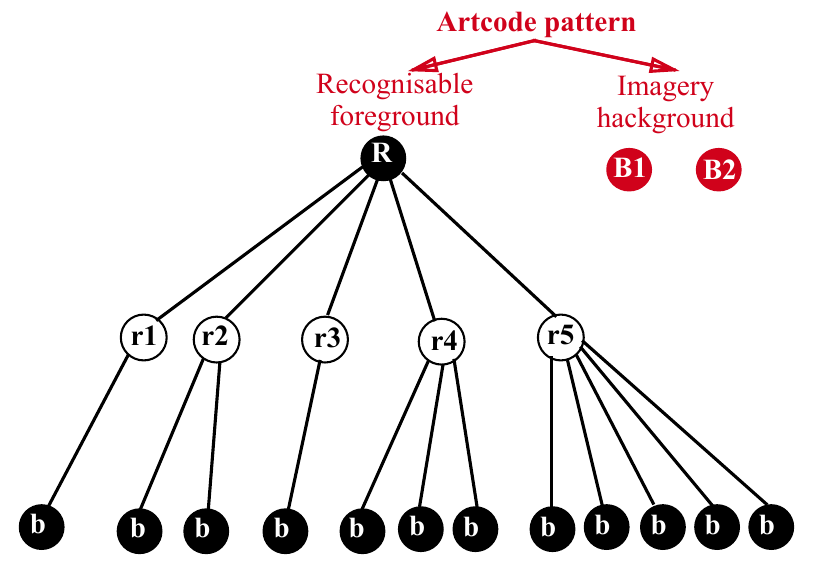}  
        \caption{Region adjacency tree of the recognisable foreground}
        \label{fig:artcodeIllustration-b}
    \end{subfigure}  
    \caption{
        Illustration of the components of an Artcode (code: ``1-1-2-3-5'') and the region adjacency tree of its recognisable foreground. 
        (b) presents the region adjacency trees of the penguins and the two examples (``B1'' and ``B2'') of background imagery.
        The trees of background imagery have only root nodes, and they are not d-touch codes, not to mention Artcodes.
    } 
    \label{fig:artcodeIllustration}
\end{figure}

Artcodes \citep{meese2013codes} are a visual markers that extend the implementation of d-touch \citep{costanza2009designable} markers with additional design rules, allowing designers to create both machine-readable and human meaningful patterns (see \figureautorefname\ref{fig:artcodeIllustration-a}).
Artcodes therefore adopt the space between the visibility of the QR codes and the secrecy of more ``invisible'' markers. 
They have been used in various scenarios: 
enhancing a dining experience \citep{meese2013codes}; 
augmenting an acoustic guitar with digital footprints \citep{benford2015augmenting, benford2016accountable}; 
attaching complex narratives to public illustrations \citep{thorn2016exploring}; 
designing a mobile garden guide using Artcode-based signage \citep{ng2016design}; and 
creating hybrid gifts \citep{koleva2020designing}, etc.

Artcodes are invisible visual markers, camouflaged by their well-designed patterns and imagery background.
Alerting to people of their presence is a key step to trigger their follow-up operations. 
Unlike the restrictive geometry of other commonly used markers, such as QR codes or ARTags \citep{Fiala2005}, Artcodes are designed by following topological rules, thereby avoiding the need for a fixed geometric shape. 
These generic topological structures, consisting of a set of connected hollow {\it regions} that contain several solid {\em blobs} (see \figureautorefname\ref{fig:artcodeIllustration-a})) \citep{costanza2009designable, meese2013codes}, distinguish Artcodes from other common images.
Consequently, it is difficult to detect them by finding a salient geometric structure, such as a thick, square border using conventional differential geometry-based feature descriptors.

We formulate the problem of recognising Artcode's presence as a proposal detection problem, called Artcode proposal detection, which will be further described in \sectionautorefname\ref{sec:detectionProposal}.
Artcode proposal detection addresses the question of whether it is likely that there is code in a particular scene, ideally even when imaging conditions mean that it can not be decoded as yet. 
This point is important, as it will allow the discovery of codes which are not yet interpretable by machine, and, with localisation, can then be used to guide the user into a possibly better suited location for decoding. 
There are three essential elements for a machine learning-powered Artcode proposer: a dataset, a feature vector, and a classifier, 
A detailed description on these three aspects is given later in this paper.

The main contributions of this paper are summarised below:
\begin{enumerate}[nosep]
    \item[1)] We introduce Artcode proposal detection problem and generalise a distinct computer vision task --- topological object detection. 
    \item[2)] We define a generic topological structure and propose to quantify the smoothness and symmetry of the orientation histogram for representing such a structure, resulting in a new feature descriptor --- the Shape of Orientation Histogram (SOH).
    \item[3)] We create two datasets: the True Artcode Dataset (TAD) and the Extended Artcode Dataset (EAD), which are public available for studying Artcode or even generic topological object classification problems.
    \item[4)] We conduct two experimental studies to evaluate the performance of the SOH feature vector on detecting Artcode proposals, illustrating its effectiveness and superiority over traditional geometric feature descriptors.
\end{enumerate}

The rest of this paper is organised as follows.
\sectionautorefname\ref{sec:artcodeDetection} presents the Artcode proposal detection problem and the definition of generic topological structure.
\sectionautorefname\ref{sec:relatedWork} briefly reviews the related work on Artcodes and conventional feature descriptors for computer vision tasks.
\sectionautorefname\ref{sec:featureDescriptor} describes the SOH feature vector and details its 
constituent variables and construction process.
\sectionautorefname\ref{sec:experiments} detailed describes the datasets and experimental studies for evaluating the proposed feature vector and systems.
\sectionautorefname\ref{sec:discussion} includes discussions on the limitations and implications of this work.
Finally, \sectionautorefname\ref{sec:conclusion} concludes this paper and describes future work.

\section{Preliminaries}
\label{sec:preliminaries}
\begin{figure}[t]
    \centering
    \begin{subfigure}[b]{0.35\textwidth}
        \includegraphics[width=\textwidth]{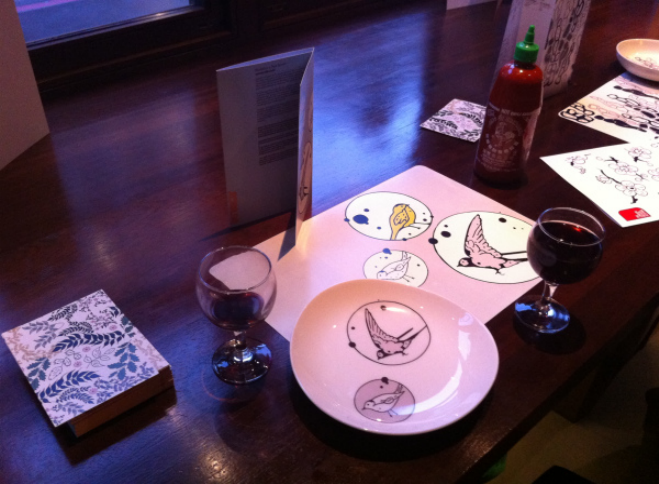}
        \caption{Artcodes in a dinner setting}
        \label{fig:artcodesDinner}
    \end{subfigure}
    \begin{subfigure}[b]{0.35\textwidth}
        \includegraphics[width=\textwidth]{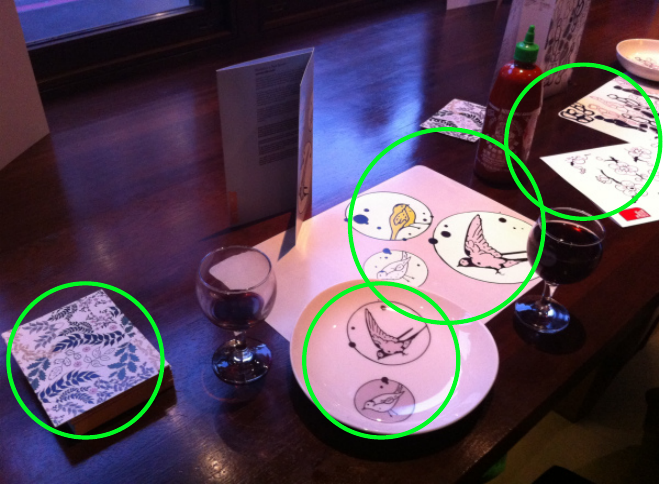}
        \caption{Proposed Artcode locations}
        \label{fig:detectedArtcodes}
    \end{subfigure}
    \caption{
        An example image of Artcodes in a dinning setting. 
        This image should be classified as Artcode class as it contains multiple Artcodes, highlighted by blue circles.
    }
    \label{fig:artcodeClassificationExample}
\end{figure}

\subsection{Artcode Detection Scenario}
\label{sec:artcodeDetection}
\figureautorefname\ref{fig:artcodeClassificationExample} illustrates an augmented dinning scenario, in which the menu, plate and placemat are decorated with Artcode patterns.
Artcode detection often involve two steps: first, determining whether this scene contains Artcodes or not; second, predicting the locations of Artcodes (as illustrated in \figureautorefname \ref{fig:detectedArtcodes}). 
The located Artcodes then need a further examinatopm, such as decoding the exact topological structure of the code (\figureautorefname\ref{fig:artcodeIllustration-b}). 
The decoded result is a string of the numbers of blobs in each region, separated by a hyphen or a comma sign. 
However, a clearer view is required for decoding the Artcodes, and thereby presence detection would be used to guide the user closer to the Artcodes.

\subsection{Generic Topological Structures}
\label{sec:gts}
Generating the Artcode candidates contains a basic operation --- Artcode classification. 
It is a binary classification problem, which classifies an input image as either containing Artcodes, or not --- labelled Artcode or non-Artcode class, respectively.
While the decoding is based on the {\it exact} topological structures of Artcodes, the proposed detection method only uses their {\it generic} topological structure. 
The generic topological structure is defined as follows: 
\begin{mydef}[Generic Topological Structure]
\label{def:genericTopologicalStructure}
The generic topological structure is composed of several connected regions (branches) and each region may or may not contain blobs (leaves). 
Regions and blobs are both connected components: a blob is solid, while a region contains a necessary hole. 
The connectedness described here is under the view of the human perspective, which would fills gaps in how objects are perceived (i.e., Law of Closure in Gestalt Psychology \citep{koffka2013principles}).
 \end{mydef}

This definition imposes no constraints on the number of regions (branches), the levels of nesting, and the minimum and maximum number of blobs (leaves) being contained in a region; all of which are specified in the d-touch or Artcode definition. 
This definition also ignores the three mechanisms (validation regions, checksum and redundancy) \citep{meese2013codes} used for improving decoding reliability in Artcode applications.

\subsection{Artcode Proposal Detection}
\label{sec:detectionProposal}
\begin{figure}
   \centering
   \begin{subfigure}[b]{0.17\textwidth}
        \includegraphics[width=\textwidth]{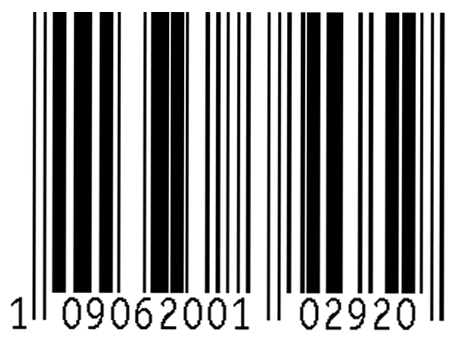}
        \caption{Barcodes}
        \label{fig:vm_barcode}
    \end{subfigure}
    \begin{subfigure}[b]{0.125\textwidth}
        \includegraphics[width=\textwidth]{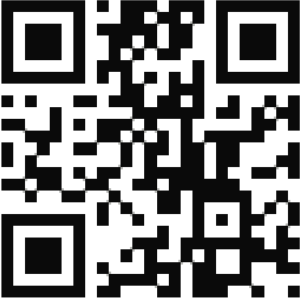}
        \caption{QR code}
        \label{fig:vm_qrcode}
    \end{subfigure} 
    \begin{subfigure}[b]{0.125\textwidth}
        \includegraphics[width=\textwidth]{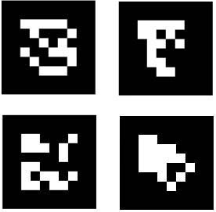}
        \caption{ARTags}
        \label{fig:vm_artag}
    \end{subfigure} 
    \begin{subfigure}[b]{0.125\textwidth}
        \includegraphics[width=\textwidth]{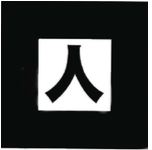}
        \caption{ARToolkit}
        \label{fig:vm_artoolkit}
    \end{subfigure}
    \begin{subfigure}[b]{0.135\textwidth}
        \includegraphics[width=\textwidth]{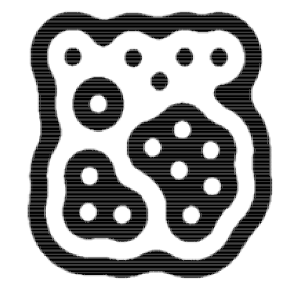}
        \caption{reacTIVision}
        \label{fig:vm_reactivision}
    \end{subfigure}    
    \begin{subfigure}[b]{0.125\textwidth}
        \includegraphics[width=\textwidth]{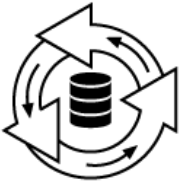}
        \caption{d-touch}
        \label{fig:vm_dtouch}
    \end{subfigure}
    \begin{subfigure}[b]{0.12\textwidth}
        \includegraphics[width=\textwidth]{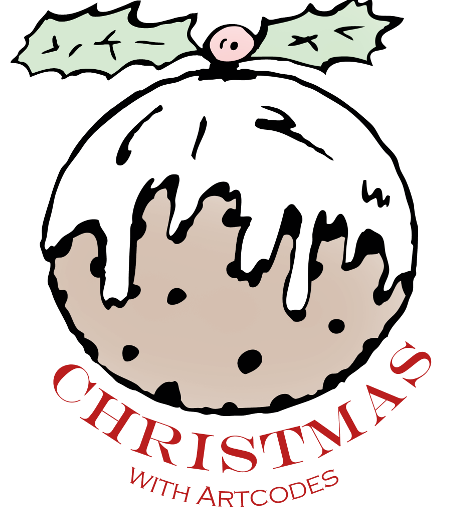}
        \caption{Artcode}
        \label{fig:vm_artcode}
    \end{subfigure}
    \caption{Visual marker examples.}
    \label{fig:vm}
\end{figure}

An image of which topological structure follows the above definition of a generic topological structure is not necessarily a guaranteed ``Artcode'', but instead can be considered to be ``Artcode-like''. 
Object detection detects a category of objects (cars, faces, humans, etc.) that have common features.
Proposal detection is to generate candidate window regions which {\it may} contain desired objects, with the goal of reducing the set of positions that need to be analysed further. 
Artcode instances vary in their exact topological structure which can only be known by analysing of their respective RATs \citep{Costanza2003}. 
Therefore, Artcode proposal detection detect candidate Artcode-like areas --- the areas that have generic Artcode topological structures and therefore likely contain Artcode instances. 
Artcode detection can actually be thought of as Artcode-{\it like} object detection, or Artcode {\em detection proposal} \citep{uijlings2013selective}.

Unlike object detection,imposing stringent performance metrics on detectors, an object proposal detection emphasises two critical properties: {\it high recall} and {\it efficiency} \citep{hosang2016makes, zitnick2014edge}. 
Specifically, object proposal detection generates bounding boxes that cover all potential instances as much as possible, even though this will increase false positives (i.e., reduce precision). 
In Artcodes detection scenarios, the generic topology definition can introduce instances that follow this definition but are not actual Artcodes. 
It may increase the number of false positives. 
Artcode propoal generation acts as an initial step to trigger subsequent precise recognition (localisation and/or decoding).
Therefore, recall is much more important than precision here.

\subsection{Detection by Decoding or Matching}
\label{sec:detectionByRecognition}
This section discusses the previous methods for visual marker or fiducial presence detection and why they are unsuitable for Artcode proposal detection. 
The previous methods detect them by looking up their predefined geometric shapes such as rectangles, ellipses or by matching the decoded topology (represented by such as region adjacency tree (RAT)) with the template topology. 
Upon initial inspection, Artcode detection seems similar to marker or fiducial recognition in a scene. 
The widely-used AR systems, such as ARTags and ARToolkit, initially require localise the markers based on their salient geometry --- thick, square borders --- as shown in the examples such as \figuresautorefname \ref{fig:vm_artag} and \ref{fig:vm_artoolkit}. 
The thick, square borders can be used for both detection and calibration. 
These detection systems are based on the geometric features (e.g., a thick, square border) that do not corresponds to the code interiors. 
These visual features has similar role to the generic topology in Artcode systems. 
However, generic topology is designed to be undetectable by human eye.

\begin{figure}
    \centering
    \includegraphics[width=0.75\textwidth]{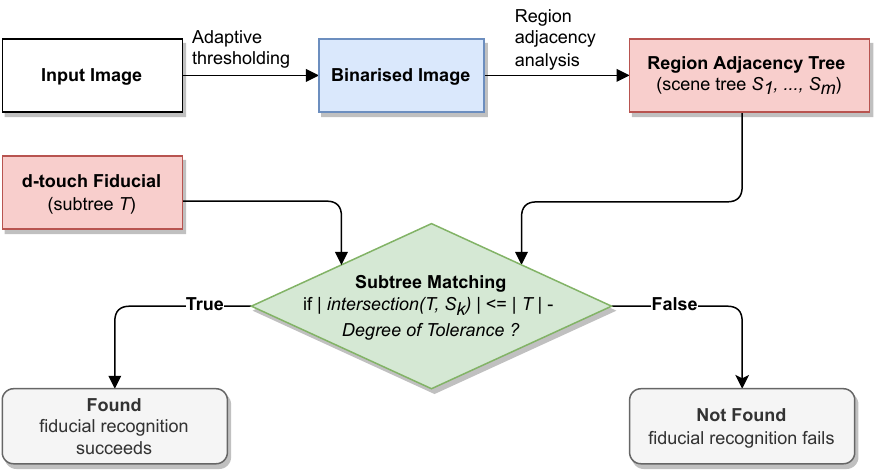}
    \caption{
        Process of the d-touch marker (fiducial) recognition. 
        Trees $T$ and $S_1, ..., S_m$ are represented as the sets of leaves, such as $T = \{ l_1, ..., l_n\}$; $| \cdot |$ denotes the cardinality of the set ($\cdot$).}
    \label{fig:dtouchRecognitionSteps}
\end{figure}

For less constrained systems such as d-touch, detecting d-touch markers (or fiducials) can be achieved by solving a simplified {\it subtree isomorphism problem} \citep{reyner1977analysis} --- matching the RAT of the predefined d-touch marker with the RAT of the scene image. 
\figureautorefname\ref{fig:dtouchRecognitionSteps} illustrates the d-touch detection process, in which the input image is first binarised using a Laplacian-based adaptive thresholding method \citep{Costanza2003} and the RAT of this two-level image is then constructed. 
Due to the constraints on the number of empty branches and the number of levels of nesting, the resulting RATs are {\it unrooted} and {\it non-oriented} trees.  
The subtree matching problem between fiducial RAT $T$ and scene RATs $S_1, ... , S_m$ can be solved in real time \citep{reyner1977analysis}. 
Because the allowed level of nesting for a d-touch code is three, the subtree matching can start from leaves, and then examine the nodes linked to leaves as branches. 
Likewise, the node linked to any branches can be labelled as a potential root. 
Therefore, the subtree matching can be done by comparing the {\it cardinality} ($| intersection(T, S_k) | $) of the intersection set of fiducial RAT $T$ and the scene RATs of $| T |$ degree with a predefined {\it degree of tolerance} --- a value used to control the relaxing condition to accept a node as the root of a fiducial instance. 
The degree of tolerance ranges from 0 to $| T |$, denoting the maximum allowed number of mismatched branches in a fiducial RAT $T$.

\begin{figure}[t]
    \centering
    \begin{subfigure}[t]{0.35\textwidth}
        \includegraphics[width=\textwidth]{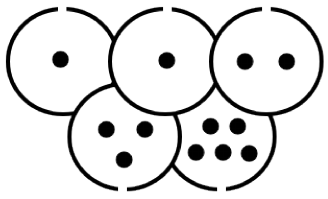}
        \caption{Synthetic Artcode-like image}
        \label{fig:real-im}
     \end{subfigure}
     \begin{subfigure}[t]{0.35\textwidth}
        \includegraphics[width=\textwidth]{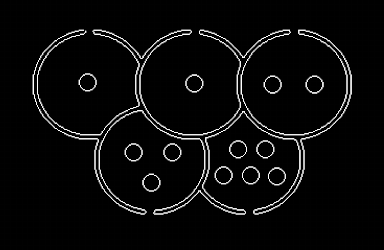}
        \caption{Corresponding edge image}
        \label{fig:real-em}
     \end{subfigure}
     \begin{subfigure}[t]{0.35\textwidth}
        \includegraphics[width=\textwidth]{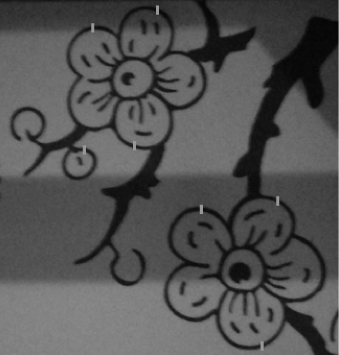}
        \caption{Real Artcode-like image}
        \label{fig:synthetic-im}
     \end{subfigure}
     \begin{subfigure}[t]{0.35\textwidth}
        \includegraphics[width=\textwidth]{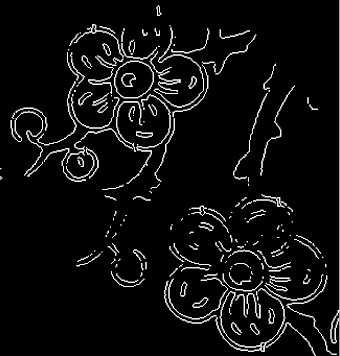}
        \caption{Corresponding edge image}
        \label{fig:synthetic-em}
     \end{subfigure}
     \caption{
        Example synthetic (top) and real (bottom) Artcode-like images that cannot be decoded due to gaps and bad lighting. 
        The small gaps on the region boundaries are intentionally added for showcasing the situations where the detection by decoding fails. 
        In contrast, these two examples can be ``detected'' by the proposed method in this work.
     }
     \label{fig:decodingFails}
\end{figure}

This detection-by-matching method requires us to examine fiducials' interior content --- namely the topological structure. 
Therefore, they depend upon which specific fiducial is provided, as it is different from the conventional fiducial localisation that is based on generic geometrical properties, such as the thick square border as mentioned earlier. 
To some extent, Artcodes can be understood as a special kind of d-touch markers, and the detection by matching or recognition paradigm would be effective for discovering Artcodes from the scene if: 
1) the codes (topological structures) of all searched Artcodes are known beforehand (which is not required for the proposed feature-based method); and 
2) the RAT of the scene can be correctly constructed, thereby the code details can be suitably restored.

However, these two conditions cannot be always satisfied Artcode scenarios. 
In Artcode scenarios, a scene may contain many Artcodes with different topological structures, which are unknown to a detector before decoding. 
A solution to this is to match all the available Artcode RATs with the scene RAT. 
However, this would be impossible because Artcodes are open to design by end users and all available Artcode RATs that should be used to match with the scene RAT are unknown beforehand. 
Moreover, computation would be expensive when the number of existing Artcodes is large because a great number of subtree matching tasks must be executed. 
This condition cannot be met in all Artcode scenarios. 
Instead of fiducials, which are inherently designed to be robustly located, Artcodes aim to be used as interactive decorative patterns --- which, by nature, means codes may exist as part of complex imagery. 
Therefore, the scene RAT may be difficult to construct correctly due to bad imaging conditions.
Two examples are presented in \figureautorefname \ref{fig:decodingFails} to illustrate decoding failures.
The gaps on the region boundaries in \figureautorefname\ref{fig:decodingFails} are intentionally added in for better explanation. 
These gaps may be caused by such complexities as {\em specular reflections}, {\em occlusions}, and {\em shadows}, which are very common in Artcode application contexts. 
The detection-by-decoding paradigm predicts whether a code is present or not by examining $| intersection(T, S_k) |$ --- a quantity influenced by the number of branches (regions) and the number of leaves (blobs) linked to them. 
The erroneous addition and reduction of regions and/or blobs can result in detection failure. 
Situations such as being too far away from the code, blurring, and occlusion, may also result in the inability to decode, and could lead to the failure of the detection-by-decoding method.

Artcode proposal detection propose possible Artcodes by examining their generic features rather than their specific fine structure through matching their RATs with the scene RAT. 
The Artcode class contains patterns that follow the generic topological definition of Artcodes, whereas the non-Artcode class simply consists of images that do not conform to these topological rules. 
As can be seen from the examples in \figuresautorefname\ref{fig:artcodesExamples} and \ref{fig:nonartcodesExamples}, there is no obvious difference in {\it geometric} shape or appearance between Artcodes and non-Artcodes, that is, the absence of thick, rectangular borders in more traditional examples. 
It is therefore impossible to categorise them based on the geometric features alone, as is easily observed. 
The geometric variation associated with Artcodes is very different to that of other well-known and well-structured markers, such as Barcodes \citep{maynard1993classifying}, QR codes,  ARTags, ARToolKit \citep{kato1999marker}, and reacTIVision \citep{Bencina2005}, which have fixed geometric shapes and strict encoding structures represented in that design.  
This flexible geometry of Artcodes make their detection difficult. 
We therefore propose a new feature vector (described in \sectionautorefname \ref{sec:featureDescriptor}) to help uncover (to detect) these hidden topological properties.

\section{Related Work}
\label{sec:relatedWork}
This work is a cross-disciplinary topic, spanning human-computer interaction and computer vision. 
In this section, we briefly review the work on traditional geometric feature descriptors, which is directly relevant to this work.
Over the past decades, there has been extensive research into feature detection and description, mainly including two classes: geometry-based features (Harris \citep{harris1988combined}, SIFT \citep{lowe2004distinctive}, and SURF \citep{bay2008speeded}), and binary features such as FAST \citep{rosten2006machine}, BRIEF \citep{calonder2010brief} and ORB \citep{rublee2011orb}). 
Geometric features commonly detect the keypoints at the most distinctive locations (local extrema after using Hessian or Harris operators, for instance) in the Laplacian of Gaussian (LoG) or approximate LoG scale space of the images, such as {\it corners}, {\it  blobs} and {\it T-junctions}. 
Next, they construct feature descriptors based on local information around the keypoints, such as histogram of oriented gradients (SIFT), Haar wavelet responses (SURF) or convolved orientation maps (DAISY) \citep{tola2008fast}).

However, such differential geometry-based feature detectors and descriptors impose a large computational burden, particularly for real-time systems or for low-power computational platforms such as mobile phones. 
Binary feature detectors were proposed to accelerate computational speed and to have similar descriptive performance. 
Binary feature detectors describe keypoints by examining the intensities of local patches. 
\citet{rosten2006machine} proposed the FAST detector, a decision tree-based detector which is able to effectively classify the image patches based on a relatively small number of pair-wise intensity comparisons. 
\citet{calonder2010brief} proposed BRIEF, directly comparing the intensities of pairs of pixels in an image patch after applying Gaussian smoothing to build the descriptor vector without requiring a training phase. 
\citet{rublee2011orb} proposed an Oriented FAST and Rotated BRIEF (ORB) descriptor, which is invariant to rotation and robust to noise, as well as being much faster than both the FAST and BRIEF features.

However, both geometric and binary features are based on local image patches; they detect and/or describe keypoints depending on local geometrical information around these keypoints. 
They are not designed to model such global information as topology, so they are not appropriate for modeling topological markers such as Artcodes. 
Another widely used approach for {\em generic visual categorisation} is the Bag of Visual Words (BoW)  (or Bag of Features) \citep{csurka2004visual}. 
The BoW method is based on vector quantisation of geometrical feature descriptors of image patches. 
A BoW feature vector corresponds to a histogram of the number of occurrences of particular image patterns in a given image. 
The particular image patterns are generated by clustering the geometric feature descriptors using methods such as K-means clustering \citep{Lloyd1982,Kanungo2002}. 
To achieve good performance, a large number of feature descriptors should be examined and clustered; therefore, the BoW method is computationally expensive.

Another useful feature for detecting objects is the histogram of oriented gradients (HoG) as proposed by \citet{dalal2005histograms}. 
HoG represents a given image by weighted vote over spatial and orientation cells and connects HoGs of every cell into a large HoG feature vector. 
Using histograms to describe features has been well studied. 
\citet{lowe2004distinctive} used HoGs in the neighbourhood of the keypoints to build the feature descriptor.
\citet{mikolajczyk2004human} used orientation-location histograms combined with thresholded gradient magnitude as feature descriptors to represent different parts of an object. 
However, \citet{dalal2005histograms} used the {\it dense} HoG, allowing overlapping cells. 
Combined with appropriate normalisation methods, dense HoG significantly outperforms existing features such as Haar wavelets \citep{papageorgiou2000trainable} for humans detection. 
These dense HoG descriptors are reminiscent of orientation histograms and HoG in SIFT; however, for human detection, fine orientation sampling turns out to be better than coarse spatial sampling as HoG used in SIFT. 
Therefore, they were computed on a dense grid of uniformly spaced cells and used overlapping local contrast normalisations to improve performance.
These features can effectively represent semantic objects, such as humans, faces, and cars, and are not appropriate for describing topological objects such as Artcodes. 
Additionally, the aforementioned approaches such as BoW and HoG are used to describe objects with high representation power and thereby are computationally expensive. 
Therefore, they are more appropriate for detection rather than for detection proposal, which demands higher recall and efficiency.

In this paper, we propose a new topological feature descriptor, referred to as SOH, for representing the Artcode image's generic topology.
A SOH is built from the commonly used orientation histogram \citep{freeman1995orientation}. 
However, rather than directly using the orientation histogram, a SOH is computed upon the shape of orientation histogram of a given image. 
The shape is described by a feature vector, consisting of the variables for measuring the symmetry and smoothness of the orientation histogram and some other auxiliary variables. 
An Artcode proposal generation system that uses SOH features and random forests or SVM classification methods, called {\sc ArtcodePresence}, is also developed, being evaluated to discover the likely locations of Artcodes in an image.

Orientation histograms were first used by \citet{freeman1995orientation} for hand gesture recognition, where orientation histograms were computed using steerable filters \citep{freeman1991design} and the orientations with magnitudes below a threshold were suppressed. 
Rather than based on pixel intensities, which are sensitive to lighting changes, orientations are essentially calculated based on the differences of pixel intensities over various directions (gradients). 
Thus, as described in \citet{freeman1995orientation}, these orientation histograms have a certain level of robustness to lighting changes and invariance to transformations such as rotations.

\begin{figure}[t]
    \centering
    \includegraphics[width=\textwidth]{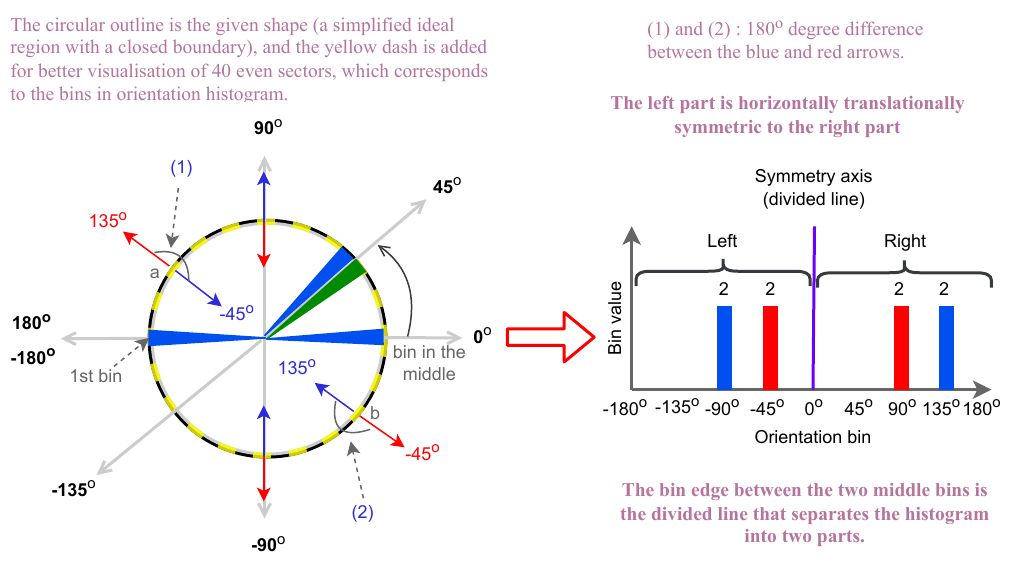}
    \caption{
        An illustration of the tuition that uses orientation histogram's symmetry and smoothness for representing generic topology. 
        The circular outline is the given sample shape of a {\it simplified ideal region} with a thick boundary, and the yellow dashed lines are added for better visualisation of fourty even sectors. 
        The blue and red arrows are the orientations of the gradients at points $a$ and $b$, which are a pair of arrows (orientations) at the corresponding points on both sides of the edge. 
        Therefore, the pairs of all points distributed over the circle may generate a (horizontally translational) symmetrical and smooth orientation histogram.
    }
    \label{fig:arrowIllustration}
\end{figure}

\section{The SOH Feature Descriptor}
\label{sec:featureDescriptor}
The SOH will be used for {\it proposing} the presence of Artcode-like objects that have a generic topological structure. 
This section describes the motivation for the SOH and the process to construct a SOH feature vector. 
Orientation histograms, or oriented gradients, have been extensively studied for the description of local information in various applications. 
Previous work \citep{freeman1995orientation, papageorgiou2000trainable, belongie2001matching, mohan2001example, lowe2004distinctive, mikolajczyk2004human, dalal2005histograms} based on gradient orientation only concerns the local geometric representation property of the orientation histograms, but pays little attention to the shape of the histogram itself, which is a potentially useful property for describing the topological aspects of a given image. 
Any geometrical shapes are possible for a valid Artcode if they follow the predefined topological rules (\sectionautorefname \ref{sec:artcodeDetection}). 
Therefore, as discussed, traditional geometrical and binary feature descriptors are inappropriate for representing this topological class of objects. 
Instead, a feature vector that can capture the Artcode's general topological features (see \definitionautorefname \ref{def:genericTopologicalStructure}) is desirable.

\subsection{The Shape of Orientation Histogram}
\label{subsec:shapeOrientationHistogram}
\begin{figure}[t]
    \centering
    \begin{subfigure}[b]{0.65\textwidth}
        \includegraphics[width=\textwidth]{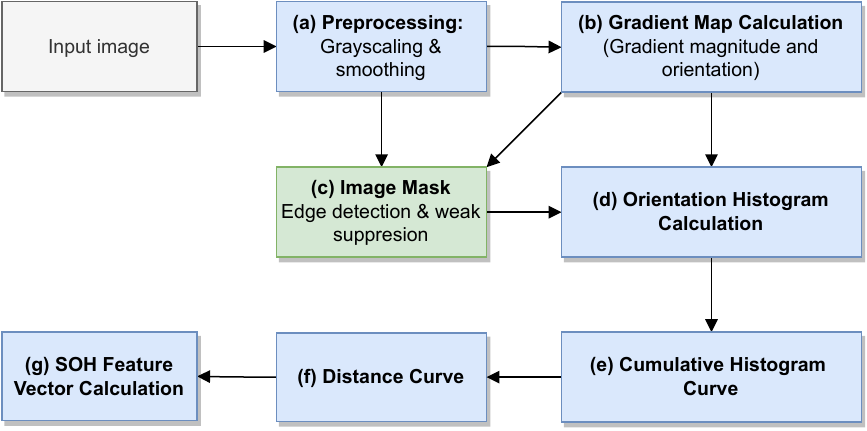}
    \end{subfigure} 
    \caption{
        Illustration of the process to construct SOH feature vector, with labels from (a) to (g) representing key steps.
    }
    \label{fig:SOHFeatureVectorConstruct}
\end{figure}
Unlike rigidly-structured visual markers, Artcodes are defined by rules which may result in almost any geometric shapes. 
Colour is of importance to the overall beauty of the code, but it is less critical for detection as long as it is not too pale. 
Artcode proposal detection must depend on their generic topology --- a closed boundary containing  closed several inner regions dominating the recognisable part of the Artcodes. 
Any images that mainly contain this topology should be considered as potential Artcodes (or Artcode-like). 
The \textit{intuition} that inspires me to study the shape of the orientation histogram is described as follows. 
Considering the circle outline that has both inside and outside circle boundaries, the directions (red and blue arrows) of the gradient at the two points across the boundaries are collinear with the {\em normal vector} at those points. 
However, it must noted that this is an ideal case, and some deviation from this will still follow this property to some extent. 
The orientations of the gradients at the two points are two pairs of opposite directional vectors, with an orientation difference of 180$^\circ$, as shown in \figureautorefname \ref{fig:arrowIllustration}. 
This relation is based on an assumption that 
1) the intensity of the boundary of a closed shape is uniform; and 
2) the curvature over the boundary is continuously changing. 
The orientations of the gradients at the edge of the shape are likely to be distributed over the range from $-180^\circ$ to $180^\circ$: the more circular the shape is, the more uniformly distributed the orientation of the gradients. 
An orientation histogram resulting from the gradient orientations on these edges has uniformly distributed bins, i.e. the general shape of this orientation histogram is not sharply changed.  

Meantime, the left part ($-180^\circ$ to $0^\circ$) of the orientation histogram is {\it horizontally translational symmetric} to the right part ($0^\circ$ to $180^\circ$). 
The bin edge between the bin of $[x^\circ, 0^\circ]$ and the bin of $[0^\circ, x^\circ]$, i.e., the line perpendicular to the $x-axis$ at the $0^\circ$, is the {\it symmetry axis} --- the divided line separates the orientation histogram into the left and right parts, in which one part is moving horizontally from the other part.

The two properties that are used to describe the shapes of these histograms are referred to as {\it smoothness} and {\em symmetry}, defined in detail as follows: 
\begin{mydef}[Symmetry]
\label{def:symemtry}
Symmetry is the similarity between the left ($-180^\circ$ to $0^\circ$) and the right ($0^\circ$ to $180^\circ$) part of the orientation histogram ranges from $-180^\circ$ to $180^\circ$, which is actually a translational symmetry.
\end{mydef}
\begin{mydef}[Smoothness]
\label{def:smoothness}
Smoothness is the overall rate of variation between the adjacent bins of the orientation histogram. 
\end{mydef}

Note that this symmetry is not {\it reflective} symmetry, a.k.a. mirror symmetry, which occurs when a line is drawn to divide a shape in halves so that each half is a reflection of the other, but is instead {\it horizontally translational} symmetry, where one half results from the other half  undergoing horizontal translation.

This assumption is based on an ideal case and, Artcodes can satisfy this assumption to some extent. Artcodes are an extension of d-touch as they add aesthetic design considerations, which  are usually beautiful interactive decorative patterns. 
Designers are likely to use smooth lines and shapes rather than triangles and rectangles to create the primary structure of the Artcode, as reported by \citet{meese2013codes}: 
\begin{displayquote}
They felt that the rectangular border had a very profound ``emotional'' effect in restricting the design from growing naturally in all directions to the point where it was.
\end{displayquote}
Meanwhile, a set of design guidelines introduced to advise designers when creating both aesthetic and recognisable Artcodes, such as {\em redundancy} and {\it smooth borders}, contribute to satisfying the assumption. 
We conducted so-called shape studies to study whether the two proposed properties can empirically capture the topological information.
Readers can refer to the supplementary material Appendix \ref{sec:shapeStudies} for more details on shape studies.

\subsection{Feature Vector Construction}
\label{subsec:featureVectorConstruction} 
\begin{algorithm}[!t]
\caption{SOH feature vector construction}
\label{algo:constructSOHFeat}
\hspace*{\algorithmicindent} Input: $im$ \Comment{input image} \\
\hspace*{\algorithmicindent} Output: $SOHFeat$ \Comment{SOH feature vector}
\begin{algorithmic}[1]
    \Procedure{constructSOHFeat}{$im$}
        \State $gray \gets \textproc{rgb2gray}(im); $ \label{line:grayscaling} \LineComment{convert $im$ from RGB to grayscale.}
        \State $gray \gets \textproc{imSmooth}(gray); $ \label{line:smoothing} \LineComment{smooth the grayscale image $gray$ using filters such as Gaussian and Bilateral filters if needed.}
        \State $ori, mag \gets \textproc{imGradient}(gray); $ \label{line:gradient} \LineComment{get the orientations and magnitudes of all pixels in $gray$}
        \State $em \gets \textproc{edgeDetector}(gray); $ \label{line:edgeDetection} \LineComment{return the edges using Sobel edge detector.}
        \State $strongMag \gets \textproc{weakSuppression}(mag); $ \label{line:weakSuppression} \LineComment{remove weak magnitudes, where the orientations of gradients with magnitude below a threshold are suppressed.}
        \State $newOri \gets \textproc{imMask}(ori, em, strongMag); $ \label{line:imageMask} \LineComment{remove the orientations that are not on the edge and whose magnitudes are weak.}
        \State $oriHist \gets \textproc{calcOrientationHist}(newOri, nBins); $ \label{line:calcOrientationHist} \LineComment{calculate the orientation histogram by binning the orientations into $nBins$ bins from $-180^\circ$ to $180^\circ$.}
        \State $cumHist \gets \textproc{calcCumulativeHist}(oriHist); $ \label{line:calcCumulativeHist} \LineComment{calculate the cumulative histogram of the provided histogram $oriHist$.}
        \State $distCurve \gets \textproc{calcDistanceCurve}(cumHist); $ \label{line:calcDistanceCurve} \LineComment{calculate the perpendicular distance from the cumulative histogram curve to the fitted line.}
        \State $SOHFeat \gets \textproc{calcSOHFeat}(gray, oriHist, cumHist, distCurve); $ \label{line:calcSOHFeat} \LineComment{calculate SOH feature vector based on $oriHist$, $cumHist$, $distCurve$, and $gray$}
        \State \textbf{return} $SOHFeat;$ 
    \EndProcedure
\end{algorithmic}
\end{algorithm} 

We in this section present the steps to build the SOH feature vector, as illustrated in \figureautorefname \ref{fig:SOHFeatureVectorConstruct} and detailed in \algorithmautorefname \ref{algo:constructSOHFeat}.
Specifically, we describe the construction of its two key components (orientation histogram and cumulative orientation histogram). 
We further present the quantification of symmetry and smoothness level of the orientation histogram and the variables for constructing the SOH feature vector.

\subsubsection{Orientation Histogram}
\label{subsubsec:orientationHistogram}
As shown in \figureautorefname \ref{fig:SOHFeatureVectorConstruct}, the input image is first converted RGB into grayscale using the formula \citep{bt1990studio}: $0.2989 * R + 0.5878 * G + 0.1140 * B$ (where $R$, $G$, $B$ are the values in the red, green, and blue channels, respectively), and then smoothed using a small Gaussian kernel to remove the colour information and noise (Step (a) in \figureautorefname \ref{fig:SOHFeatureVectorConstruct} and \linesautorefname \ref{line:grayscaling} and \ref{line:smoothing} in \algorithmautorefname \ref{algo:constructSOHFeat}). 
The gradient map at each pixel of the grayscale image was then calculated (Step (b) and \lineautorefname \ref{line:gradient}). 
The output of this step is a {\it magnitude} and an {\it orientation} map with the same dimensions as the input image.  
The next step (Step (c)) is to filter out pixels that are unrelated to the generic topology. 
The primary structure of an Artcode is the branches (regions) and leaves (blobs) in the RAT. 
The contours of regions and blobs are critical to topological analysis; therefore, only the orientations of the gradients on the edges of these objects are used to compute the orientation histogram (\lineautorefname\ref{line:edgeDetection}). 
These edges are also relevant to Artcode decoding.
Correct binarisation of these edge areas might be more important than other areas of the image, as discussed in \sectionautorefname\ref{sec:detectionByRecognition}.

The orientations with weak magnitude should be suppressed: the orientations with magnitudes below a predefined threshold are removed (\lineautorefname \ref{line:weakSuppression}).
Meanwhile, the orientations of the pixels that are not on the edges of boundaries were also removed (\lineautorefname\ref{line:edgeDetection}). 
Edges can be found by edge detection approaches \citep{Canny1986, gonzalez2012digital}, example edge maps of both synthetic and real images are shown in \figuresautorefname\ref{fig:shapeStudies} and \ref{fig:realImageStudies}) of Appendix\ref{app:supplementaryMaterial}. 
Orientation histogram is then computed from this filtered orientation map (Step (d) in \figureautorefname\ref{fig:SOHFeatureVectorConstruct} and \linesautorefname \ref{line:imageMask} and \ref{line:calcOrientationHist} in \algorithmautorefname \ref{algo:constructSOHFeat}).

We next normalise the resultant orientation histogram: its bin values are {\em normalised} using the following equation:
\begin{equation}
\hat{v}_i = \frac{v_i}{N}, \quad i = 1, \dots, nBins
\end{equation}
where $v_i$ is the number of elements in the $i$th bin, and $\hat{v}_i$ is its normalised value, $N$ is the number of elements in the input data and $nBins$ is the number of bins in this orientation histogram (Step (d)). 
The left and right halves of orientation histogram are respectively denoted as: 
\begin{align}
\mathcal{H_L} = \Bigl\{ \hat{v}_i \mid i = 1, \dots, n \Bigr\} \\
\mathcal{H_R} = \Bigr\{ \hat{v}_i \mid i = n + 1, \dots, nBins \Bigr\} 
\end{align}
 where $n = \frac{nBins}{2}$, indicating the half number of the orientation histogram $\mathcal{H}$. 
 The value of $nBins$ is empirically set to 72 in the experiments (i.e., the range of bin is $\frac{180^\circ}{72} =  5^\circ$)).

\begin{figure}[!t]
    \centering
    \begin{subfigure}[b]{0.45\textwidth}
        \includegraphics[width=\textwidth]{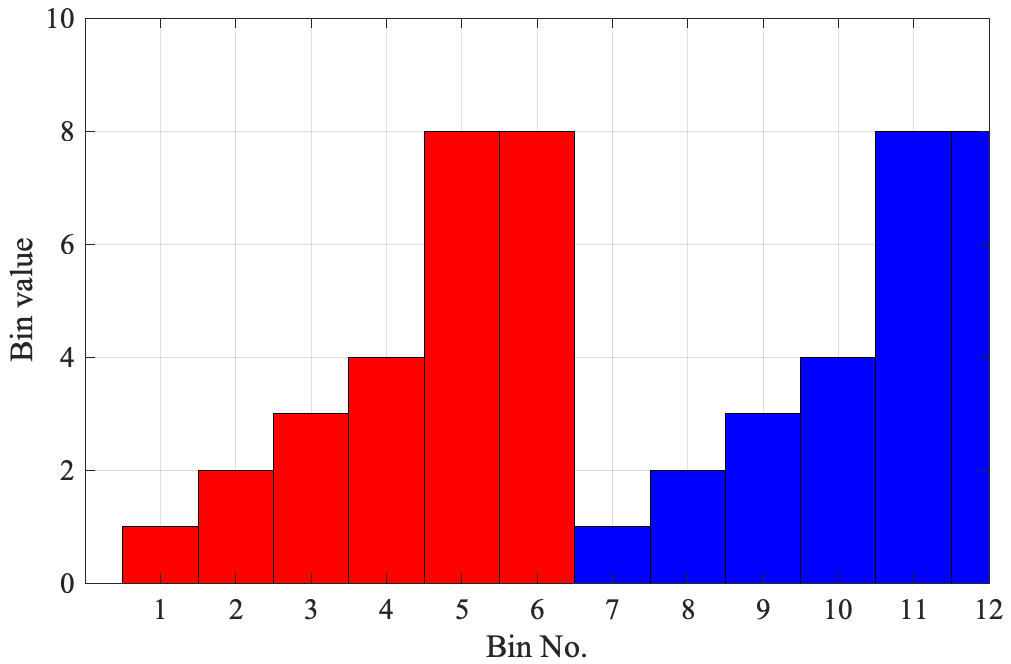}
        \caption{Histogram}\label{fig:cumhistIllustration_1}
    \end{subfigure}
    \begin{subfigure}[b]{0.45\textwidth}
        \includegraphics[width=\textwidth]{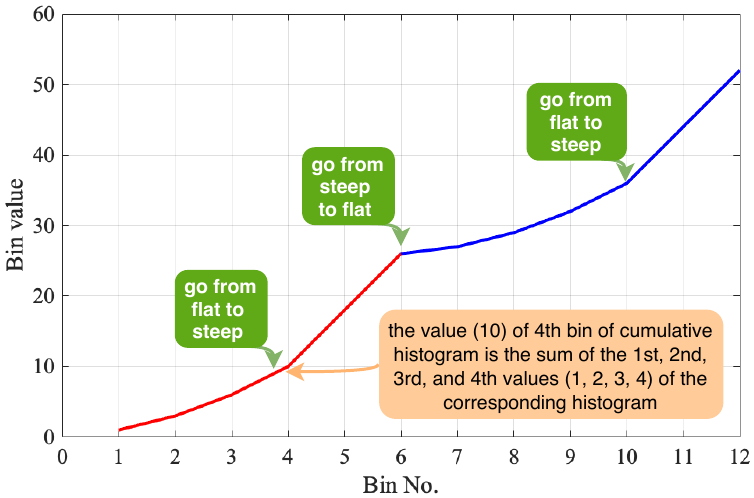}
        \caption{Cumulative histogram}\label{fig:cumhistIllustration_2}
    \end{subfigure}
    \caption{
        Illustration of an example histogram and its corresponding cumulative histogram. 
        The cumulative histogram is plotted as a line graph rather than bar chart for better observation of the level of symmetry and smoothness. 
        The right half (blue) of the histogram is {\it translational symmetric} to the left half (red). 
        (\subref{fig:cumhistIllustration_2}): The three incremental changing points represent the three sharp changes (from bin 4 to bin 5, 6 to 7, and 10 to 11), respectively.
    }
    \label{fig:cumhistIllustration}
\end{figure}

\subsubsection{Cumulative Orientation Histogram}
\label{subsubsec:cumulativeHistogram}
The {\it symmetry} of an orientation histogram can be calculated directly by comparing the left and right parts of the orientation histogram and its cumulative histogram using similarity measures. 
The smoothness of the orientation histogram is difficult to evaluate directly from orientation histogram. 
Thus, the {\em cumulative} orientation histogram (see \figureautorefname \ref{fig:cumhistIllustration}) is calculated as an alternative to an orientation histogram (Step (e)). 
A cumulative histogram is just a histogram whose values are cumulative frequencies rather than frequencies, and the value of its $k$th bin is calculated by: 
\begin{equation}
\mathcal{C}(k) = \sum_{i=1}^{k} bin(i), \quad k = 1,\dots, nBins
\end{equation}
where $bin(i)$ is the value of the $i$th bin of the orientation histogram. 
Using a cumulative orientation histogram, the smoothness is visually represented. 
The convex and concave parts of the cumulative orientation histogram reflect the increase and decrease in corresponding parts of the orientation histogram. 
The changes (increases and decreases) in the orientation histogram are reflected by the changing rate of increase in cumulative orientation histogram. 
The parts of the cumulative histogram curve that changes from {\it high gradient} to {\it no gradient}, or from {\it no gradient} to {\it high gradient}, may represent a sharp change at the corresponding parts in orientation histogram (see illustration in \figureautorefname \ref{fig:cumhistIllustration}).

\begin{figure}[t]
	\centering
	 \begin{subfigure}[t]{0.45\textwidth}
                  \includegraphics[width=\textwidth]{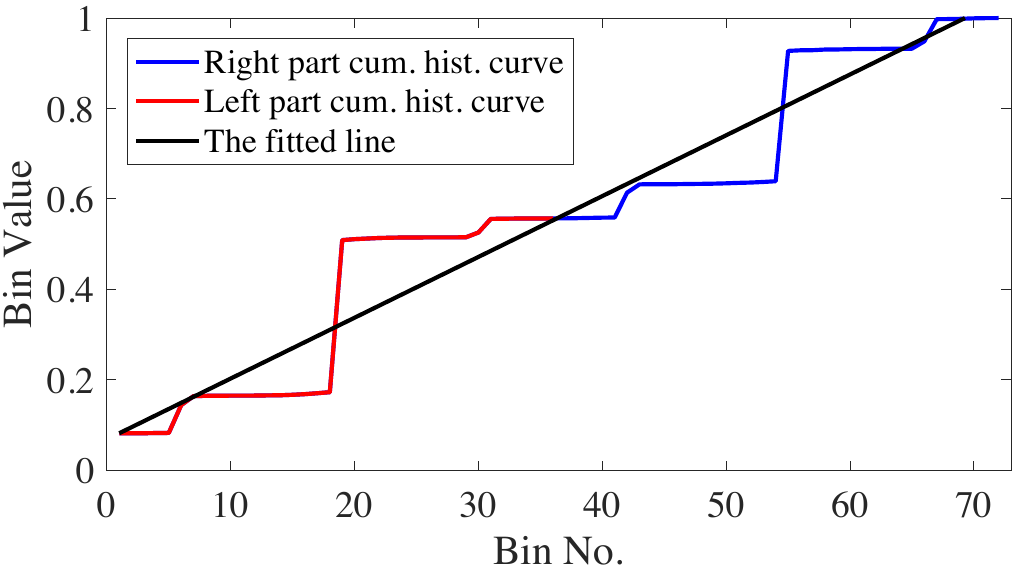}
                  \caption{Cumulative histogram curve}
                  \label{fig:curveIllustration-1}
         \end{subfigure}
         \begin{subfigure}[t]{0.45\textwidth}
                  \includegraphics[width=\textwidth]{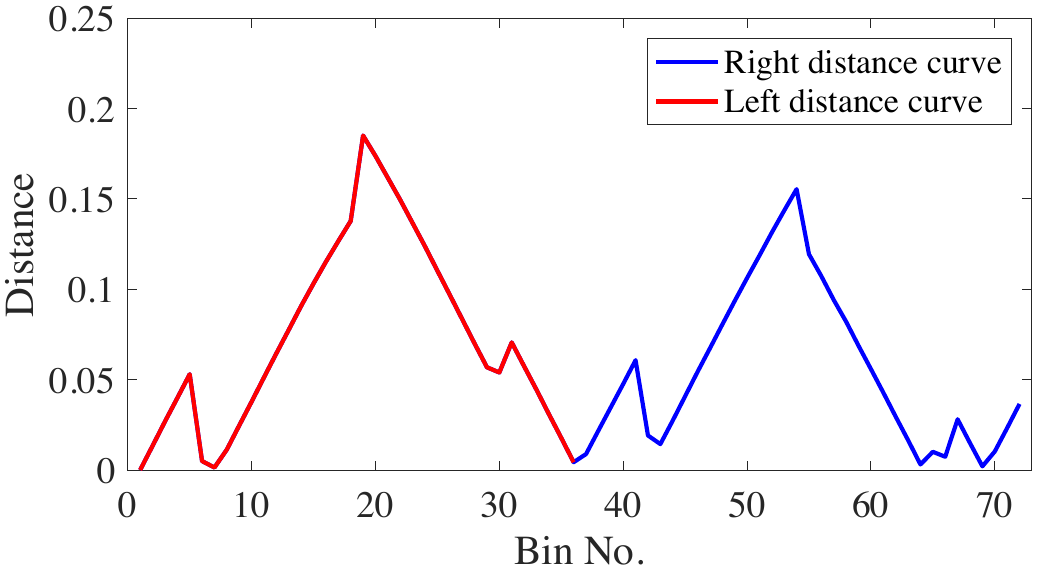}
                  \caption{Distance curve}
                  \label{fig:curveIllustration-2}
         \end{subfigure}
         \caption{
            An example cumulative histogram curve with a total number of 72 bins ($5^\circ$ for each bin), and its corresponding distance curve: the left red curve (bins from 1 to 36) and the right blue curve (bins from 37 to 72). 
         }
         \label{fig:curveIllustration}
 \end{figure}

The next step calculates the corresponding distance curve of the cumulative histogram curve, as shown in Step (e) and (f) of \figureautorefname \ref{fig:SOHFeatureVectorConstruct}. 
A straight line needs to be fitted to the points at the cumulative histogram curves, and the resultant fitted line is the black straight line as shown in \figureautorefname \ref{fig:curveIllustration-1}. 
Subsequently, the perpendicular distances from the points at the cumulative histogram curve to this straight line is calculated, generating the distance curves shown in Step (f) of \figureautorefname \ref{fig:SOHFeatureVectorConstruct}. 
The left and right parts of the cumulative histogram curve or of the distance curves can be used to calculate similarity using the Procrustes distance \citep{klingenberg2015analyzing} (as illustrated in \figuresautorefname \ref{fig:superimpositionIllustration} and \ref{fig:procrustes}) or other similarity measures.

It is known that orientation histogram is sensitive to image rotation \citep{freeman1995orientation}. 
However, we observe that its level of smoothness and symmetry is invariant to image rotations. 
We therefore conducted a study on the effect of image rotations on these properties of the resulting orientation histogram, intuitively providing evidence to this invariance.
Due to limited space, the result is presented in Appendix~\ref{sec:effectOrientationHistogram}.

\subsubsection{Variables of the SOH Feature Vector}
\label{subsubsec:calcSOHFeatureVector}
As described in previous sections, the SOH feature vector is based on the analysis of the shape of the orientation histogram, by quantifying the symmetry and smoothness of the orientation histogram. 
Artcodes have connected closed regions and blobs of which edges are often smooth; these characteristics of Artcodes can be used as cues to discover Artcodes, as they are represented by a certain level of smoothness and symmetry of their orientation histogram. 
These two properties are calculated from the distance curve, separating the distance curve into two equal-size parts, the left and right curves (denoted $\mathcal{C_L} $ and $\mathcal{C_R}$, as illustrated in \figuresautorefname \ref{fig:curveIllustration}), before calculating the similarity between the two curves using similarity metrics. 
The SOH feature vector mainly consists of three categories of variables (or components): variables for measuring symmetry, variables for measuring smoothness, and auxiliary variables. 
The following sections specify these three categories of variables.

\subsubsection*{Variables for Measuring Symmetry} 
Similarity metrics are used for measuring symmetry. 
Two classic similarity metrics: {\em Procrustes} \citep{moser1965volume} and {\em Chi-Square} \citep{pearson1900x} distances are adopted in this work.
Formally, the two curves with $n$ points are denoted as: 
\begin{align}
\mathcal{C_L} =\Bigl\{ (x_i, y_i) \mid i = 1, \dots, n \Bigr\} \\
\mathcal{C_R} = \Bigl\{ (u_i, v_i) \mid i = 1, \dots, n \Bigr\} 
\end{align}
 where $n = \frac{nBins}{2}$, indicating the half number of the cumulative histogram curve $\mathcal{C}$, and $x_i$ and $u_i$ are { the} corresponding $x$-coordinates of the $i$th curve points: $y_i$ and $v_i$.  \\

\begin{figure}[t]
    \centering
    \begin{subfigure}[t]{0.95\textwidth}
        \includegraphics[width=\textwidth]{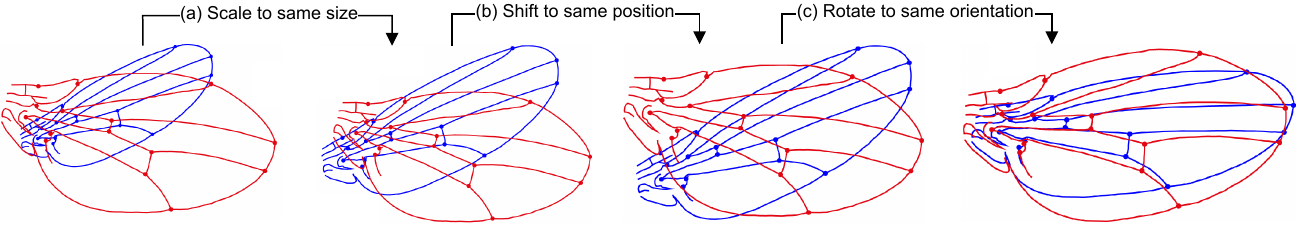} 
    \end{subfigure}
    \caption{
        An illustration of Procrustes superimposition: uniform scaling, translation, and rotation (reproduced from \citet{klingenberg2015analyzing}).
    }
    \label{fig:superimpositionIllustration}
 \end{figure}

\begin{figure} 
    \centering
    \begin{subfigure}[t]{0.45\textwidth}
        \includegraphics[width=\textwidth]{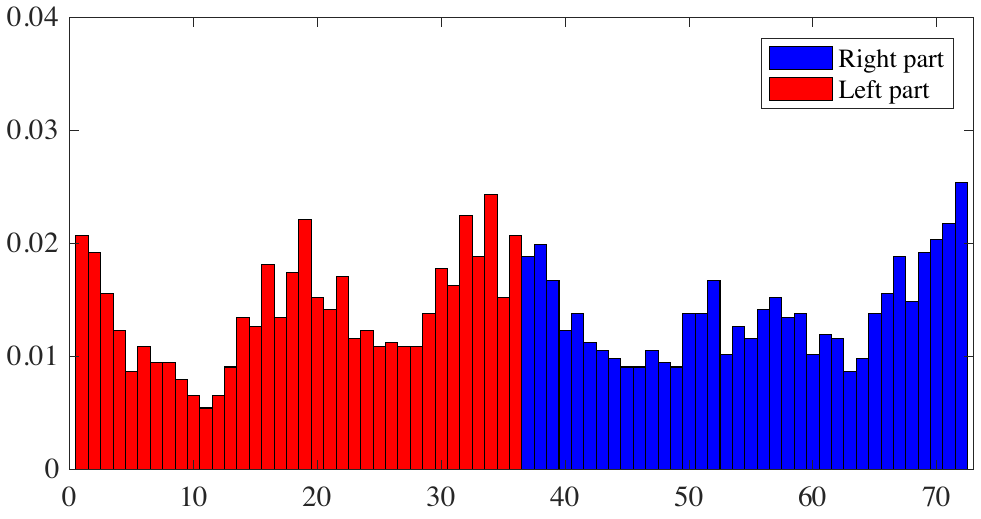}
        \caption{An example orientation histogram}
        \label{fig:procrustes-1}
    \end{subfigure}
    \begin{subfigure}[t]{0.45\textwidth}
        \includegraphics[width=\textwidth]{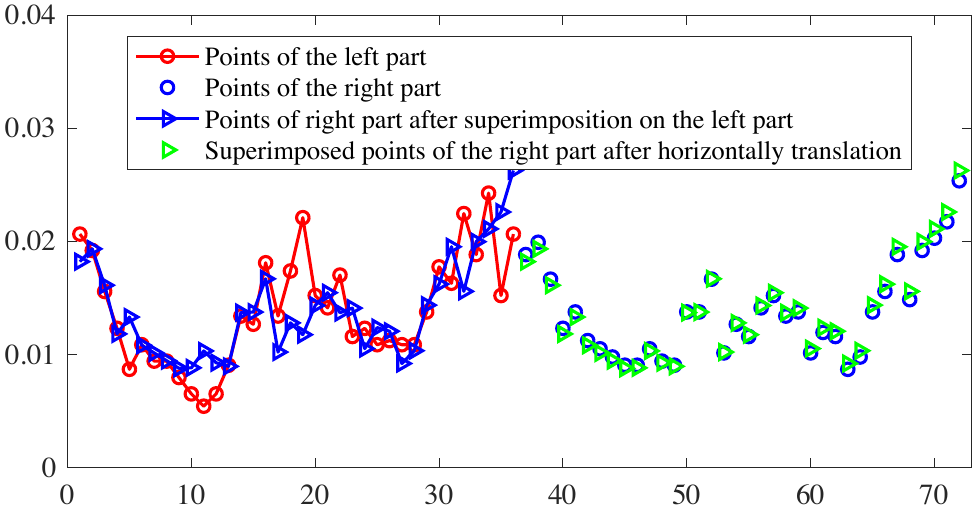}
        \caption{Bin values before and after Procrustes superimposition}
        \label{fig:procrustes-2}
    \end{subfigure}
    \caption{
        An example of an orientation histogram and its bin values before and after Procrustes superimposition. 
        (\subref{fig:procrustes-1}): An example orientation histogram: the left and right halves of the orientation histogram are marked by red and blue, respectively, and each have 36 bins (points). 
        (\subref{fig:procrustes-2}): Illustration of the corresponding bin values (frequency values) before and after superimposition: the orientation histogram in (\subref{fig:procrustes-2}) is presented using circles (red for the left part, and blue for the right); the superimposed points of the right part are denoted as blue triangles; and the green triangles are its horizontally translated version.
    }
    \label{fig:procrustes}
 \end{figure}

A Procrustes analysis \citep{moser1965volume,kendall1989survey} is a form of statistical shape analysis measuring the similarity between two shapes.
In this work, the left part ($\mathcal{C_L}$) of the orientation histogram is selected as the one to be compared, whereby the other shapes (i.e., the right part $\mathcal{C_R}$) are first superimposed and then compared to the left part. 
Before calculating of the Procrustes distance, it is necessary to perform Procrustes superimposition (see \figureautorefname \ref{fig:superimpositionIllustration} for an illustration of Procrustes superimposition): {\em translation}, {\em rotation}, {\em uniform scaling}, and {\em reflection}. 
After transformation, the Procrustes distance between the two curves $\mathcal{C_L}$ and $\mathcal{C_R} $, is calculated by:
\begin{equation}\label{eq:proDistance}
	 prodist(\mathcal{C_L}, \mathcal{C_R}) = \sqrt{\sum_{i=1}^{n}\Big((\hat{x}_i - \hat{u}_i)^2 + (\hat{y}_i - \hat{v}_i)^2 \Big)}
\end{equation}
where $(\hat{x_i}, \hat{y}_i)$ and $(\hat{u_i}, \hat{v}_i)$ are the coordinates of the $i$th points of curves $\mathcal{C_L}$ and $\mathcal{C_R}$ after superimposition. 
However, in our case, $\mathcal{C_R}$ is at the same scale as $\mathcal{C_L}$, and the $x$-coordinates of $\mathcal{C_R}$ are horizontally translated from these of $\mathcal{C_L}$; therefore, the $x$-coordinates of $\mathcal{C_L}$ and $\mathcal{C_R}$ after Procrustes superimposition are approximately equivalent, \equationautorefname\ref{eq:proDistance} can thus be simplified as follows: 
\begin{equation}\label{eq:simplifiedProDistance}
	 prodist(\mathcal{C_L}, \mathcal{C_R}) \approx \sqrt{\sum_{i=1}^{n}\Big((\hat{y}_i - \hat{v}_i)^2 \Big)}
\end{equation}

As illustrated in \figureautorefname \ref{fig:procrustes}, the points of the right part after Procrustes superimposition are slightly offset compared to the original points (see the blue circles and their superimposed version as denoted by the green triangles). 
After superimposition, the Procrustes distance between the left and the right parts of the orientation histogram is essentially the Euclidean distance (\equationautorefname \ref{eq:simplifiedProDistance}) between the points of the left part and the superimposed points of the right part. 
Due to the superimposition, the Procrustes distance is more effective than the Euclidean distance in comparing shapes as a whole. 
Therefore, the Procrustes distance is useful for calculating shape similarity and thus the level of symmetry.

Despite its effectiveness for comparing shapes, the Procrustes distance is sensitive to noise. 
As presented in \equationautorefname\ref{eq:simplifiedProDistance}, the distance is based on the sum of the square of the difference between two comparison points. 
Consequently, a big difference only between two points would largely increase the Procrustes distance. 
In real situations, this sharp difference between $y_i$ and $v_i$ is probably due to outliers. 
The difference on some of the point pairs should not therefore take much weight on the similarity comparison as a whole. 
Accordingly, {\it Chi-square} distance ($\chi^2$) is introduced to alleviate this problem. 
The square of differences between the pair of points is divided by their sum, i.e., each term is not the square absolute difference between the two bin values but rather a relative difference. Therefore, the Chi-square distance could be complementary to the Procrustes distance in measuring the similarities between histograms.

In this case, the Chi-square distance between the two curves $\mathcal{C_L}$ and $\mathcal{C_R}$ is calculated using the equation as follows:
\begin{equation}\label{eq:chiDistance}
    \chi^2(\mathcal{C_L}, \mathcal{C_R}) = \sum_{i=1}^{n}\frac{(y_i  - v_i)^2}{(y_i + v_i)}
\end{equation}
The symmetrical property of the orientation histogram rather than the cumulative histogram is also measured by the similarity between the left and right parts of the orientation histogram $\mathcal{H}$ by the Procrustes distance ($prodist(\mathcal{H_L}, \mathcal{H_R})$) and the Chi-Square distance ($\chi^2(\mathcal{C_L}, \mathcal{C_R})$). 
As derivative measures the changing rate, the symmetry of the orientation histogram is also measured by computing the similarity between the first derivatives of its corresponding distance curves ($\mathcal{C_L}$ and $\mathcal{C_R}$) using the Procrustes and Chi-square distances, denoted as $prodist(d\mathcal{C_L}, d\mathcal{C_R})$ and $\chi^2(d\mathcal{C_L}, d\mathcal{C_R})$, respectively.

\begin{table}[!t]
\caption{The full list of variables of the SOH feature vector. 
$S_1$ -- $S_6$: variables for measuring the symmetry of an orientation histogram; 
$S_7$ -- $S_8$: variables for measuring the smoothness of an orientation histogram; 
$S_9$ -- $S_{12}$: auxiliary variables.}
\label{tbl:SOHVariableList}
\centering
\small{
    \begin{tabularx}{\textwidth}{llX}
    \hline
    Variable & Equation & Description \\ \hline
    $S_1$ & $prodist(\mathcal{C_L}, \mathcal{C_R})$ &  Measure symmetry property of the orientation histogram $\mathcal{H}$ using the Procrustes distance between the curves $\mathcal{C_L}$ and $\mathcal{C_R}$ \\ \hline
    $S_2$ & $\chi^2(\mathcal{C_L}, \mathcal{C_R})$ &  Measure symmetry property of the orientation histogram $\mathcal{H}$ using the Chi-Square distance between the curves $\mathcal{C_L}$ and $\mathcal{C_R}$ \\ \hline
    $S_3$ & $prodist(d\mathcal{C_L}, d\mathcal{C_R})$ &  Measure the symmetry property of the orientation  histogram $\mathcal{H}$ using the Procrustes distance between the derivatives of the curves $\mathcal{C_L}$ and $\mathcal{C_R}$ \\ \hline
    $S_4$ & $\chi^2(d\mathcal{C_L}, d\mathcal{C_R})$ &  Measure the symmetry property of the orientation  histogram $\mathcal{H}$ using Chi-Square distance between the derivatives of the curves $\mathcal{C_L}$ and $\mathcal{C_R}$ \\ \hline
    $S_5$ & $prodist(\mathcal{H_L}, \mathcal{H_R})$ &  Measure the symmetry property of the orientation histogram $\mathcal{H}$ using the Procrustes distance between the left and right parts ($\mathcal{H_L}$ and $\mathcal{H_R}$) of $\mathcal{H}$ \\ \hline
    $S_6$ & $\chi^2(\mathcal{H_L}, \mathcal{H_R})$ &  Measure the symmetry property of the orientation  histogram  $\mathcal{H}$ using the Chi-Square distance between the left and right parts ($\mathcal{H_L}$ and $\mathcal{H_R}$) of $\mathcal{H}$ \\ \hline
    $S_7$ & $\mu(\mathcal{R})$ &  Measure the smoothness property of the orientation histogram $\mathcal{H}$ by calculating the mean of the residual $\mathcal{R}$ \\ \hline
    $S_8$ & $\sigma(\mathcal{R})$ &  Measure the smoothness property of the orientation histogram $\mathcal{H}$ by calculating the standard deviation of the residual $\mathcal{R}$ \\ \hline
    $S_9$ & $\mu(\mathcal{I})$ &  Combine with $S_{10}$ to measure the intensity variations of the image $\mathcal{I}$, the mean of $\mathcal{I}$ \\ \hline
    $S_{10}$ & $\sigma(\mathcal{I})$ &  Combine with $S_{9}$ to measure the intensity variations of the image $\mathcal{I}$, the standard deviation of $\mathcal{I}$ \\ \hline
    $S_{11}$ & $\mu(\mathcal{I}_{edge})$ &  Combine with $S_{12}$ to measure the intensity variations of the edge pixels $\mathcal{I}_{edge}$, the mean of $\mathcal{I}_{edge}$ \\ \hline
    $S_{12}$ & $\sigma(\mathcal{I}_{edge})$ &  Combine with $S_{11}$ to measure the intensity variations of the edge pixels $\mathcal{I}_{edge}$, the standard deviation of $\mathcal{I}_{edge}$ \\ \hline
    \end{tabularx}
}
\end{table}

\subsubsection*{Variables for Measuring Smoothness} 
The smoothness of the orientation histogram is measured by the mean ($\mu$) and standard deviation ($\sigma$) of the {\it residual} vector (\figureautorefname\ref{fig:SOHFeatureVectorConstruct} Step (e)), where the residual is the {\it difference} between the fitted line $\mathcal{P}$ (the black line in \figureautorefname \ref{fig:SOHFeatureVectorConstruct} Step (e)) and cumulative histogram curve $\mathcal{C}$ (the red and blue curves in \figureautorefname \ref{fig:SOHFeatureVectorConstruct} Step (e)).
The residual vector therefore has the same dimensions as the cumulative histogram. 
Formally, the residual vector $\mathcal{R}$ is denoted as: 
\begin{equation}
    \mathcal{R} =  \bigg\{ r_i \mid  r_i = \abs*{\mathcal{P}_i - \mathcal{C}_i}, i = 1, \dots, nBins \bigg\}
\end{equation}
where  $\mathcal{P}_i$ and $\mathcal{C}_i$ are the values of the fitted line and the $i$th bin of the cumulative histogram, respectively. 
The two variables, the {\it mean} and {\it standard deviation} (st. dev.) of $\mathcal{R}$, are then used to describe the smoothness of the orientation histogram. 
Up to now the SOH feature vector has eight feature variables, describing the symmetry and smoothness of the orientation histogram of the image and thus representing the general topological structures. 
The next section present the auxiliary variables for the SOH feature vector.

\subsubsection*{Auxiliary Variables} 
The aforementioned eight measurements are based on the observation that the recognisable part of an Artcode is the connected closed regions with high-contrast edges. 
The four additional measurements discussed below are based on two observations: 
(1) the overall intensity of the recognisable foreground of Artcode is darker than the background part; and 
(2) the pixels on the boundaries of the recognisable foreground have nearly equal intensities. Therefore, mean and standard deviation are employed to quantify these two observations.

Let $\mathcal{I}$ and $\mathcal{I}_{edge}$ denote the grayscale input image and the grayscale pixels at the boundaries found using the edge map detected by Sobel edge detection method, the four measurements are then given as follows: $\mu(I)$, $\sigma(I)$, $\mu(I_{edge})$, and $\sigma(I_{edge})$, where $\mu(\cdot)$ and $\sigma(\cdot)$ denote the mean and standard deviation of the input, respectively.

According to above analysis, these twelve measurements are the full list of variables used to construct a full SOH feature vector, as described in \tableautorefname\ref{tbl:SOHVariableList}.
By selecting subsets of these features, five variants of the SOH feature vector with varied dimensions are constructed, which will be further studied in experiments. 
\tableautorefname\ref{tbl:SOHVariableList} shows the full list of the SOH feature variables: including the aforementioned three types of variables: variables for measuring symmetry ($S_1$ -- $S_6$) and smoothness ($S_7$ -- $S_8$) of the orientation histogram, and auxiliary variables ($S_9$ -- $S_{12}$).

\section{Experimental Studies}
\label{sec:experiments}
The previous section describes the SOH feature vector, this section details experimental studies on this feature descriptor for detecting Artcode proposals. 
We conducted two experiments.
The first experiment evaluates the performance of feature sets: compare SOH variants with traditional geometrical descriptors on generating Artcode proposals. 
BoW and HoG are traditionally widely-used feature descriptors for image categorisation and object detection before the era of deep learning. 
They have been proven to be effective for describing the traditional semantic object class that shares the common geometric characteristics. 
However, it is not yet clear whether BoW and HoG may also be effective in generating proposals for such topological objects as Artcodes, especially when compared to the specially designed SOH features. 
We therefore conducted the second experiment to compare their performance on generating Artcode proposals. 
This experiment was evaluated in terms of the two most particular metrics for proposal detection evaluation --- recall and efficiency --- in addition to other metrics, such as accuracy, precision, true negative rate (TNR), F2 measure, and the Matthews correlation coefficient (MCC). 
The second experiment studies the {\it performance} of the classifier with the selected SOH variant using a balanced dataset.

\subsection{Datasets}
\label{sec:datasets}
Two datasets were collected for experimental studies. 
The first dataset is referred to as the True Artcode Dataset (TAD) and the second as the Extended Artcode Dataset (EAD). 
The TAD is the smaller and more imbalanced one of the two, but contains only true Artcodes; while the EAD is larger and more balanced, and is obtained by adding some simulated Artcode images. 
Imbalance handling techniques were also adopted to alleviate the impact of the imbalance of the TAD, resulting in an additional SMOTE-augmented True Artcode Dataset (SaTAD), described in \sectionautorefname\ref{subsec:handlingImbalance}.

\begin{figure}[t]
     \centering  
     \begin{subfigure}[b]{0.22\textwidth}
        \includegraphics[width=\textwidth]{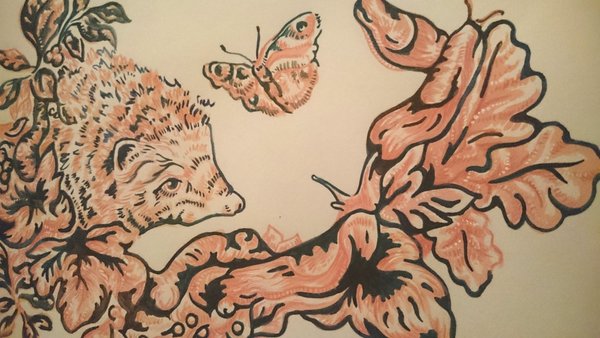}
     \end{subfigure}
     \begin{subfigure}[b]{0.125\textwidth}
        \includegraphics[width=\textwidth]{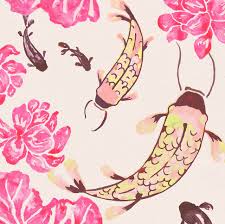}
     \end{subfigure}
      \begin{subfigure}[b]{0.14\textwidth}
        \includegraphics[width=\textwidth]{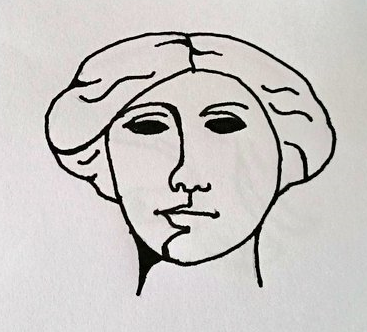}
     \end{subfigure}
     \begin{subfigure}[b]{0.16\textwidth}
        \includegraphics[width=\textwidth]{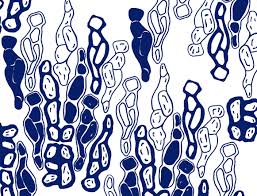}
     \end{subfigure}
     \begin{subfigure}[b]{0.16\textwidth}
        \includegraphics[width=\textwidth]{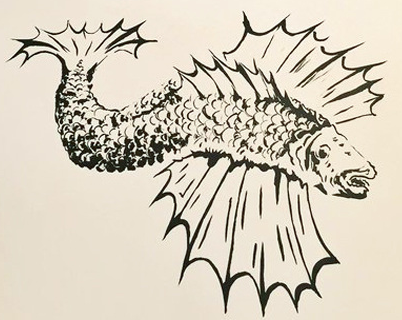}
     \end{subfigure} 
     \begin{subfigure}[b]{0.155\textwidth}
        \includegraphics[width=\textwidth]{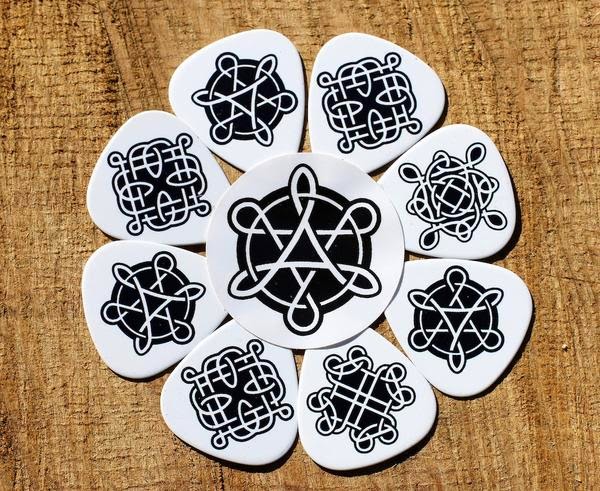}
     \end{subfigure}
     \caption{
        Artcode example selected from the True Artcode Dataset. 
        Artcodes have a flexible geometry and contain a variety of visual elements and are visually ``hidden'' markers that can be decoded, like barcodes and QR codes. 
        Artcodes in this dataset are created by designers or end users in previous Artcode relevant projects and workshops. 
     }
     \label{fig:artcodesExamples} 
\end{figure}

\begin{figure}
     \centering
     \begin{subfigure}[b]{0.18\textwidth}
        \includegraphics[width=\textwidth]{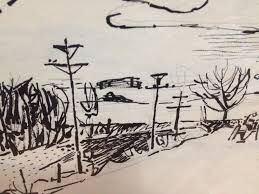}
     \end{subfigure}
     \begin{subfigure}[b]{0.135\textwidth}
        \includegraphics[width=\textwidth]{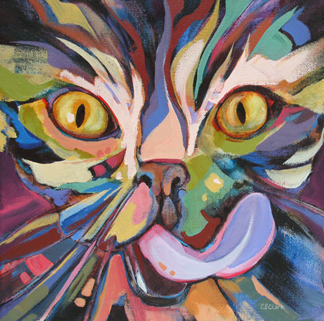}
     \end{subfigure}
     \begin{subfigure}[b]{0.135\textwidth}
        \includegraphics[width=\textwidth]{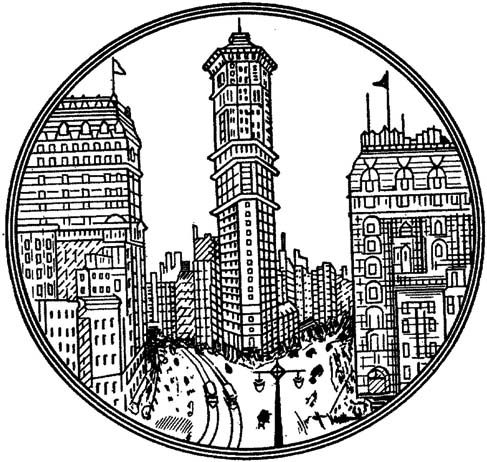}
     \end{subfigure}
     \begin{subfigure}[b]{0.16\textwidth}
        \includegraphics[width=\textwidth]{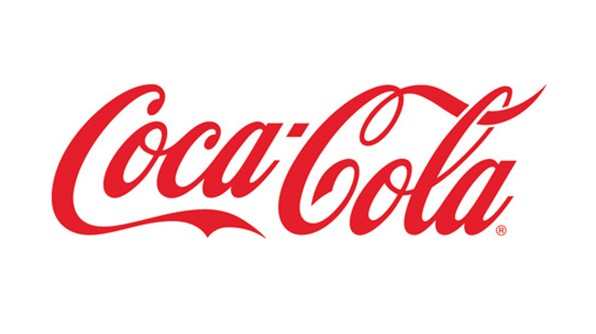}
     \end{subfigure} 
     \begin{subfigure}[b]{0.19\textwidth}
        \includegraphics[width=\textwidth]{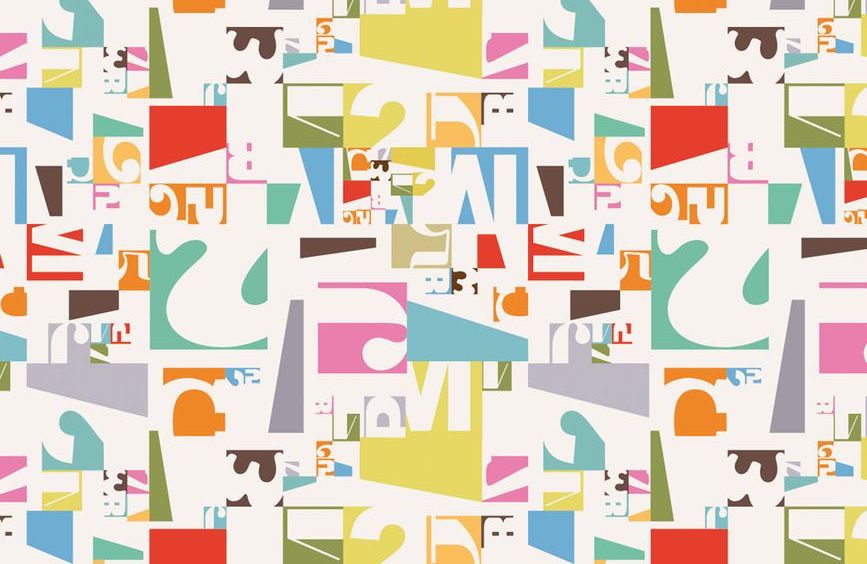}
     \end{subfigure} 
     \begin{subfigure}[b]{0.16\textwidth}
        \includegraphics[width=\textwidth]{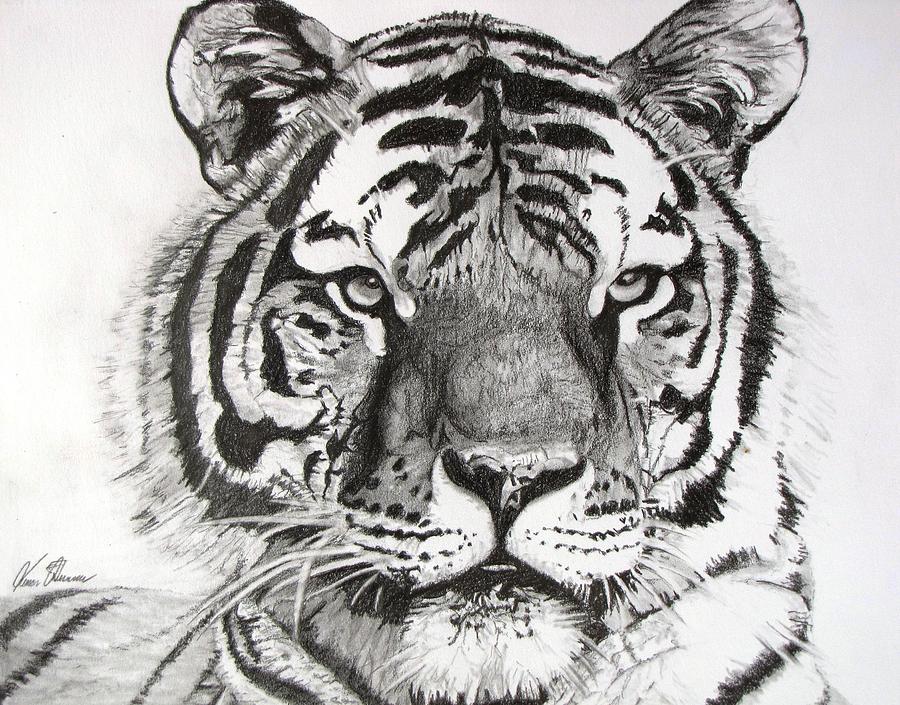}
     \end{subfigure}
     \caption{
        Non-artcode examples selected from the TAD dataset. 
        They are deliberately selected in order to confuse the Artcode presence recognition system. 
        They are visually very similar to the Artcodes in \figureautorefname \ref{fig:artcodesExamples}, and are challenging to classify through visual inspection alone.
     }
     \label{fig:nonartcodesExamples}
\end{figure}

\subsubsection{True Artcode Dataset (TAD)}
\label{subsec:dataset_tad}
Artcode classification, the aim of which is to label an input image as an Artcode (i.e. Artcode-like) or not, is a basic procedure for generating Artcode proposals. 
Ts so study the Artcode classification problem, a dataset called the True Artcode Dataset (TAD), which contained only {\it true} Artcode and non-Artcode images was created. 
Artcodes are real, decodable Artcode images in real-world situations, while non-Artcodes are images that do not include any Artcodes. 
\figureautorefname \ref{fig:artcodesExamples} presents some Artcode examples selected from the TAD dataset, including Artcode samples of {\em flowers}, {\em fishes}, {\em a human face}, and {\em Celtic knots}, which demonstrate very different geometry. 
Rather than completely randomly selected images, the non-Artcodes were intentionally picked to be confusing for the classifier. 
Some images, such as landscape images, would be very easy to differentiate from Artcodes, and hence, they are excluded from this dataset for study. 
Consequently, this dataset will be a tough test for the SOH. 
The aim here is a strict distribution between true, aesthetic Artcodes, and other similar non-Artcode images --- a very challenging test.

\figureautorefname \ref{fig:nonartcodesExamples} shows six non-Artcode images, including a {\em nature scene drawing}, {\em illustration}, a {\em tiger}, the {\em Coca-Cola logo}. 
Although they are human-created images that are visually very similar to Artcodes, they do not have any predefined topological structure. 
It is difficult, or even impossible, to classify the two groups of images through inspection of the appearance and geometry alone. 
This dataset includes non-Artcodes collected from various human-created images, like {\em logos}, {\em drawings}, {\em advertisements}, {\em paintings} and {\em graphics}. 
We exclude other image types such as {\em natural scenes}, {\em human images} and {\em daily life images} because these categories are obviously different from Artcodes in visual aspects and they would be easy to predict their classes.

Because Artcodes are manually created by designers, the number currently available is small (especially considering the large number of available non-Artcodes), but work is ongoing to extend this dataset. 
A consequence of this limited number of available true Artcodes is an imbalanced dataset, with a larger number of non-Artcode images: it contains 47 Artcode and 116 non-Artcode images (the entire dataset is publicly available through the link 
\footnote{\url{http://doi.org/10.5281/zenodo.2248036}}). 
However, this cannot reflect the expected usage in real life, in which many more non-Artcode images will be presented to the camera than Artcodes.

\begin{figure}[!t]
    \centering
    \begin{subfigure}[b]{0.105\textwidth}
        \includegraphics[width=\textwidth]{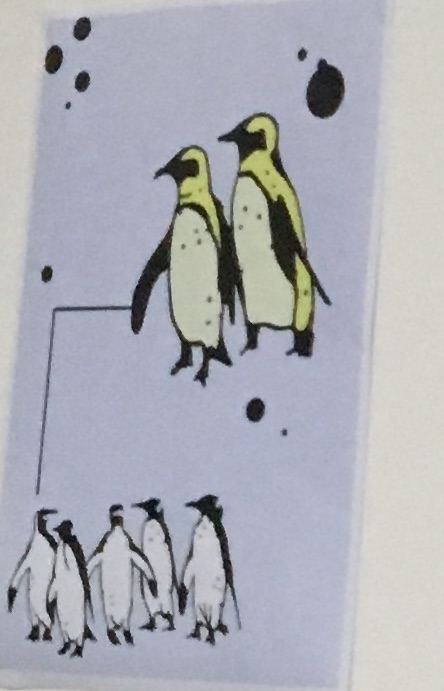}
    \end{subfigure}
    \begin{subfigure}[b]{0.165\textwidth}
        \includegraphics[width=\textwidth]{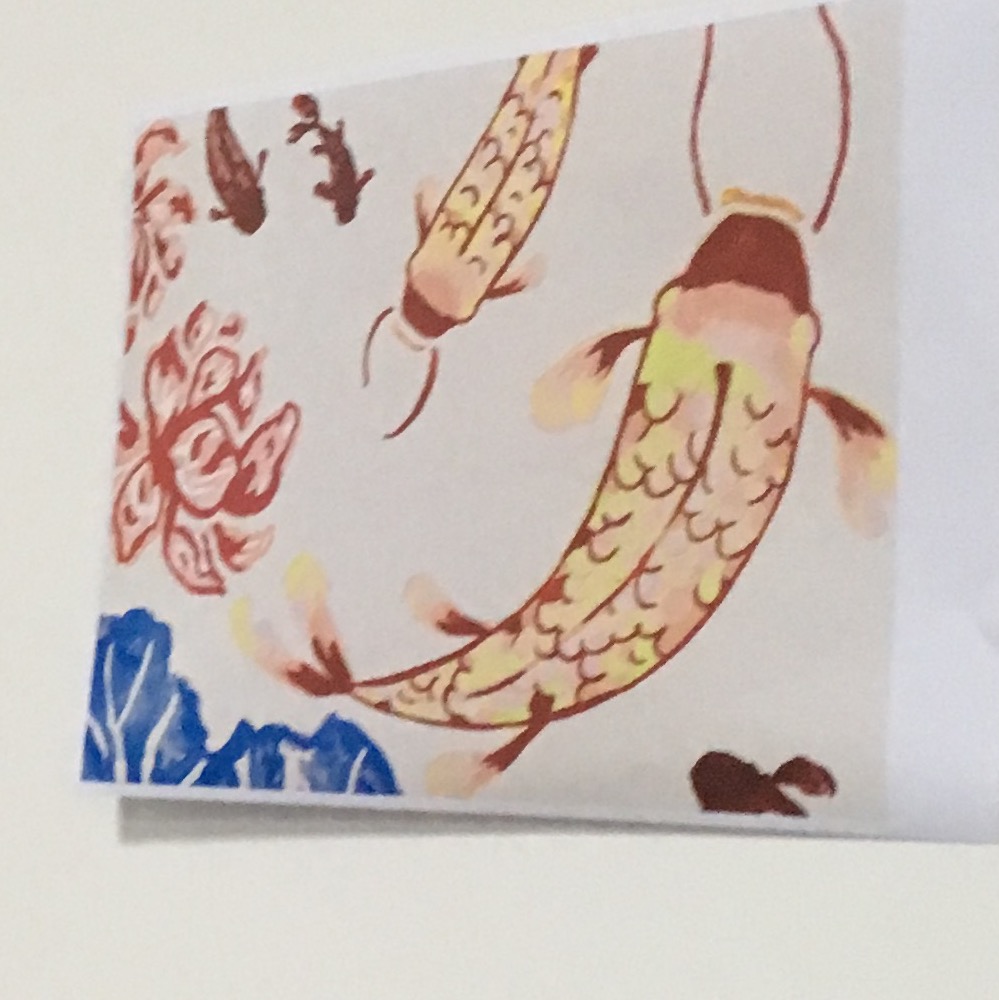}
    \end{subfigure}
    \begin{subfigure}[b]{0.1575\textwidth}
        \includegraphics[width=\textwidth]{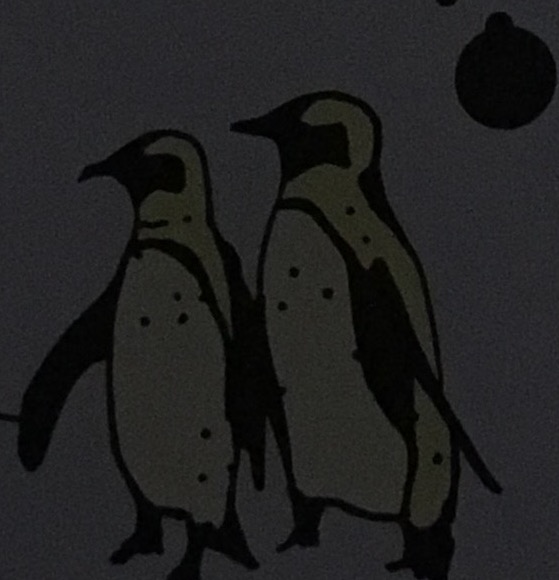}
    \end{subfigure} 
    \begin{subfigure}[b]{0.21\textwidth}
        \includegraphics[width=\textwidth]{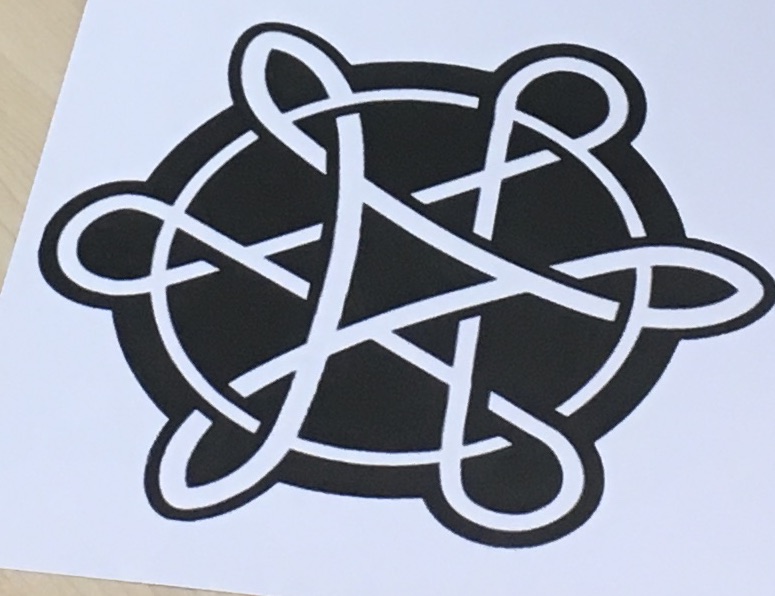}
    \end{subfigure}
    \begin{subfigure}[b]{0.325\textwidth}
        \includegraphics[width=\textwidth]{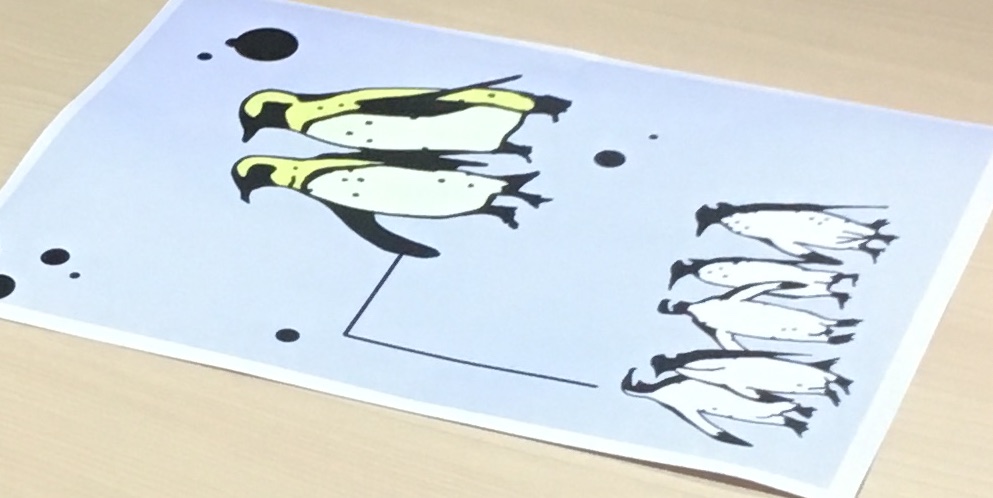}
    \end{subfigure} 
    \caption{
        Positive examples (Artcodes) added into the TAD dataset in \sectionautorefname\ref{subsec:dataset_tad}. 
        Newly added images are captured under different illuminations and camera poses.
    }
    \label{fig:additionalDataset}
\end{figure}

\subsubsection{Extended Artcode Dataset (EAD)}
\label{sec:dataset_EAD}
The TAD dataset is smaller and unbalanced, which includes many more negative examples (i.e., non-Artcodes). 
We therefore created a relatively more balanced dataset called the Extended Artcode Dataset (EAD) by adding ``simulated'' Artcode images to the TAD dataset.
The new added images are not {\it realistic} Artcode artefacts but instead are {\it printouts} (see examples in \figureautorefname \ref{fig:additionalDataset}) of Artcode examples under different illuminations and camera poses. 
The resultant dataset is more balanced than the TAD dataset, which includes 75 positive (Artcode) and 116 negative (non-Artcode) examples. 
This dataset is also publicly available through the link. \footnote{\url{http://doi.org/10.5281/zenodo.2248036}}

Although the EAD is larger and more balanced than the TAD, it contains many ``fake'' Artcodes, which are printouts of some existing Artcode samples in TAD, rather than being specially created on a specific medium, e.g., ceramic with new Artcodes. 
Therefore, to some extent, it can be considered an augmented dataset, albeit augmented in the physical world.

\subsection{Evaluation Metrics}
\label{subsec:performanceMetrics}
Considering the sensitivity of single performance metrics, a group of measurements were selected to provide an informative view of the proposed classifier's performance. 
Recall,  precision, True Negative Rate (TNR), accuracy,  F$_\beta$ measure and Matthews Correlation Coefficient (MCC) \citep{matthews1975comparison} were all used to examine the performance. 
Based on the confusion matrix in \tableautorefname\ref{tbl:confusionMatrix}, these metrics are defined as follows:

\begin{equation}\label{eq:rec}
	recall = \frac{TP}{TP+FN}
\end{equation}
\begin{equation}\label{eq:prec}
	precision = \frac{TP}{TP+FP}
\end{equation}
\begin{equation}\label{eq:tnr}
	TNR = \frac{TN}{TN + FP }
\end{equation}
\begin{equation}\label{eq:acc}
	accuracy = \frac{TP+TN}{TP + FP+FN+TN}
\end{equation}
\begin{equation}\label{eq:fbeta}
	F_\beta = \frac{(1+\beta)^2\times recall \times precision}{\beta^2 \times recall + precision}
\end{equation}
\begin{equation}\label{eq:mcc}
	MCC = \frac{TP \times TN - FP \times FN}{\sqrt{(TP+FP)(TP+FN)(TN+FP)(TN+FN)}}
\end{equation}

\begin{table}[!t]
\centering
\caption{Confusion matrix for Artcode prediction}
\label{tbl:confusionMatrix}
\small{
    \begin{tabular}{@{}lcll@{}}
    \hline
    \multicolumn{1}{c}{} &  & \multicolumn{2}{c}{Predicted} \\ \cline{3-4} 
    \multicolumn{1}{c}{} &  & \multicolumn{1}{c}{Artcode} & \multicolumn{1}{c}{Non-Artcode} \\ \hline
    \multirow{2}{*}{Actual} & \multicolumn{1}{c|}{Artcode} & True positives (TP) & False negatives (FN) \\
     & \multicolumn{1}{c|}{Non-Artcode} & False positives (FP) & True negatives (TN) \\ \hline
    \end{tabular}
}
\end{table}

Recall (sensitivity) and precision are two measures that focus on the positive examples (Artcodes) and predictions. 
Their importance varies over learning tasks --- precision, for example, being more desirable than recall in information retrieval. 
However, in the case of Artcode detection, it may be more desirable to have higher recall than precision, because recognizing the presence of Artcodes is the basis for later decoding in AR. Similarly, TNR (specificity) measures the proportion of negative examples  (non-Artcodes) that are correctly identified as such. 
Accuracy, the F$_\beta$ measure, and MCC measure the overall performance of the classifier, considering both positive and negative classes. 
Accuracy is the overall proportion of correct predictions, for both Artcode and non-Artcode classes, and is a simple way of describing a classifier's performance on the given dataset. 
However, accuracy is sensitive to size differences among classes. 
The F$_\beta$ measure uses the weighted harmonic average of precision and recall to evaluate the classifier's preciseness (how many positive samples it predicts correctly) and robustness (e.g., it does not miss a significant number of positive samples), where $\beta$ is used to control the importance of recall over precision. 
In this experiment, considering the desirability of recall over precision, the F2 measure ($\beta = 2$ ) was used, making recall twice as important as precision. 
Compared with accuracy, the F2 measure provides more insight into the performance of a classifier,  although the sensitivity is less. 
The F2 measure can also be as sensitive to data distributions as accuracy. 
MCC is, in essence, a correlation coefficient between the observed and predicted classifications, incorporating true and false positives and negatives. 
It is generally regarded as one of the best measures for classifier performance evaluation \citep{powers2011evaluation}, and remains effective even if the dataset is imbalanced.

\begin{figure}[t]
    \centering
    \begin{subfigure}[b]{0.45\textwidth}
        \includegraphics[width=\textwidth]{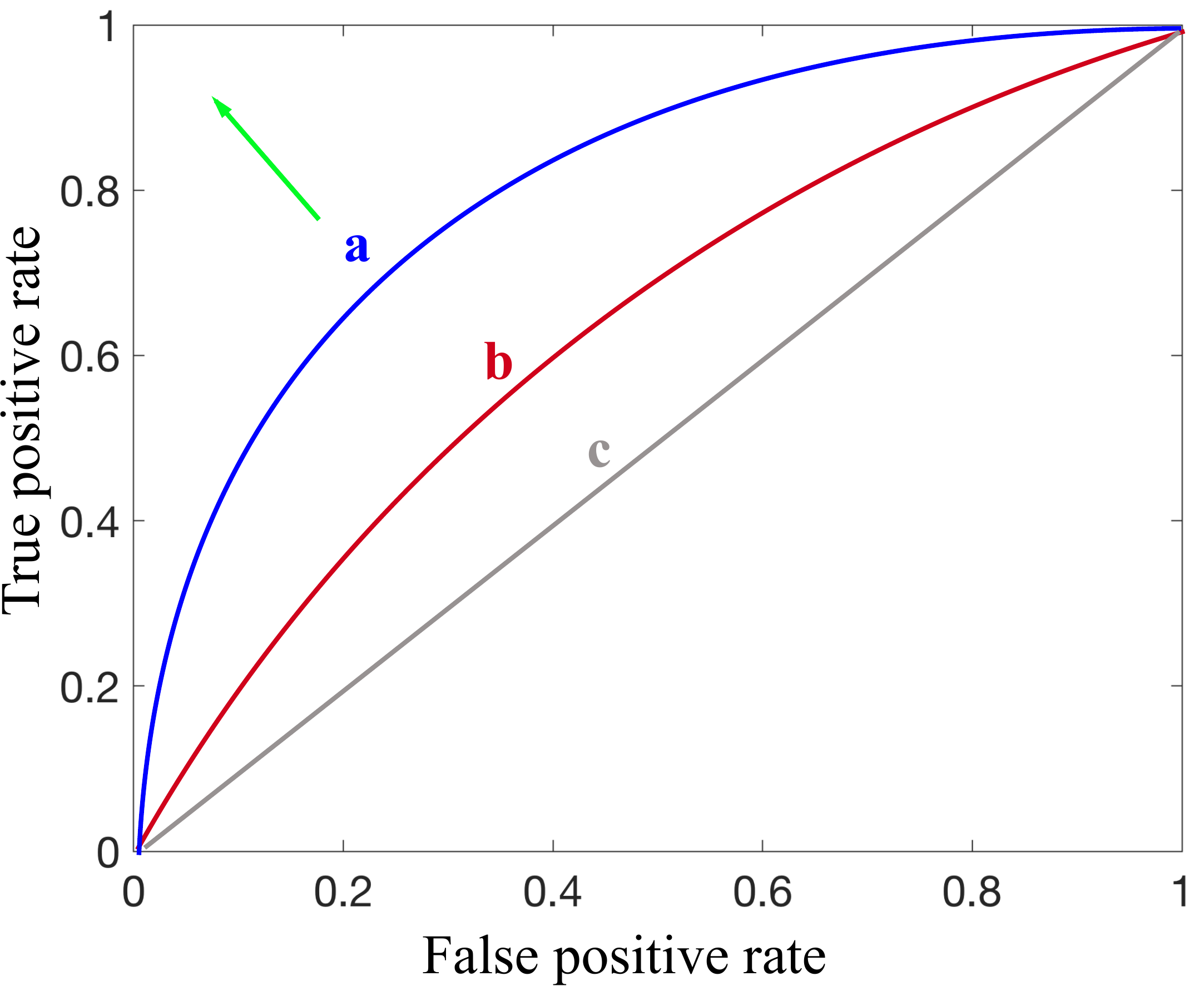}
        \caption{ROC curves}
        \label{fig:illustrationROC}
    \end{subfigure}
    \begin{subfigure}[b]{0.45\textwidth}
        \includegraphics[width=\textwidth]{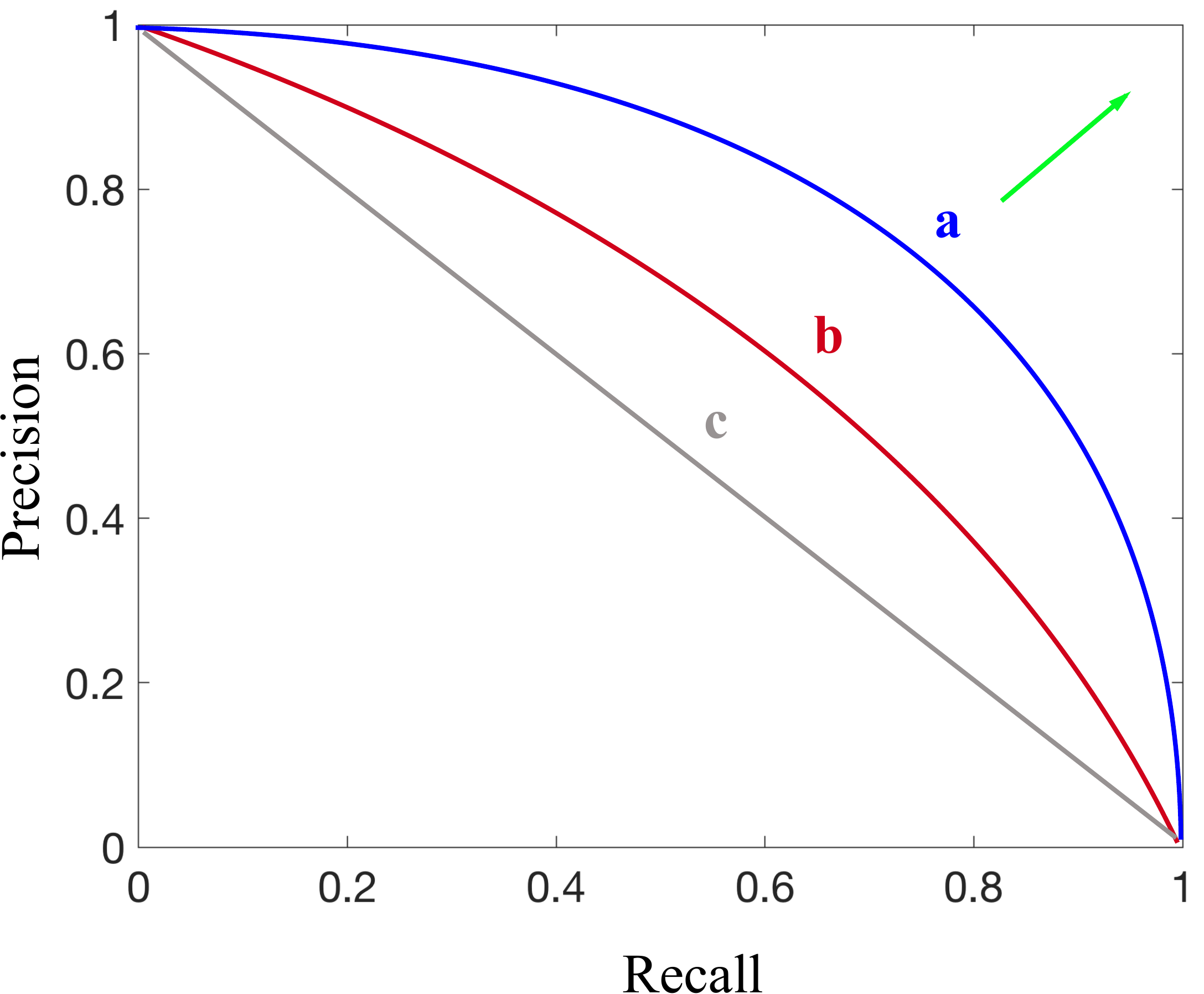}
        \caption{PR curves}
        \label{fig:illustrationPR}
    \end{subfigure}
    \caption{
        Illustration of the performance comparison between different ROC and PR curves.  
        Three curves labelled as {\color{blue}{\it a}}, {\color{red}{\it b}}, and {\color{gray}{\it c}} are plotted in both (\subref{fig:illustrationROC}) and (\subref{fig:illustrationPR}). 
        Curves {\it c} in both (\subref{fig:illustrationROC}) and (\subref{fig:illustrationPR}) indicate that the performance of the classifier is equivalent to guess by chance. 
        The curve that is closer to the top left corner (i.e., coordinates at $(0, 1)$) in the ROC plot (\subref{fig:illustrationROC}) indicates better performance of the classifier; the curve in PR plot that is closer to the top right corner (i.e., coordinate at $(1, 0)$) denotes the better performance. 
        Therefore, curves {\it a} in both (\subref{fig:illustrationROC}) and (\subref{fig:illustrationPR}) represent better performance of the classifier.
    }
    \label{fig:illustrationROC-PR}
\end{figure}

Additionally, Receiver Operating Characteristic (ROC) and Precision-Recall (PR) curves and their area under curve (AUC-ROC and AUC-PR) were used to illustrate the classifier's performance. ROC and PR curves are commonly used to evaluate the performance of the classifier in binary classification problems in machine learning. 
ROC curves plot the false positive rate on the {\em x}-axis versus the true positive rate on the {\em y}-axis, while PR curves plot recall on the {\em x}-axis and precision on the {\em y}-axis. 
The goal of the ROC curve is to be in the {\it upper-left-hand} corner, whereas the goal of the PR curve is to be in {\it upper-right-hand} corner. 
The closer the curves approach to these two corners (i.e., coordinates at (0, 1) and (1, 0)), the better the performance of the classifier. 
For example, \figureautorefname \ref{fig:illustrationROC-PR} illustrates three example ROC (\figureautorefname \ref{fig:illustrationROC}) and PR (\figureautorefname \ref{fig:illustrationPR}) curves (denoted a, b, and c), and the classifiers with the blue curves achieves the best performance ({\color{blue}a} $>$ {\color{red}b} $>$ {\color{gray}c}). 
The green allows represent the direction that the performance increases in.  
The straight, gray lines (c) uniformly separate the ROC and PR spaces, and denoting the performance of the classifier is equivalent to a chance guess (equal chance for both classes in binary classification problem). 
While ROC and PR curves provide a visual inspection of the classifier's performance, AUC-ROC and AUC-PR provide a numerical metric for evaluating the classifier's performance. Models with higher AUC values are preferred over those with lower AUC values. 
However, when dealing with highly skewed datasets, PR curves present a more informative picture of a classifier's performance --- the classifier performs better in a ROC plot would perform worse in the PR curve \citep{saito2015precision}.

\subsection{Experiment 1: Comparison of Feature Descriptors}
\label{sec:experiment1}
\subsubsection{Experimental Setting}
This experiment was conducted to compare the performance of the SOH feature descriptor (and its variants) with other two classic feature sets on the Artcode proposal generation task. 
The two selected feature sets for comparison are: the Bag of visual Words (BoW) and the Histogram of oriented Gradients (HoG). 
In addition to the three feature descriptors, as described in the following section, the TAD dataset was chosen for this study.
A five-fold cross validation was adopted for evaluating the classifier's performance, and the values of the performance metrics were the average values obtained after ten rounds of execution. 
The computational cost is the elapsed time for ten rounds of the five-fold cross validation.

\begin{table}
\centering
\caption{
    Description of the SOH variants under evaluation. 
    The definition of the SOH variants are recursively defined based on the previous SOH variants in this table. 
    The symbol $\cup$ denotes the union of the two sets.
}
\label{tbl:descriptionSOHVariants}
\small{
    \begin{tabularx}{\textwidth}{ llX }
    \hline
    SOH variant & Variables & Description \\ \hline
    SOH-05 & $\{S_1, S_2, S_3, S_7, S_8\}$ & Only contain the basic variables measuring the two properties (symmetry and smoothness) of orientation histogram. \\ \midrule
    SOH-07 & SOH-05  $\cup$ $\{S_9, S_{10}\}$ & Add $\{S_9, S_{10}\}$ (the mean and std. of the input image $\mathcal{I}$, i.e., $\mu(\mathcal{I})$ and $\sigma(\mathcal{I})$) into SOH-05. \\ \midrule
    SOH-08 & SOH-07 $\cup$ $\{S_4\}$ & Add $\{S_4\}$ ($\chi^2(d\mathcal{C_L}, d\mathcal{C_R})$, which measures the {\it Chi-Square} distance between the derivatives of curve $\mathcal{C_L}$ and $\mathcal{C_R}$) into SOH-07.  \\ \hline
    SOH-10 & SOH-08 $\cup$ $\{S_5, S_6\}$ & Add $\{S_5, S_6\}$ (measuring the similarity between the left and right parts of orientation histogram) into SOH-08. \\ \hline
    SOH-12 & SOH-10 $\cup$ $\{S_{11}, S_{12}\}$ & Add $\{S_{11}, S_{12}\}$ (the mean and std. of the edge pixels, i.e., $\mu(\mathcal{I}_{edge})$ and$\sigma(\mathcal{I}_{edge})$ ) into SOH-10. \\ \hline
    \end{tabularx}
}
\end{table}

\subsubsection*{Three Feature Descriptors}
The three feature descriptors for this experiment are described in detailed as follows:
\begin{description}

    \item[\textit{Bag of Words}] 
    To accelerate the computation of BoW features, the SURF \citep{bay2008speeded} rather than SIFT \citep{lowe2004distinctive} detector and descriptor were used in the experiment for detecting and describing the basic features for clustering. 
    K-means \citep{Lloyd1982,Kanungo2002} was used to create the visual words by clustering the SURF features built in the first stage. 
    Both low and high dimensional BoW feature vectors denoted as BoW-10 and BoW-100 were tested, which contained 10 and 100 visual words, respectively.

    \item[\textit{Histogram of Gradients}] 
    HoG can only work for detecting objects of the same size. 
    However, the sizes of the images in the EAD are varied. Therefore, to make use of HoG vector, the images in the dataset are first resize to a predefined size. 
    Two image sizes were considered in the experiment: $64 \times 128$ and $128 \times 128$ (denoted as HoG-64$\times$128 and HoG-128$\times$128), representing two possible shapes of Artcodes, rectangular and square, respectively. 

    \item[\textit{SOH Variants}]
    Considering the type of SOH feature variables and their importance for representing the Artcodes, five variants of the SOH feature vector from 5 to 12 dimensions (SOH-05, -07, -08, -10, -12), were evaluated by gradually adding new variable groups. 
    The SOH variants contain different subsets of the variables given in \tableautorefname\ref{tbl:SOHVariableList}. 
    SOH-05 contains the five basic variables to measure the two key properties (symmetry and smoothness) of the orientation histogram described earlier.
    SOH-07 adds two further measurements: $\mu(\mathcal{I})$ and $\sigma(\mathcal{I})$, which denote the mean and the standard deviation of the grayscale intensity of the input image. 
    SOH-08 adds one more variable $\chi^2(d\mathcal{C_L}, d\mathcal{C_R})$ to SOH-07, this new variable describes the similarity of the orientation histogram by comparing the derivatives of the left and right curves. 
    To study whether the intensity variation of the input image (particularly over the foreground regions) has an impact on the SOH, two another SOH variants, SOH-10 and SOH-12, which incorporate variables $S_9$, $S_{10}$, $S_{11}$ and $S_{12}$, are also used for experimental comparison (see \tableautorefname \ref{tbl:descriptionSOHVariants} for a detailed description of the SOH variants).
    
\end{description}

\subsubsection*{Classifier} 
\label{sec:svm_classifier}
In addition to feature descriptor and dataset, a detector or proposer also requires a classification method.
In this experiment, the SVM classification method was considered. 
Specifically, the SVM classifier that employs the Radial Basis Function (RBF) (i.e., Gaussian) kernel was adopted as the classification method.  
The RBF kernel on two feature vectors: $\vec{x}$ and $\vec{y}$, is defined as: 
\begin{equation}
K(\vec{x}, \vec{y}) = \exp \Big(- \frac{\parallel \vec{x} - \vec{y} \parallel ^ 2}{2\sigma^2} \Big)
\end{equation}
An equivalent definition involves $\gamma = \frac{1}{2\sigma^2}$:
\begin{equation}\label{eq:rbfKernel}
K(\vec{x},  \vec{y}) = \exp \big(- \gamma \parallel \vec{x} - \vec{y} \parallel ^ 2\big)
\end{equation}
where $\parallel \vec{x} - \vec{y} \parallel ^ 2$ is the squared Euclidean distance between the two feature vectors $\vec{x}$ and $\vec{y}$, $\gamma$ is a parameter that only affects the scale of $K(\vec{x}, \vec{y})$ values but does not change minimum and maximum $K(\vec{x}, \vec{y})$ values, meaning that it has no impact on the trained SVM classifier, which was therefore set to one in this experiment for computational convenience.

\subsubsection{Experimental Results}
\label{subsec:experiment-1Results}
Object proposal generation aims to find all possible presences of an object as fast as possible, even at the cost of a high number of  {\it false positives} --- incorrectly predicting non-Artcodes as Artcodes. 
As a result, a further detection method would be necessary --- one that is highly accurate, can only deal with these proposals and would not miss the objects. 
Consequently, in terms of the two metrics for evaluating object detection proposal methods presented in papers by \citet{uijlings2013selective,zitnick2014edge,hosang2016makes}, the methods that are unable to obtain better values in recall and efficiency could be considered as ``failing'' on object detection proposal task, even though they can obtain good values in other measurements, e.g., precision and accuracy. 
Among the evaluation metrics, recall and efficiency are the most two important metrics for evaluating the performance of Artcode proposal generation. Along with recall and efficiency, other measurements described in \sectionautorefname \ref{subsec:performanceMetrics} were also used in this experimental evaluation for aims of comparison.

This experiment was implemented using MATLAB and run on a consumer-grade laptop. 
Its results are reported in \tableautorefname  \ref{tbl:comparisonExperimentalResults}. 
As presented in \tableautorefname \ref{tbl:comparisonExperimentalResults}, the best result obtained in each performance measure is boldfaced; and the first three best results are marked by asterisks ($\ast$). 
The values of the six performance measurements: accuracy, precision, recall, TNR, F2 measure, MCC, and the computational cost measurement --- efficiency or elapsed time --- are presented in \tableautorefname \ref{tbl:comparisonExperimentalResults} as well.

\begin{table}[t]
\caption{
    Experimental results (averaged over ten rounds of execution) on the Artcode detection task using feature descriptors: BoW, HoG, and SOH variants. 
    The best result obtained in each performance metric is {\bf boldfaced}; and the first three best results are marked by asterisks ($\ast$). 
    The experiment was evaluated using the TAD dataset.}
\label{tbl:comparisonExperimentalResults}
\small{
\begin{tabular}{rllllllp{2cm}}
\hline
Methods & Accuracy & Precision & Recall & TNR & F2 & MCC & Elapsed time (sec.)  \\ \hline
BoW-10 & 0.671579$^\ast$ & 0.572725 & 0.665333 & 0.675652 & 0.644402 & 0.334638 & 2460.323123 \\
BoW-100 & 0.705263$^\ast$ & 0.611765$^\ast$ & 0.693333 & 0.713043 $^\ast$ & 0.675325 & \textbf{0.399489}$^\ast$ & 2634.86228 \\ \hline
HoG-64x128 & \textbf{0.714737}$^\ast$ & \textbf{0.70206}$^\ast$ & 0.482667 & \textbf{0.866087}$^\ast$ & 0.514407 & 0.383506$^\ast$ & 245.903276 \\
HoG-128x128 & 0.703158$^\ast$ & 0.688081$^\ast$ & 0.453333 & \textbf{0.866087}$^\ast$ & 0.486334 & 0.355885$^\ast$ & 287.299745 \\ \hline
SOH-05 & 0.482105 & 0.420108 & 0.817333$^\ast$ & 0.263478 & 0.687018 & 0.093636 & \textbf{0.442411}$^\ast$ \\
SOH-07 & 0.547368 & 0.461529 & \textbf{0.869333}$^\ast$ & 0.337391 & \textbf{0.738419}$^\ast$ & 0.232054 & 0.448229$^\ast$ \\
SOH-08 & 0.540526 & 0.45229 & 0.784 & 0.381739 & 0.683404 & 0.175738 & 0.450316 \\
SOH-10 & 0.641579 & 0.531017 & 0.813333$^\ast$ & 0.529565 & 0.734614$^\ast$ & 0.343802 & 0.448416$^\ast$ \\
SOH-12 & 0.654737 & 0.544111 & 0.772 & 0.578261 & 0.712252$^\ast$ & 0.345082 & 0.468994 \\ \hline
\end{tabular}
}
\end{table}

\paragraph{BoW versus HoG}
In terms of the two aggregated evaluation metrics, accuracy and MCC,  BoW and HoG, achieve very close performance in Artcode proposal detection. 
BoW gives relatively balanced performance in both positive (Artcode) and negative (non-Artcode) examples compared with HoG, with slight differences between precision and TNR. 
However, HoG is an order of magnitude faster than BoW, showing that BoW is less appropriate for real-time applications than HoG. 
The expensive computational cost of BoW is mainly because of the large number of features required to calculate and the large computational cost to compute a bag of visual words. 
HoG obtains the highest value of TNR, which is significantly higher than that of SOH variants, showing that HoG is better in predicting non-Artcode objects. 
However, HoG achieves very low recall, which measures the fraction of the detected positives over the total amount of true positives. 
Thus, the relatively high value of precision obtaibed by the HoG is mainly due to the higher TNR value. 
Although HoG is significantly faster than BoW, neither are fast enough to satisfy the computational speed requirements for Artcode propose generation.
Additionally, they obtain relatively lower recall values but comparatively higher values in other evaluation metrics, indicating that they are not appropriate for object detection proposal tasks. 
In fact, they are designed for object detection rather than object proposal detection \citep{csurka2004visual, dalal2005histograms}.

\paragraph{SOH versus BoG and HoG}
Compared with BoW and HoG, the SOH variants obtained better values in recall, F2 measure and elapsed time, but lower values in accuracy, precision, TNR and MCC. 
The SOH family show opposite trends with HoG on detected positives and negatives, with better performance in positives than negatives. 
The best values of recall and F2 measure are obtained by SOH-07. 
Because of the low dimensions and simple calculation of each variable, SOH variants are extremely fast --- two orders of magnitude faster than HoW and three orders of magnitude faster than BoW. 
This is an appealing characteristic for a proposal generator. 
According to \citet{hosang2016makes,zitnick2014edge,uijlings2013selective}, recall and running speed are two critical measurements for an object proposal generator. 
Therefore, in the context of object detection proposal generation, SOH significantly outperforms both BoW and HoG.

Overall, the SOH variants obtain similar performance in recall, F2 measure, and elapsed time. 
All SOH variants are low-dimensional, with little impact on the computational cost when more variables are added into the SOH feature vector. 
With or without auxiliary variables such as the mean ($S_9$) and the standard deviation ($S_{10}$) of image gray intensity, SOH performs well in terms of recall and F2 measure. 
Because F2 measure pays twice importance over recall than precision, it would also be an evaluation indicator for the object detection proposal method. 
As the variables are added in, SOH starts to obtain better performance in negative examples while preserving its good performance in positive examples (Artcodes), which is reflected by the higher accuracy and TNR, but lower recall of SOH-10 and SOH-12 than the low-dimensional SOH variants (see the SOH family section (4th row) in \tableautorefname \ref{tbl:comparisonExperimentalResults}).  
For Artcode proposal generation task, considering its stringent demands on recall and efficiency; SOH-07 obtains higher recall and F2 measure than other SOH variants, and fast execution time (only slightly slower than SOH-05), hence SOH-07 is the recommended one for this purpose.

\subsection{Experiment 2: In-Depth Evaluation of the SOH Descriptor}
\label{sec:experiment2}
The first experiment compared the SOH variants with other two classic feature vectors for Artcode proposal generation on the EAD dataset, showing that the SOH achieves very good performance overall in this context. 
Among the SOH variants, the SOH-07 is best overall in terms of balancing performance and efficiency, and was therefore selected as the feature vector for Artcode proposal generation in this experiment.
We in this section present further experimental study on the Artcode detection proposer based on the SOH-07 feature vector, using random forests or SVM classifiers. 
This Artcode proposal generation system will hereafter be referred to as {\sc ArtcodePresence}.
Three sub-experiments were conducted in this experiment to comprehensively evaluate the performance of this system with different configurations.

\subsubsection{Classifier Selection}
\label{subsubsec:classifier}
Having decided on a SOH-07 configuration, there is now a need to select a suitable classifier. Since the Artcode proposal generator and the decoding system will be deployed through a mobile platform, speed and memory are key considerations when choosing an appropriate classifier. 
Although recently almost all of the state of the art of computer vision tasks (e.g., CoAtNets \citep{dai2021coatnet} for classification, YOLOv5 \citep{jocher2021} for object detection, and Mask R-CNN \citep{he2017mask} for semantic segmentation) were obtained by deep neural networks, they are unsuitable for this task because 
1) this experiment aims to evaluate the performance of the SOH feature vector on proposing Artcodes rather than to achieve its state of the art.
2) the datasets for training are relatively small and unable to train a proper neural network model; and  
3) this work studies a proposed feature descriptor but deep neural networks require no feature engineering;
Therefore, we focus on selecting classifiers from traditional statistical machine learning models.

Random forests \citep{breiman2001random}, as an ensemble learning method, can provide accurate prediction for classification, without much overfitting, and they are relatively easy to train. 
More importantly, the random forests technique is computationally efficient when running, and would be sufficient for real-time application. 
Therefore, random forests were chosen as the {\em baseline} classifier. 
Additionally, due to its superior computational and classification performance, SVM was also used an alternative to random forests.

Random forests were proposed by \citet{breiman2001random}, as an ensemble of decision trees such that each tree depends on the values of a random vector sampled independently, with the same distribution for all trees in the forest. 
Random forests have desirable characteristics for the learning outcomes in this study: they are as accurate as Adaboost \citep{freund1995desicion}; robust to outliers and noise; faster than bagging \citep{breiman1996bagging} and boosting \citep{freund1995desicion}; and they do not overfit. 
Moreover, they have intuitive and easily tunable parameters --- only the number of decision trees needs to be considered. 
These properties indicate that random forests are desirable as the baseline classification method, demonstrating the problem of Artcode proposal generation.

\subsubsection{SMOTE-augmented True Artcode Datasets}
\label{subsec:experiment2Dataset}
The TAD dataset, as shown in \sectionautorefname \ref{subsec:dataset_tad}, mainly contains images captured during real interaction with Artcode artefacts  --- the real-life Artcode images. 
TAD is the base dataset for Artcode detection proposal evaluation, but it is small and unbalanced. 
In order to achieve good performance, a dataset should introduce diversity, including positive and negative samples, which cover a wide range of variation types. 
Although work is ongoing to extend the TAD dataset by Artcode designers, the current number of available Artcodes is limited (or at least less than the number of available non-Artcodes);  the imbalance of the TAD dataset will be expected to exist in the near future. 

It is well known that learning from an imbalanced dataset usually produces {\it biased} classifiers that have higher predictive accuracy over the {\it majority} class, but poorer predictive accuracy over the {\it minority} (rare) class \citep{seiffert2010rusboost}; an imbalance handling technique was therefore needed to deal with the imbalance of the TAD dataset.

The EAD dataset alleviates the imbalance of the TAD dataset by adding ``fake`` Artcode images (as shown in \figureautorefname\ref{fig:additionalDataset}), which in this case are images of printouts of true Artcode images taken under various illuminations and camera poses. 
However, this augmentation technique operates in image space (i.e., the raw data space) rather than feature space; and hence, the resultant EAD dataset is appropriate for comparison studies between different feature vectors. 
This experiment evaluates in detail the performance of the SOH-07 feature as part of a classifier, operating in the feature space. 
Therefore, the imbalance handling technique that operates on feature space is more appropriate than the augmentation techniques acting in image space, and was adopted to deal with imbalance of the TAD dataset.

\subsubsection*{Dealing with Imbalance via SMOTE}
\label{subsec:handlingImbalance}
A number of techniques \citep{chawla2002smote, han2005borderline, seiffert2010rusboost, chawla2003smoteboost, elkan2001foundations, fan1999adacost} have been proposed to handle imbalanced classification. 
These techniques are mainly categorised into two groups: methods at data level or methods at algorithmic level. 
The methods at data level \citep{chawla2002smote, han2005borderline, seiffert2010rusboost, chawla2003smoteboost} attempt to balance distributions by examining the representative proportions of class examples in the dataset, while the methods at algorithmic level \citep{elkan2001foundations, fan1999adacost} consider the costs associated with misclassifying examples, also known as {\it cost-sensitive} learning. 
The section describes the methods operating at data level; a more comprehensive survey on investigating learning from imbalanced data can be found in works, such as that by \citet{he2009learning}.

One simple and effective imbalance handling technique is random resampling: either random oversampling or undersampling. 
The former randomly adds samples from the {\em minority} class and augments the original dataset, whereas the latter achieves a balanced distribution by randomly removing samples from the {\em majority} class. 
Random oversampling simply appends replicated data to the original dataset, easily leading to overfitting (the situation that a classifier works perfectly in the training data but works badly for unseen data). 
With regard to avoiding overfitting, \citet{chawla2002smote} proposed a synthetic sampling method --- the Synthetic Minority Oversampling TEchnique (SMOTE), where the minority class is oversampled by creating ``synthetic'' examples rather than simply oversampling with replacement. 
SMOTE generates synthetic examples in a less application-specific manner, by working in ``feature space'' rather than ``data space''. 
The minority class is oversampled by taking each minority class sample and introducing synthetic examples along the line segments connecting any/all of $k$ nearest neighbours of the minority class. 
The synthetic examples $\mathcal{F}$ at the current example $\mathcal{F}_{curr}$ are generated by the following formula: 
\begin{equation}\label{eq:smote}
    \mathcal{F} = \bigg\{ \mathcal{F}_{new}^{i} \mid i = 1, \dots, \floor*{ N/100 } \bigg\}
\end{equation}
with the $\mathcal{F}_{new}^i$ defined as follows:
\begin{equation}
    \mathcal{F}_{new}^i = \mathcal{F}_{curr} + r \cdot  (\mathcal{F}_{curr} - \mathcal{F}_k), \; k = 1, \dots, K
\end{equation}
where $\mathcal{F}$ is the new set of synthetic feature vectors (examples) at the current example $\mathcal{F}_{curr}$, $\mathcal{F}_k$ denotes the $k$th neighbour of the feature vector $\mathcal{F}_{curr}$ under consideration, $r$ is a random number between 0 and 1, and $N$ is a parameter determining the total number of oversampling examples generated by SMOTE. 
The number of neighbours that are randomly chosen from the $K$ nearest neighbors is based on the required amount of synthetic examples ($(N/100) \times T$, where $T$ is the number of minority class samples).

Over the past few years, SMOTE has had a great deal of success in various applications, and a number of variants have been proposed. \citet{han2005borderline} presented Borderline-SMOTE, a synthetic sampling techniques built on SMOTE that only oversamples or strengthens the {\em borderline} minority examples, which is more easily misclassified than those far from the borderline.  
\citet{chawla2003smoteboost} proposed SMOTEBoost, which introduces SMOTE data sampling into the Adaboost \citep{freund1995desicion} algorithm, showing improved prediction performance on the minority class and overall $F_\beta$ measure.  
\citet{seiffert2010rusboost} proposed RUSBoost, which applies random undersampling in the Adaboost algorithm. 
RUSBoost gives a comparable performance to SMOTEBoost, but is simpler and faster. 
However, the improved performances of these SMOTE variants is very small. 
Considering the simplicity of SMOTE and its verified performance levels across various applications, SMOTE was employed in this work to deal with the imbalance of the Artcodes dataset. 

\begin{table}
\centering
\caption{Example SOH (SOH-07) feature vector from the SaTAD-4 dataset.}
\label{tbl:quantitativeDataset}
\small{
    \begin{tabular}{@{}lllllllc@{}}
    \hline
    \multicolumn{7}{c}{Variables/Features}                                                                                                                    & Class/Label     \\ \hline
    \multicolumn{1}{c}{$S_1$} & \multicolumn{1}{c}{$S_2$} & \multicolumn{1}{c}{$S_3$} & \multicolumn{1}{c}{$S_9$} & \multicolumn{1}{c}{$S_{10}$} & \multicolumn{1}{c}{$S_6$} & \multicolumn{1}{c}{$S_7$} & \multicolumn{1}{c}{1/0} \\ \hline
    .021247                & .086649                & .385715                & 223.070675             & 51.21163               & .019988                & .011607                & 0                        \\
    .050593                & .003012                & .081212                & 164.618898             & 51.93487               & .005639                & .003802                & 1                        \\
    .14202                 & .015217                & .10068                 & 122.87                 & 53.933                 & .018584                & .010488                & 0                        \\
    .052458                & .099058                & .007762                & 145.535080             & 25.685932              & .012161                & .006863                & 1                        \\
    .015428                & .001412                & .03387                 & 193.38                 & 80.952                 & .008229                & .004432                & 0                        \\ \hline
    \end{tabular}
}
\end{table}

\begin{table}[t]
\centering
\caption{Numbers of positive (Artcode) and negative (non-Artcode) examples in the TAD and SaTAD daatsets.}
\label{tbl:smotedDatasets}
\small{
    \begin{tabular}{@{}llllll@{}}
    \hline
    \textbf{} & \multicolumn{5}{c}{Dataset} \\ \cline{2-6} 
     & TAD & SaTAD-1 & SaTAD-2 & SaTAD-3 & SaTAD-4 \\ \hline
    Number of Artcodes & 47 & 94 & 94 & 94 & 141 \\
    Number of non-Artcodes & 116 & 70 & 94 & 116 & 147 \\ \hline
    \end{tabular}
}
\end{table}

\begin{figure}[!t]
    \centering
     \begin{subfigure}[b]{0.45\textwidth}
        \includegraphics[width=\textwidth]{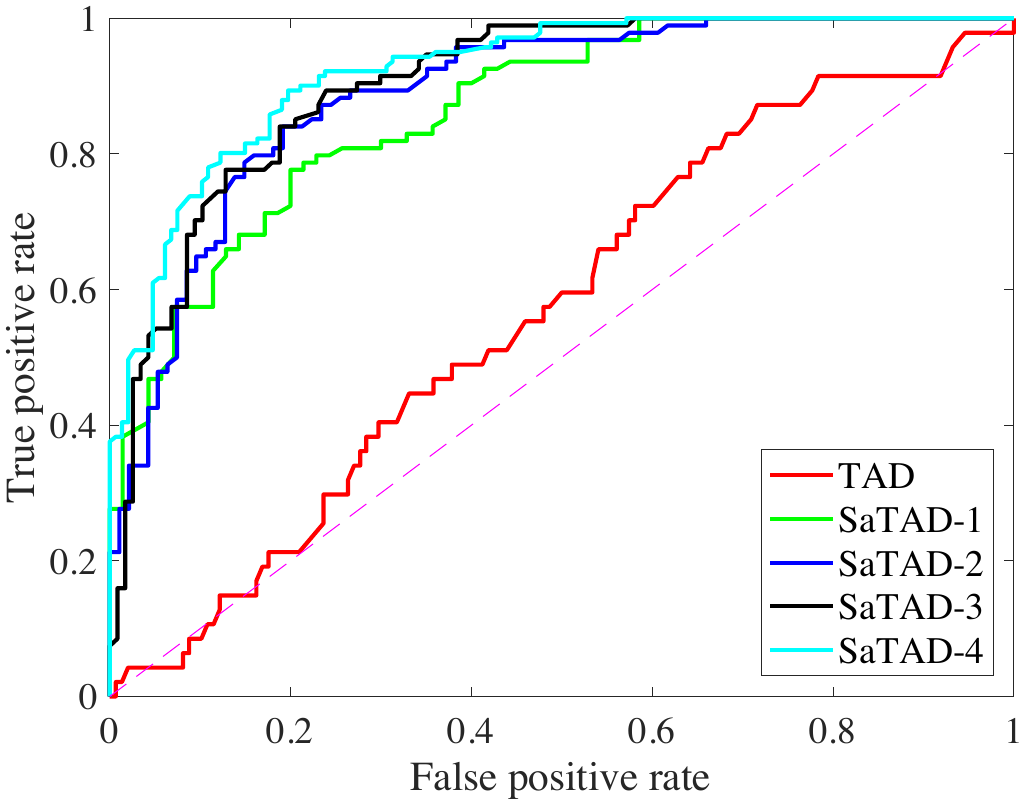}
        \caption{ROC curves}
        \label{fig:compROC}
    \end{subfigure}   
     \begin{subfigure}[b]{0.45\textwidth}
        \includegraphics[width=\textwidth]{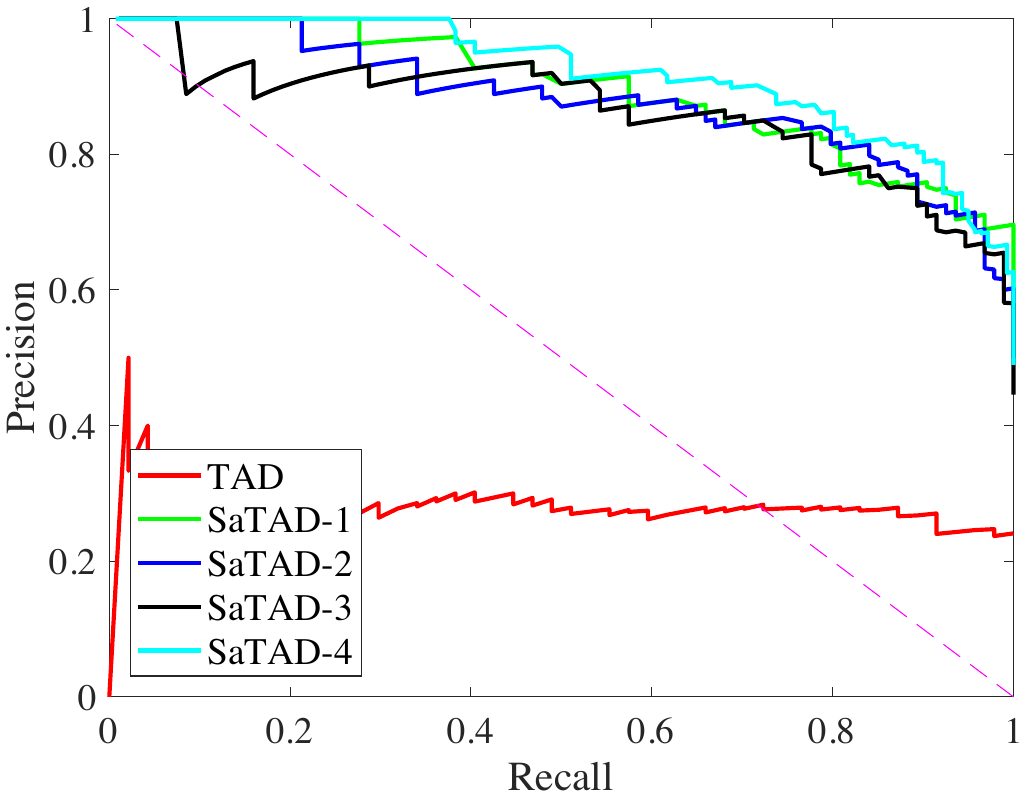}
        \caption{PR curves}
        \label{fig:compPR}
    \end{subfigure}
     \begin{subfigure}[b]{0.45\textwidth}
        \includegraphics[width=\textwidth]{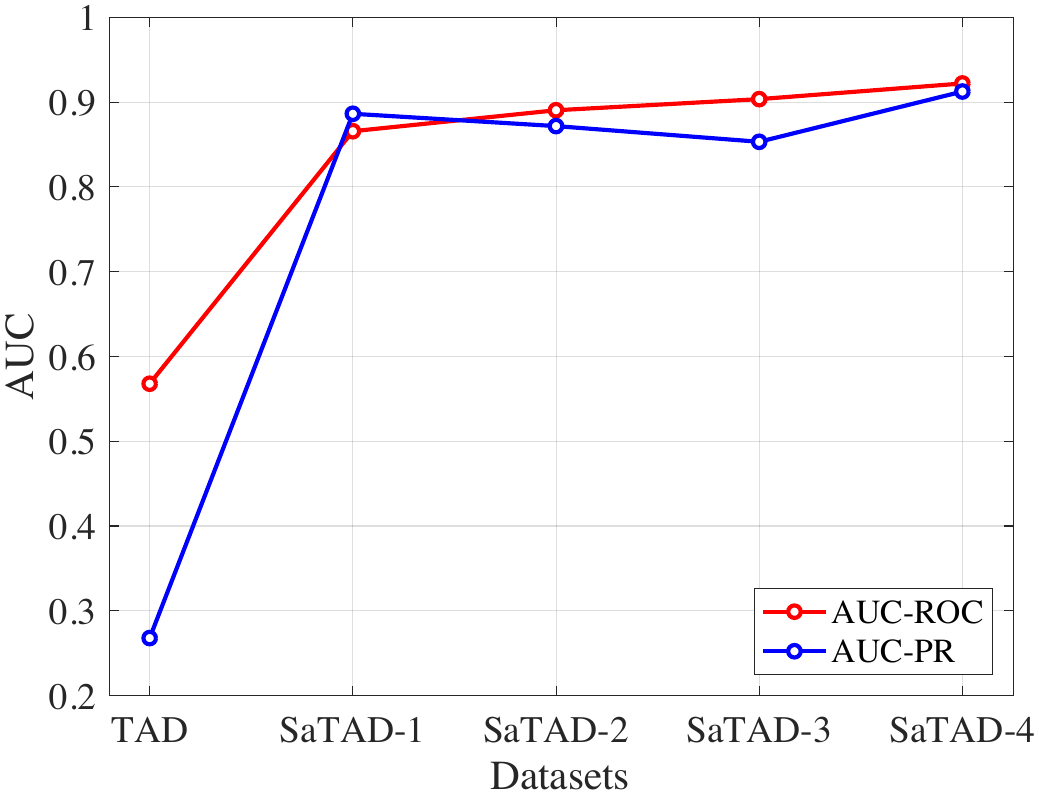}
        \caption{Area under the curve (AUC)}
        \label{fig:compAUC}
    \end{subfigure} 
    \begin{subfigure}[b]{0.45\textwidth}
        \includegraphics[width=\textwidth]{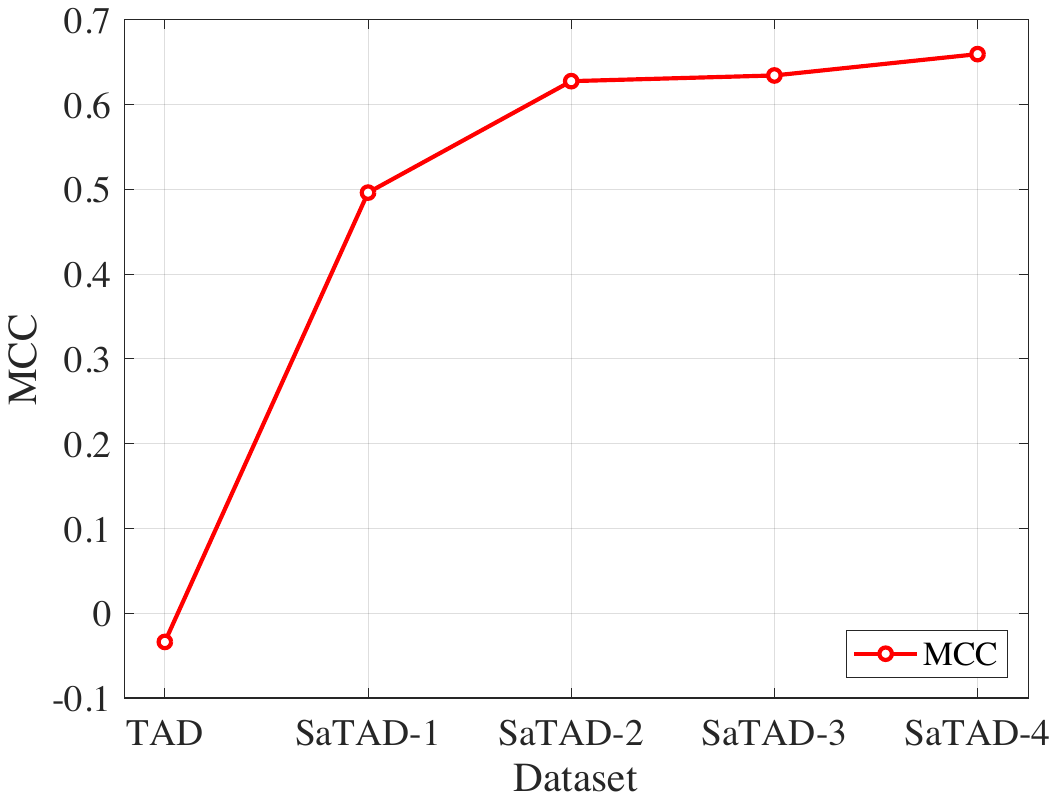}
        \caption{MCC}
        \label{fig:compMCC}
    \end{subfigure} 
    \caption{
        Performance comparison among the random forests classifiers ({\tt nTrees} = 80) trained these datasets: TAD, SaTAD-1, SaTAD-2, SaTAD-3, and SaTAD-4. (\subref{fig:compROC}) and (\subref{fig:compPR}) present the ROC and PR curves of the five classifiers. 
        The classifier trained on SaTAD-4 dataset achieves the best performance in terms of ROC and PR curves; its ROC and PR curves are closer to the corners of $(0, 1)$ and $(1, 0)$, respectively. 
        (\subref{fig:compAUC}) presents the area under the curve (AUC) of the five classifiers. 
        As a higher value of AUC indicates better performance; the classifier trained on SaTAD-4 obtains the best performance. 
        The values of MCC of the five classifiers are shown in (\subref{fig:compMCC}): the classifier trained on SaTAD-4 also obtains the higher value, indicating its best prediction performance.
    }
    \label{fig:compDataset}
\end{figure}

\subsubsection{Selection of the ``Optimal'' SMOTE-augmented TAD}
SMOTE is unable to be used directly in an Artcode dataset that only contains images. 
Due to the proposed SOH feature vector, each image in the Artcode dataset needs to be converted into a quantitative feature vector (\tableautorefname \ref{tbl:SOHVariableList}), producing a {\it quantitative} dataset with the same amount of samples as the original dataset. 
In this dataset, for computation convenience, Artcode and non-Artcode classes are assigned the integers 1 and 0, denoting Artcode and non-Artcode class labels, respectively, as shown in \tableautorefname\ref{tbl:quantitativeDataset}.

To obtain an ``optimal'' SMOTE-augmented TAD dataset contributing to Artcode detection proposal performance, four SMOTE-augmented True Artcode Datasets (SaTADs) (SaTAD-1, SaTAD-2, SaTAD-3, and SaTAD-4) (\tableautorefname\ref{tbl:smotedDatasets}) with varied proportion of Artcodes and non-Artcodes were generated using SMOTE. 
The random forests classifiers were trained on both the TAD and the SMOTE-augmented TAD, denoted as the {\it Original} classifier, and the {\it SMOTE-augmented} (SaTAD-1 -- SaTAD-4) classifier. 
The performance of these classifiers was then evaluated based on their own datasets using ROC and PR curves (\sectionautorefname \ref{subsec:performanceMetrics}) and their respective Area Under Curve (AUC). 
Because MCC is an informative measure for both balanced and imbalanced datasets, it is also adopted for evaluation.

As illustrated in \figureautorefname \ref{fig:compDataset}, the SMOTE-augmented classifiers significantly outperform the Original classifier in terms of ROC, PR curves, AUC-ROC, AUC-PR and MCC, showing that the SMOTE-augmented dataset significantly enhanced the classifier's performance. 
Among the four SaTAD classifiers, the SaTAD-4 classifier achieved the best performance in terms of all four measures --- the improved performance was not significantly greater than SaTAD2--3 but was still consistently better. 
Therefore, the SaTAD-4 classifier and SaTAD-4 dataset were adopted for further experimental evaluation of the proposed method.

\subsubsection*{SaTAD-4 as a Basis for Further Development}
\label{par:smote_sd4}
This SMOTE-augmented classifier trained on the SaTAD-4 dataset is the basis of the {\sc ArtcodePresence} system in this work.
The non-SMOTE-augmented classifier --- the Original classifier --- can be further enhanced via the exploration of Metamorphic Relations \citep{chen1998metamorphic} for post-training fine turning, a technique from software engineering, as reported in \citet{xu2021using}.

\subsubsection{Investigation of {\sc ArtcodePresence} using Random Forests and SaTAD-4}
\label{subsec:experimentalSetting}
This experiment investigates the performance of the {\sc ArtcodePresence} (i.e., Artcode proposal generator) with the selected classifer --- random forests --- and training dataset --- SaTAD-4.

\subsubsection*{Experimental Setting} 
This {\sc ArtcodePresence} is configured to use random forests classifier and trains on the SaTAD-4 dataset, which was implemented using MATLAB and run on a same consumer-grade laptop as the previous experiment.
Its performance was evaluated using $k$-fold cross-validation on the SaTAD-4 dataset. 
Because random forests are used in the classifier, the performance exhibits a certain level of variation on the execution due to the random variable (feature) selection from the feature vector. 
Therefore, this procedure was executed 50 times, and its performance was evaluated based on the its descriptive statistics using notched boxplots of quantitative performance measures. Experiments were then conducted to study the impact of the tuning parameter of random forests --- the number of trees ({\tt nTrees}) in the forests --- on the performance of {\sc ArtcodePresence}.

\begin{figure}[!t]
    \centering
    \begin{subfigure}[t]{0.325\textwidth}
        \includegraphics[width=\textwidth]{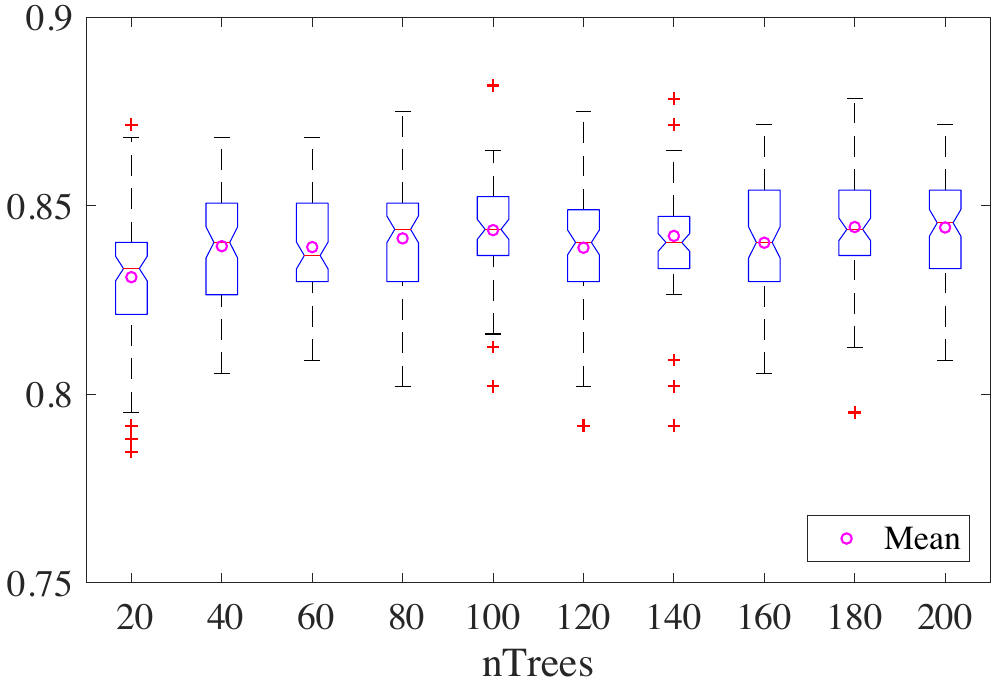}
        \caption{Accuracy}
        \label{fig:performanceMetrics_acc}
    \end{subfigure}   
    \begin{subfigure}[t]{0.325\textwidth}
        \includegraphics[width=\textwidth]{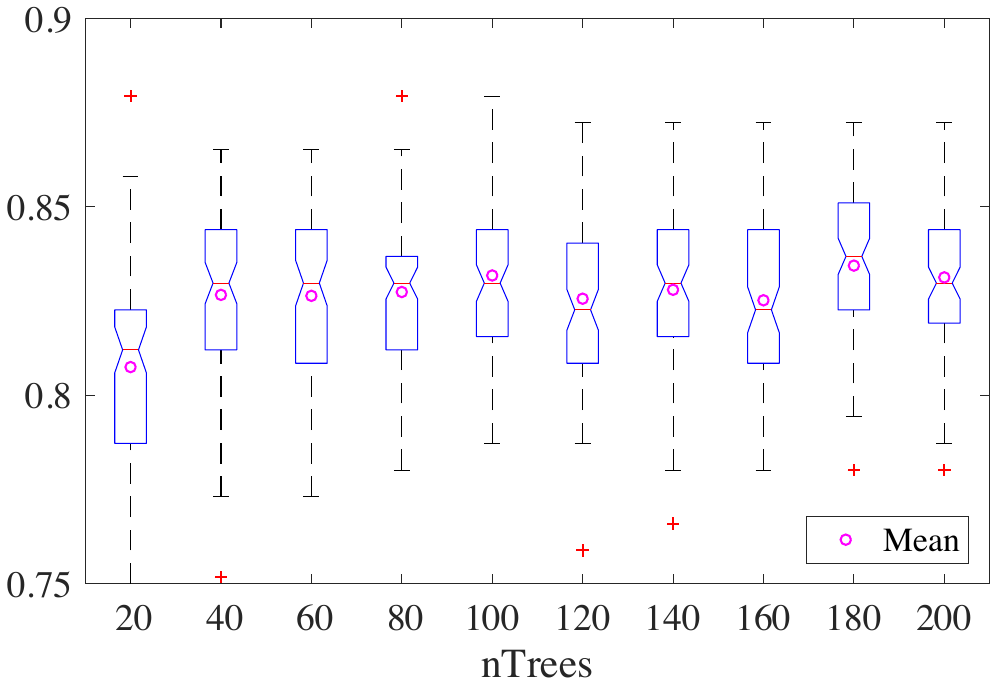}
        \caption{Recall}
        \label{fig:performanceMetrics_rec}
    \end{subfigure} 
    \begin{subfigure}[t]{0.325\textwidth}
        \includegraphics[width=\textwidth]{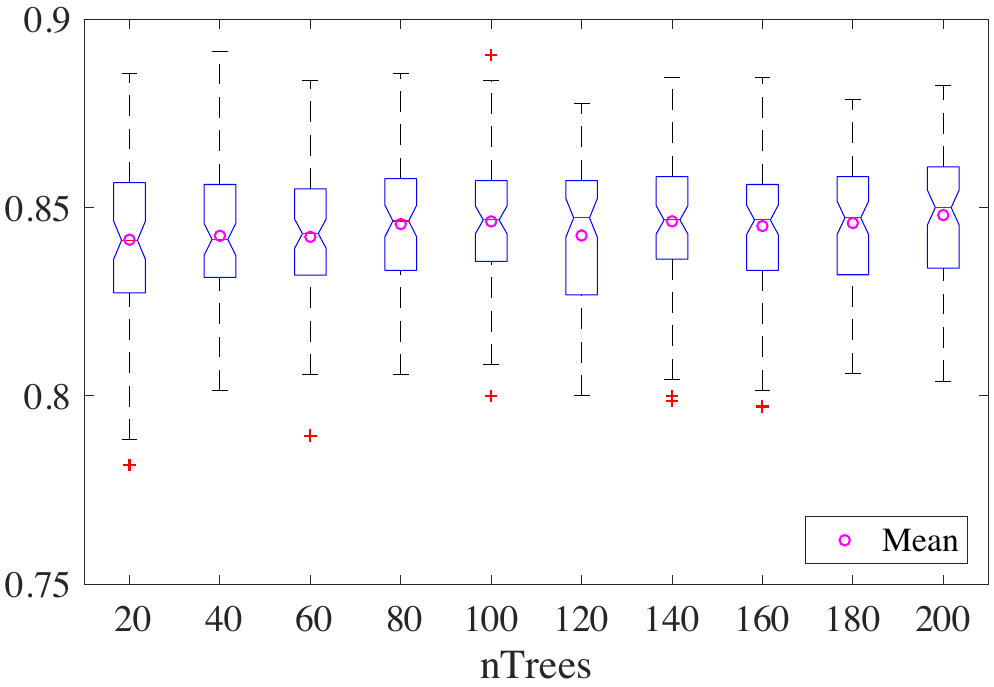}
        \caption{Precision}
        \label{fig:performanceMetrics_prec}
    \end{subfigure} 
    \begin{subfigure}[t]{0.325\textwidth}
        \includegraphics[width=\textwidth]{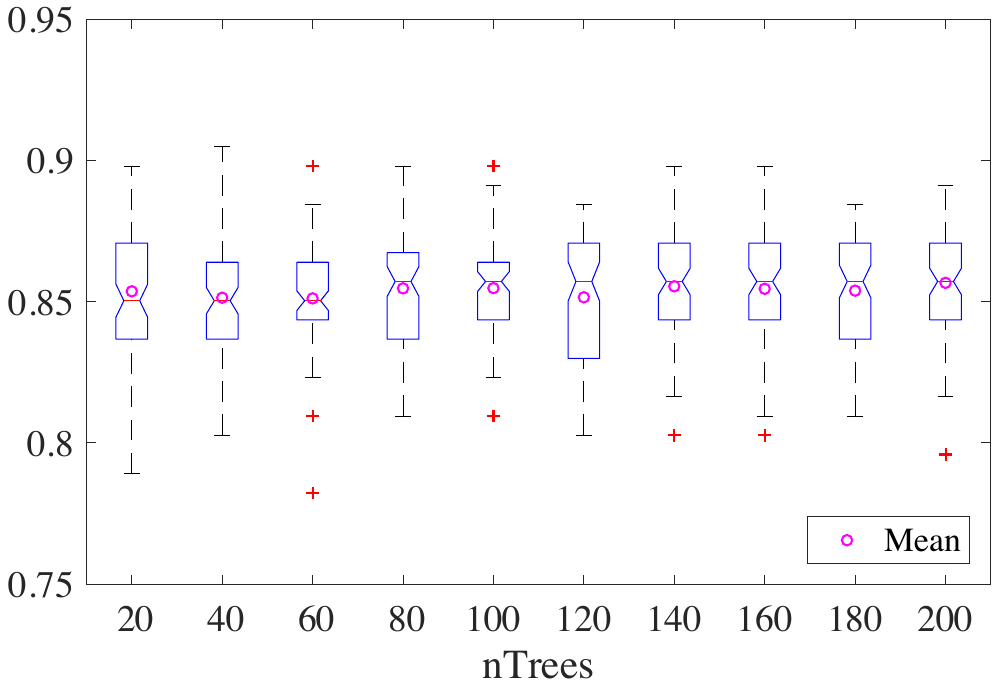}
        \caption{TNR}
        \label{fig:performanceMetrics_tnr}
    \end{subfigure}
    \begin{subfigure}[t]{0.325\textwidth}
        \includegraphics[width=\textwidth]{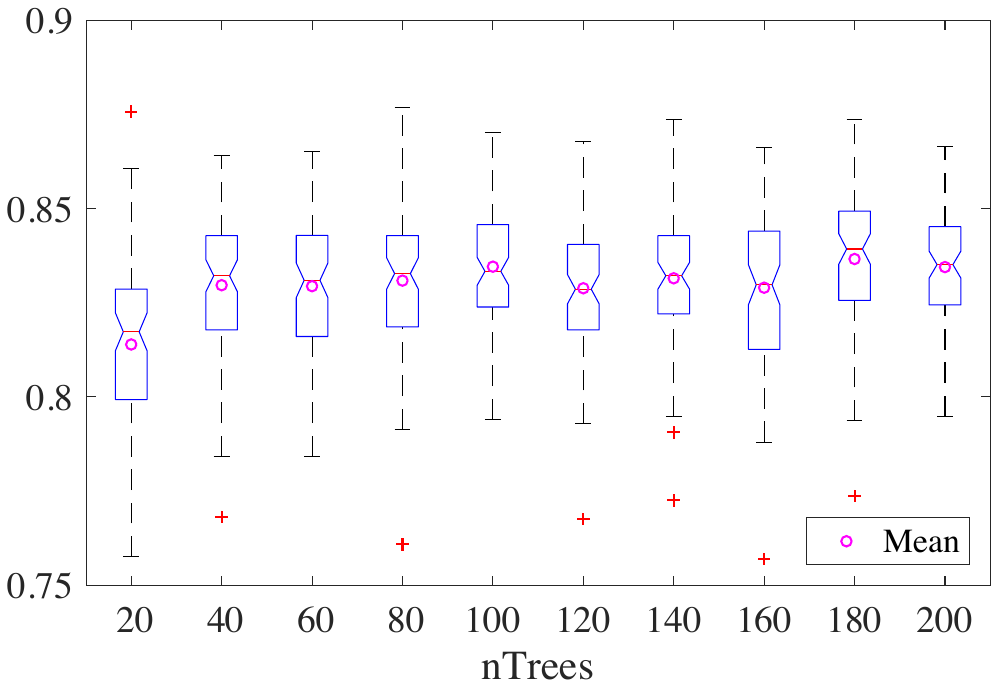}
        \caption{F2 measure}
        \label{fig:performanceMetrics_fbeta}
    \end{subfigure} 
    \begin{subfigure}[t]{0.325\textwidth}
        \includegraphics[width=\textwidth]{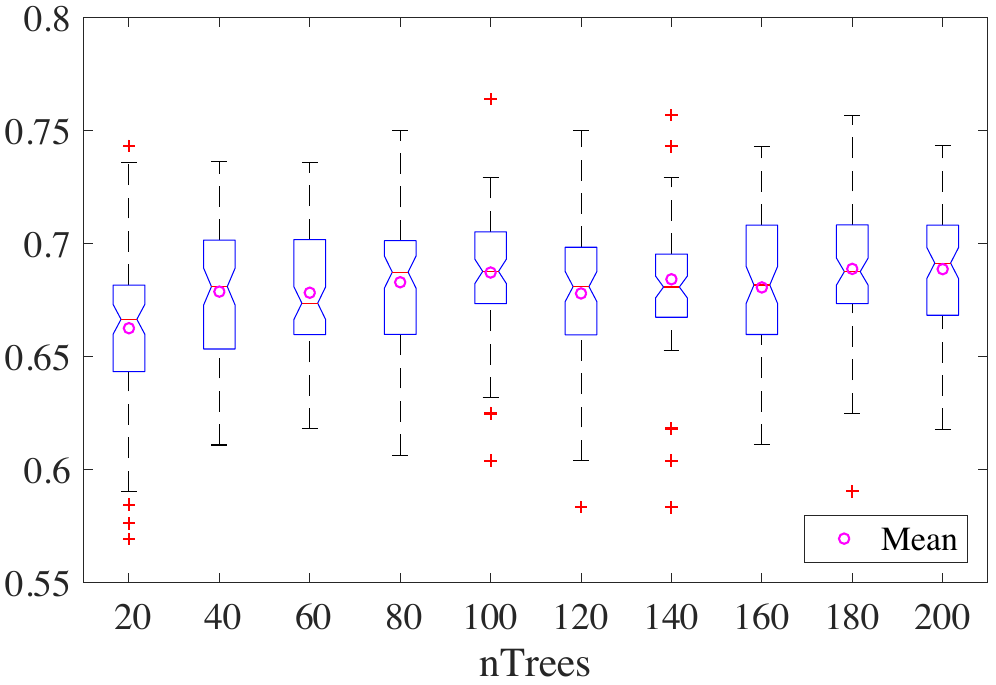}
        \caption{MCC}
        \label{fig:performanceMetrics_mcc}
    \end{subfigure}  
    \caption{
        Performance of the proposed classifier evaluated on the SaTAD-4 dataset. 
        The performance measures are denoted by the notched box plots. 
        The boxes show the interquartile range (IQR) (the 25th to 75th percentile). 
        The small horizontal lines at the middle of the boxes indicate the media of the data. 
        The notches display the conference intervals around the median. 
        The whiskers indicates the data that are lesser than the 75th percentile or greater than the 25th percentile. 
        The outliers are denoted (+), and the means of the data are denoted ($\circ$).
    }
    \label{fig:performanceMetrics}
\end{figure}

\subsubsection*{Experimental Results}
\label{sebsec:experimentalResults}
\begin{figure}[t]
    \centering
    \begin{subfigure}[b]{0.45\textwidth}
        \includegraphics[width=\textwidth]{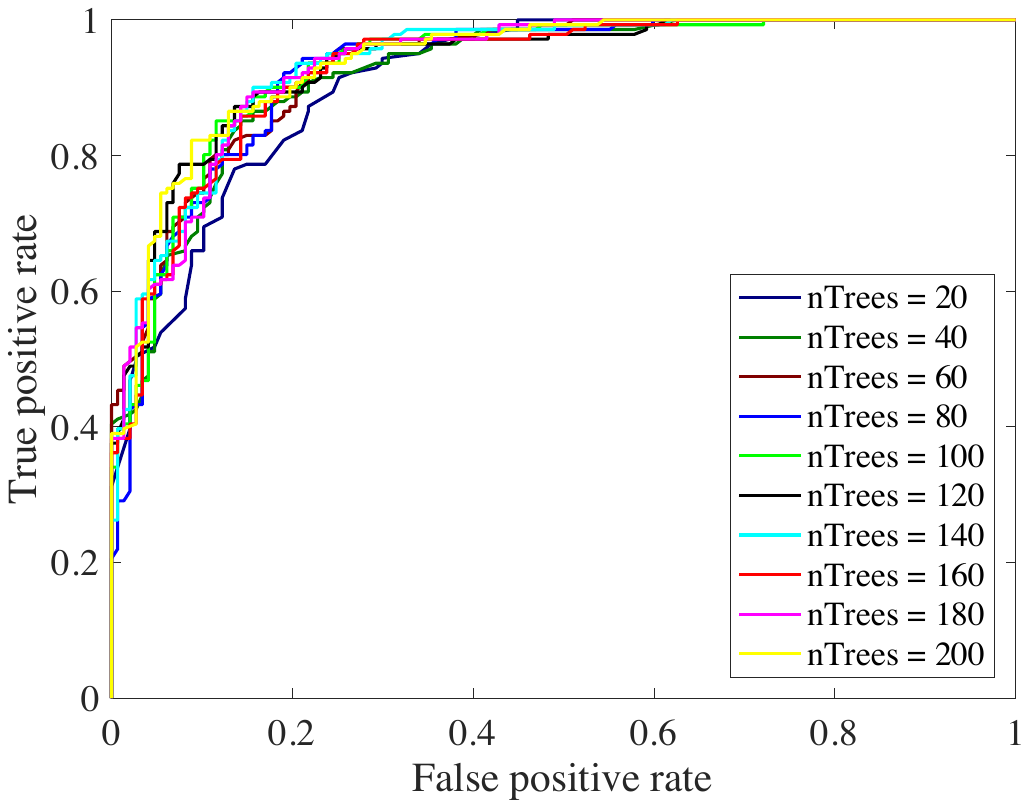}
        \caption{ROC curves}
        \label{fig:pc_roc}
    \end{subfigure}   
    \begin{subfigure}[b]{0.45\textwidth}
        \includegraphics[width=\textwidth]{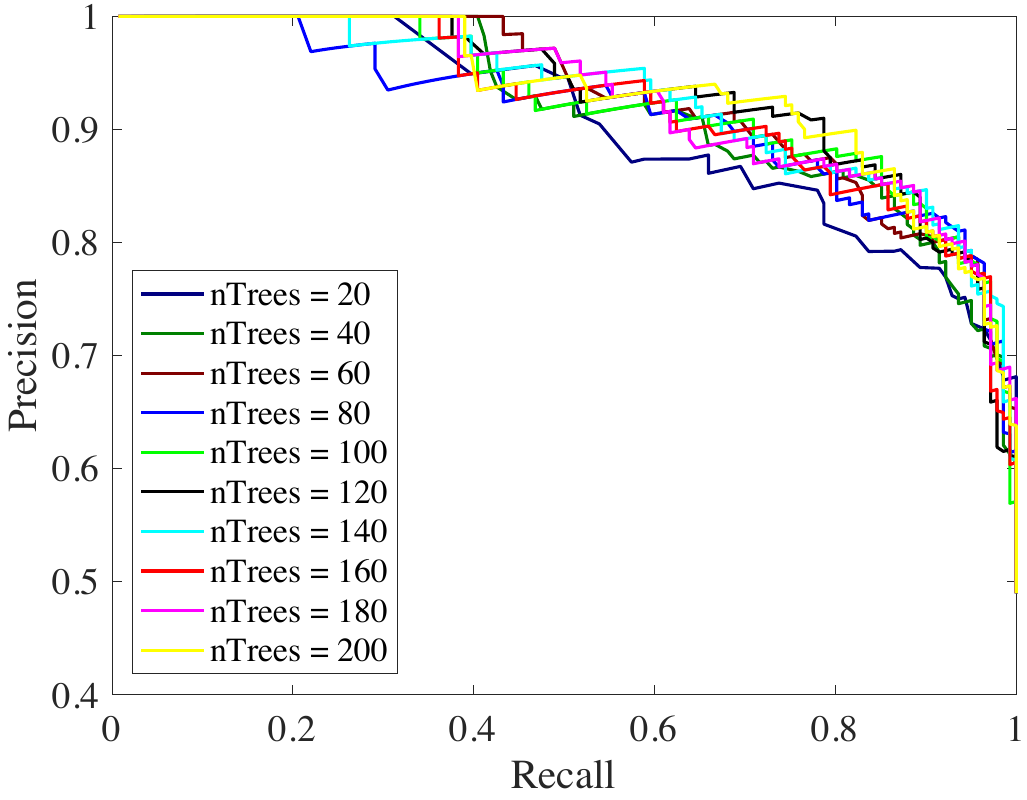}
        \caption{PR curves}
        \label{fig:pc_pr}
    \end{subfigure} 
    \begin{subfigure}[b]{0.60\textwidth}
        \includegraphics[width=\textwidth]{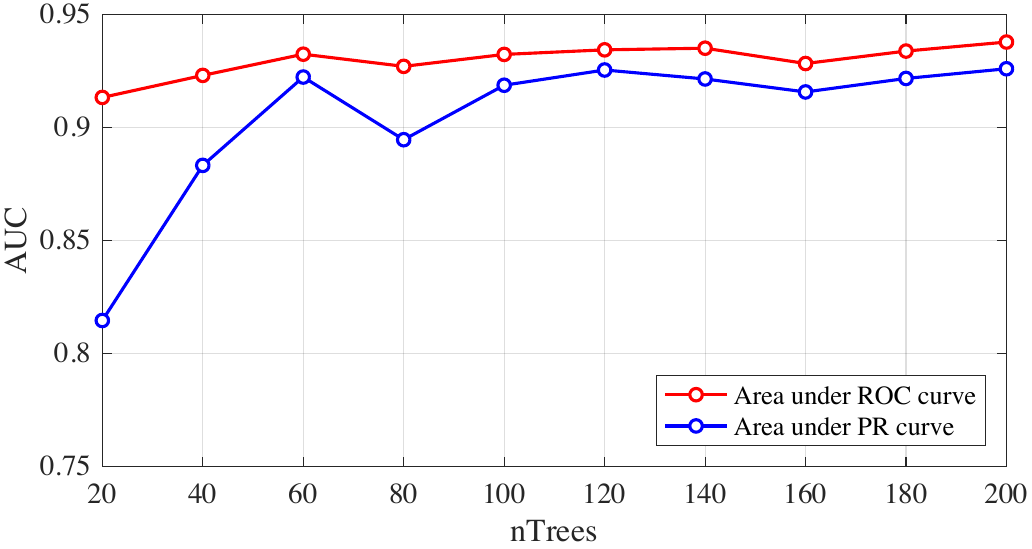}
        \caption{Area under ROC and PR curves}
        \label{fig:pc_auc}
    \end{subfigure} 
    \caption{
        Performance curves of the proposed classifier (detection proposer) with different values of {\tt nTrees}, evaluated on SaTAD-4 dataset. 
        (\subref{fig:pc_roc}) and (\subref{fig:pc_pr}) present the ROC and PR curves. 
        The areas under these curves (AUC) are shown in (\subref{fig:pc_auc}), which is a plot of \tableautorefname \ref{tbl:performanceAuc}.
    }
    \label{fig:performanceCurves}
\end{figure}

\figuresautorefname \ref{fig:performanceMetrics} and \ref{fig:performanceCurves} present the classifier's performance with different values of {\tt nTrees} (the number of decision trees in the random forests). 
\figureautorefname \ref{fig:performanceMetrics} presents the six performance metrics generated from the 50 rounds of running cross-validation of the classifiers. 
The figure also shows the average values (pink circles in the boxplots) of these performance metrics, also calculated from 50 rounds of running.

The impact of {\tt nTrees} on the classifier's performance is illustrated in \figuresautorefname \ref{fig:performanceMetrics} and \ref{fig:performanceCurves}, showing that the classifier stabilizes at approximately {\tt nTrees} $= 40$, in terms of the evaluation metrics, with small variations in the performance when {\tt nTrees} ranges from 40 to 200. 
This is reflected by whether the notches between the classifiers with different values of {\tt nTrees} overlap or not. 
As shown in \figuresautorefname\ref{fig:performanceMetrics_acc}, \ref{fig:performanceMetrics_fbeta}, and \ref{fig:performanceMetrics_mcc}, the notch of the first box ({\tt nTrees} $= 20$) does not overlap with the notches of other boxes for these three overall performance metrics (accuracy, F2 measure, and MCC). 
According to \citet{chambers2018graphical}, if two boxes' notches do not overlap, then there is a strong evidence (95\% confidence) that they have different medians. 
The performance of other classifiers are evidently better than the classifier with {\tt nTrees} $= 20$. 
Additionally, the notches of these boxes ({\tt nTrees} $= 40, \dots,200$) all overlap, indicating that their medians are not significantly different. 
This is also shown in \figureautorefname\ref{fig:pc_auc} by the sharp increase of PR-AUC when the value of {\tt nTrees} is increased from 20 to 40. 
The following analysis focuses on the performance evaluation of the classifiers with {\tt nTrees} $\ge 40$.

\begin{table}[!t]
\centering
\caption{
    Area under the ROC and PR curves of the classifier in \figureautorefname \ref{fig:performanceCurves}. 
    The best values of AUC of ROC and PR curves are boldfaced.}
\label{tbl:performanceAuc}
\small{
    \begin{tabular}{@{}ccccccccccccc@{}}
    \hline
     & \multicolumn{10}{c}{Number of decision trees (${\tt nTrees}$)} &  &  \\ \cline{2-11} 
    AUC & 20 & 40 & 60 & 80 & 100 & 120 & 140 & 160 & 180 & 200 & Mean & St Dev \\ \hline
    ROC & .9133 & .9231 & .9324 & .9271 & .9323 & {\bf.9344} & .9351 & .9283 & .9338 & .9378 & .9298 & .0072 \\ \hline
    PR & .8146 & .8832 & .9223 & .8946 & .9187 & .9254 & .9215 & .9157 & .9217 & {\bf .9260} & .9044 & .0346 \\ \hline
    \end{tabular}
}
\end{table}

Regarding the classifier's performance on predicting the positive class (Artcode), as measured by recall and precision, the classifier obtains scores of about 0.82 and 0.84 for recall and precision, respectively.
This means that the classifier correctly predicts 82\% of Artcodes from the given dataset and predicts an Artcode class to be a true Artcode with 84\% accuracy. 
Likewise, the classifier performs well in the negative class (non-Artcodes), as measured by TNR (\figureautorefname\ref{fig:performanceMetrics_tnr}), with 85\% of non-Artcodes correctly predicted. 
Using 1 and 0 to denote Artcode and non-Artcode class, respectively, and $\hat{y}$ as the predicted class of the true class $y$, the proposed classifier has the following conditional probabilities: 
\begin{align}
	recall = \Pr(\hat{y} = 1 \mid y = 1) = 0.82  \\
	precision = \Pr(y =1 \mid \hat{y} = 1) = 0.84  \\
	TNR = \Pr(\hat{y} = 0 \mid y = 0) = 0.85
\end{align}

The accuracy, F2 measure, and MCC assess the classifier's overall performance, considering both positive and negative classes. 
As shown in \figuresautorefname\ref{fig:performanceMetrics_acc} and \ref{fig:performanceMetrics_fbeta}, the accuracy and F2 measure of the classifiers are about 0.84 and 0.83, respectively, indicating that the classifier correctly predicts 84\% of samples in SMOTE-augmented Artcode dataset (SaTAD-4), and achieves a good tradeoff (0.83) between precision and robustness. 
Because the given dataset --- SaTAD-4 --- is almost balanced, accuracy is as an effective performance measure as is F2 measure: both show good overall performance of the chosen classifier. 
The MCC of the classifier on the given dataset is approximately 0.68, as a specific case of Pearson Correlation Coefficient (PCC) \citep{pearson1895note} of two binary variables, MCC has similar interpretation as the PCC \citep{pearson1895note}: MCC = 0.68 exhibits a ``strong'' positive agreement between {\it predictions} and {\it observations}, showing the good predictive power of the classifier.

The good predictive power of the classifier is also visually reflected by the ROC and PR curves. 
In \figuresautorefname\ref{fig:pc_roc} and \ref{fig:pc_pr}, the ROC and PR curves of the classifiers with different numbers of {\tt nTrees} approach to the {\it upper-left} and {\it upper-right} corners, respectively, showing good overall performance of the classifiers. 
Moreover, the areas under the curves (ROC-AUC and PR-AUC) of the classifiers are reasonably high, compared to a random guess, as shown in \figureautorefname\ref{fig:pc_auc}, with means of 0.9298 (St dev = 0.0072) and 0.9044 (St dev = 0.0346) across all given values of {\tt nTrees}, respectively. 
This shows the {\it probability} that the classifier ranks a randomly chosen positive example (Artcode) higher than a randomly chosen negative example (non-Artcode), denoted as follows: 
\begin{equation}
	\Pr\bigl(score(y = 1) > score(y = 0)\bigr) = 0.93
\end{equation}
This also reflects the good performance of the configured classifier in the experiment --- the {\sc ArtcodePresence} with random forests.

\subsubsection{{\sc ArtcodePresence} with SVM versus Random Forests on SaTAD-4}
\label{subsec:artcodePresenceSVM}
This section further studies the performance of {\sc ArtcodePresence} with a SVM \citep{cortes1995support} classification method, comparing its performance with the previous random forests-powered {\sc ArtcodePresence}.
The {\sc ArtcodePresence} with SVM is implemented similarly to {\sc ArtcodePresence} with random forests, except that it uses a different classification method. 
As described in \sectionautorefname \ref{sec:svm_classifier}, SVM is considered as the most effective machine learning method that can handle both linear and non-linear classification problems using the {\it kernel method} \citep{smola1998learning}. 
Compared to random forests, a SVM classifier needs to adjust with kernels and some other tuning parameters to achieve its optimal performance. 
As evaluated in \sectionautorefname \ref{sebsec:experimentalResults}, the performance of random forests classifier stabilised when {\tt nTrees} was greater than 40. 
The random forests achieved close performance between {\tt nTrees} $ = 80$ and {\tt nTrees} $ = 200$. 
Considering the computational cost, this experiment used the random forests that contains 80 decision trees, comparing its performance with a SVM classifier on a same dataset --- the SaTAD-4 dataset.

\begin{figure}[!t]
    \centering
    \begin{subfigure}[b]{0.45\textwidth}
        \includegraphics[width=\textwidth]{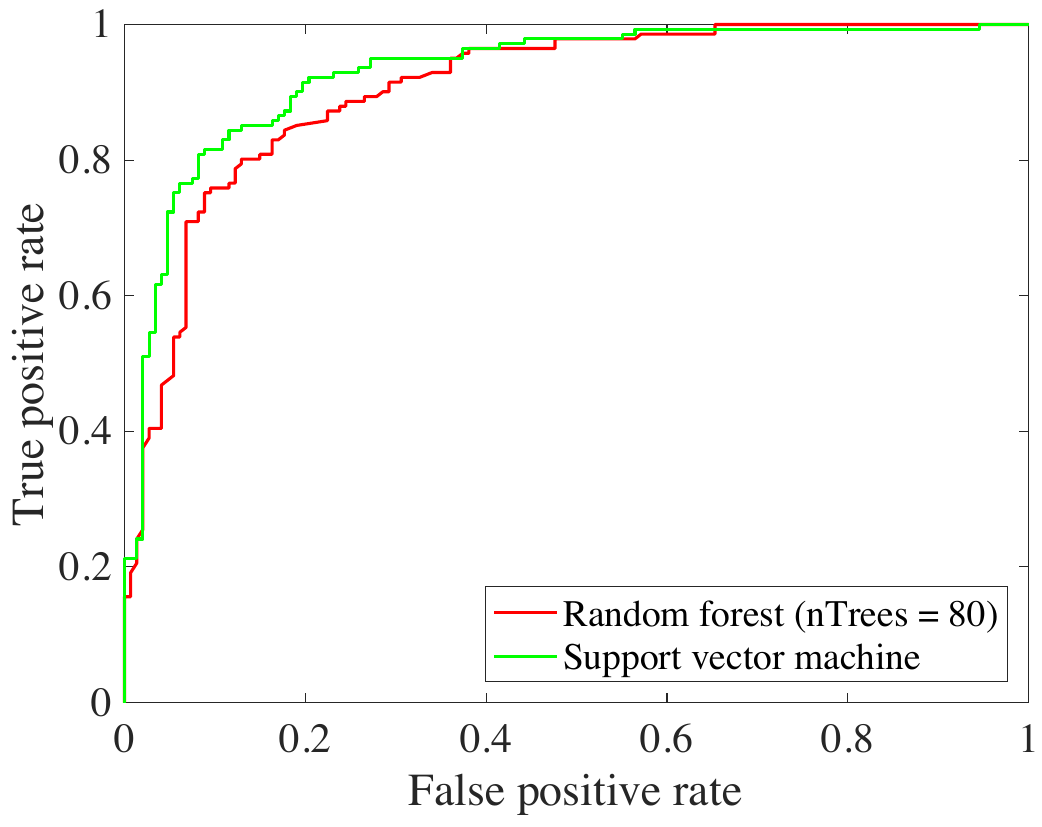}
        \caption{ROC curves}
        \label{fig:cp_roc}
    \end{subfigure}   
     \begin{subfigure}[b]{0.45\textwidth}
        \includegraphics[width=\textwidth]{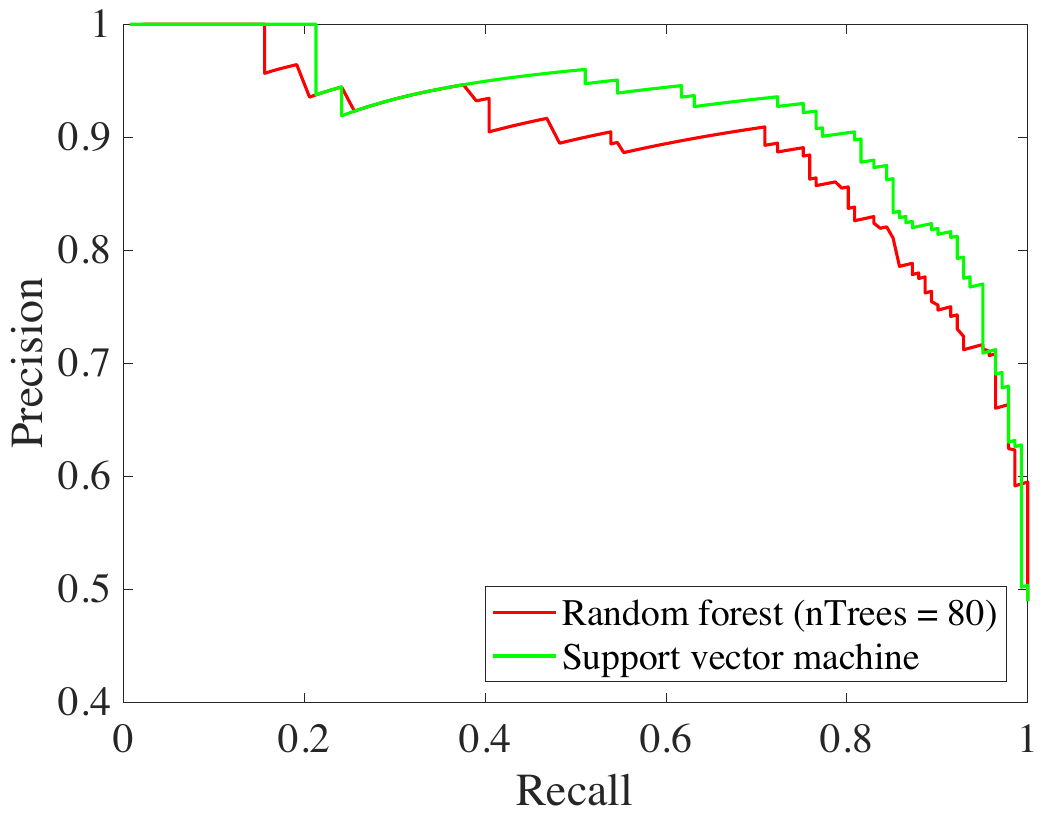}
        \caption{PR curves}
        \label{fig:cp_pr}
    \end{subfigure}  
     \begin{subfigure}[b]{0.6\textwidth}
        \includegraphics[width=\textwidth]{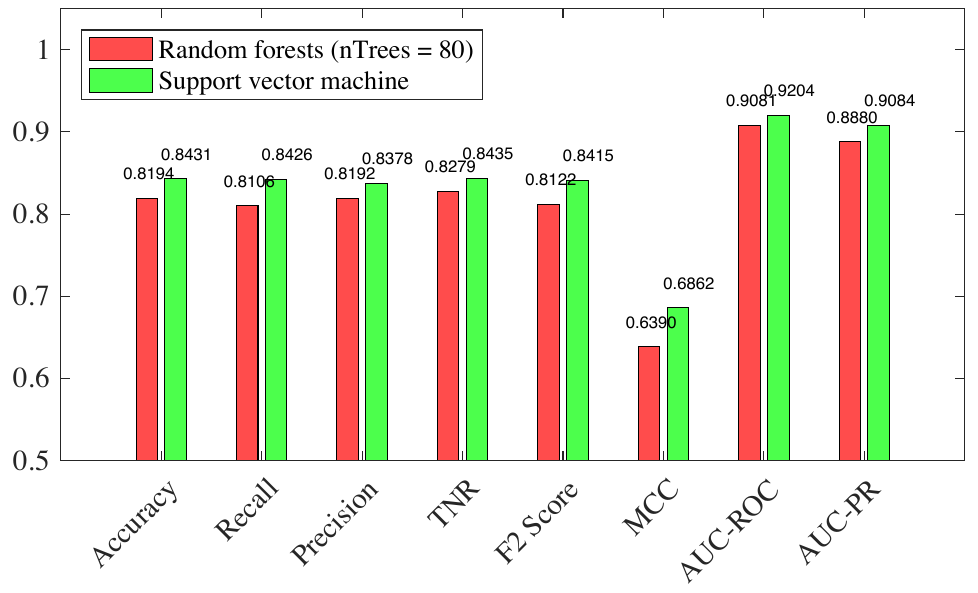}
        \caption{Performance metrics and the areas under the ROC and PR curves (ROC-AUC and PR-AUC).}
        \label{fig:cp_bar}
    \end{subfigure}   
    \caption{
        Performance comparison between {\sc ArtcodePresence} with random forests ({\tt nTrees} = 80) and SVM. 
        In all graphs, {\sc ArtcodePresence} with SVM performs better performance than random forests.
    }
    \label{fig:comparePerformance}
\end{figure}

In this experiment, the SVM classifer that employs the Gaussian or Radial Basis Function (RBF) (\equationautorefname \ref{eq:rbfKernel}) kernel was adopted as the classification method. 
Considering the differing importance between the two classes: Artcode and non-Artcode, the misclassification cost of this classifier is not equivalent for different classification errors. 
A square cost matrix is fed into the classifier to specify the misclassification cost, which if the true class of an observation is $i$, $Cost(i, j)$ is the cost of classifying a sample into class $j$. 
Tuning work was conducted to obtain a good cost matrix, and the following cost matrix achieved reasonably good results and it was therefore used in the classifier:
\renewcommand{\kbldelim}{(}
\renewcommand{\kbrdelim}{)}
\begin{equation}\label{eq:costMatrix}
	Cost_{i, j} = 
	\kbordermatrix{
	& \text{predicted non-Artcode} & \text{predicted Artcode} \\
	\text{true non-Artcode} &0 & 1 \\
	\text{true Artcode} & 3 & 0 
	}
\end{equation}

\subsubsection*{Experimental Results}
The experiments were conducted under the same experimental setting as the previous experiments. 
\figureautorefname \ref{fig:comparePerformance} shows the results of {\sc ArtcodePresence} with the SVM classifier: 
\figureautorefname\ref{fig:cp_roc}) and \figureautorefname\ref{fig:cp_pr} shows the ROC and PR curves of the random forests and SVM classifier; 
\figureautorefname\ref{fig:cp_bar} shows the performance of the two classifiers in terms of accuracy, recall, precision, TNR, F2 measure, MCC, and the areas under the ROC (ROC-AUC) and PR (PR-AUC) curves, with the numbers on the top of the bars showing their respective values.

The SVM classifier (green in \figureautorefname\ref{fig:comparePerformance}) slightly outperformed the random forests classifier, with the ROC curve being closer to the (0, 1) corner (see AUC results in \figureautorefname\ref{fig:cp_bar}) and the PR curve approaching the (1, 0) corner. 
Additionally, the average values of the evaluation metrics shown in \figureautorefname \ref{fig:cp_bar} shows the better performance of the SVM classifier --- with slightly higher values in term of all of the evaluation measurements. 
This experiment showed that Artcode classification can be effective in both random forests and SVM framework using the SOH feature vector and the SMOTE-augmented dataset. 
Compared to the baseline classifier --- {\sc ArtcodePresence} with random forests ({\tt nTrees} $= 80$) --- {\sc ArtcodePresence} with the SVM classifier has a much lower computational cost, which could be an improved version of the baseline classifier.

The two classifiers (random forests and SVM) trained on the SaTAD-4 dataset obtain good quality results, in both recall (the most important measurement for an effective detection proposal generator \citep{hosang2016makes,zitnick2014edge}) and other measurements, such as precision, F2 measure, MCC, accuracy, AUC-ROC, and AUC-PR. 
Because the classifiers are built on SOH-07, they are sufficiently time-efficient for proposing Artcodes in an input images. 
This experiment also shows the performance of {\sc ArtcodePresence} is somewhat insensitive to a certain class of classification methods, indicating the effectiveness and efficiency of the proposed SOH feature descriptor.

\section{Discussion and Implication}
\label{sec:discussion}
\begin{figure}[!t]
    \centering  
    \begin{subfigure}[b]{0.437\textwidth}
        \includegraphics[width=\textwidth]{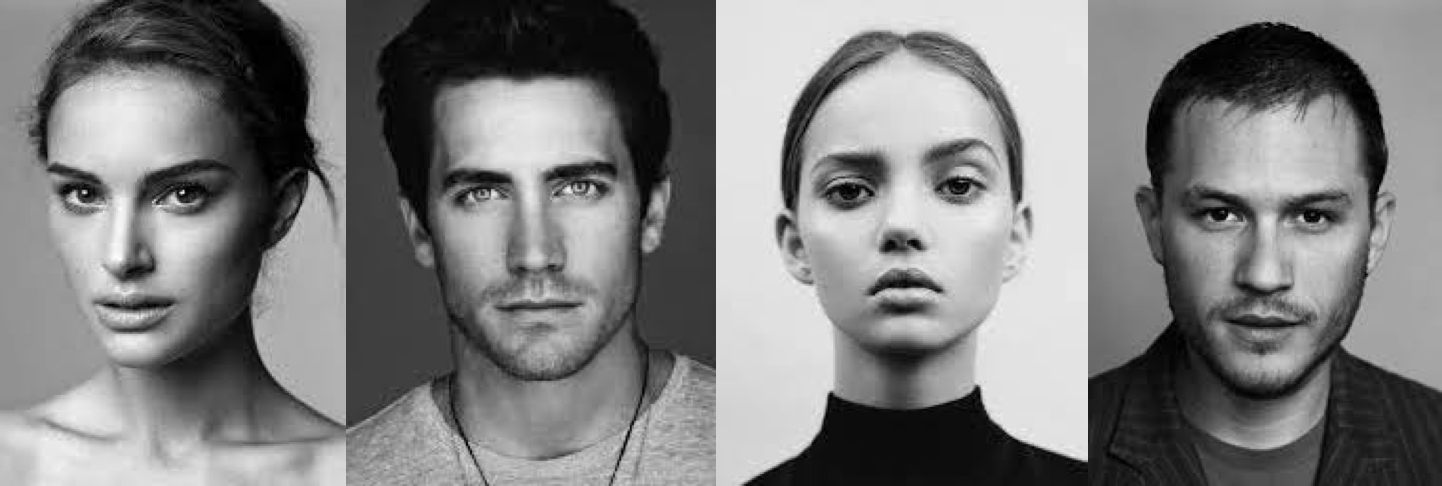}
        \caption{Faces}
        \label{fig:faceClass}
    \end{subfigure} 
    \begin{subfigure}[b]{0.513\textwidth}
        \includegraphics[width=\textwidth]{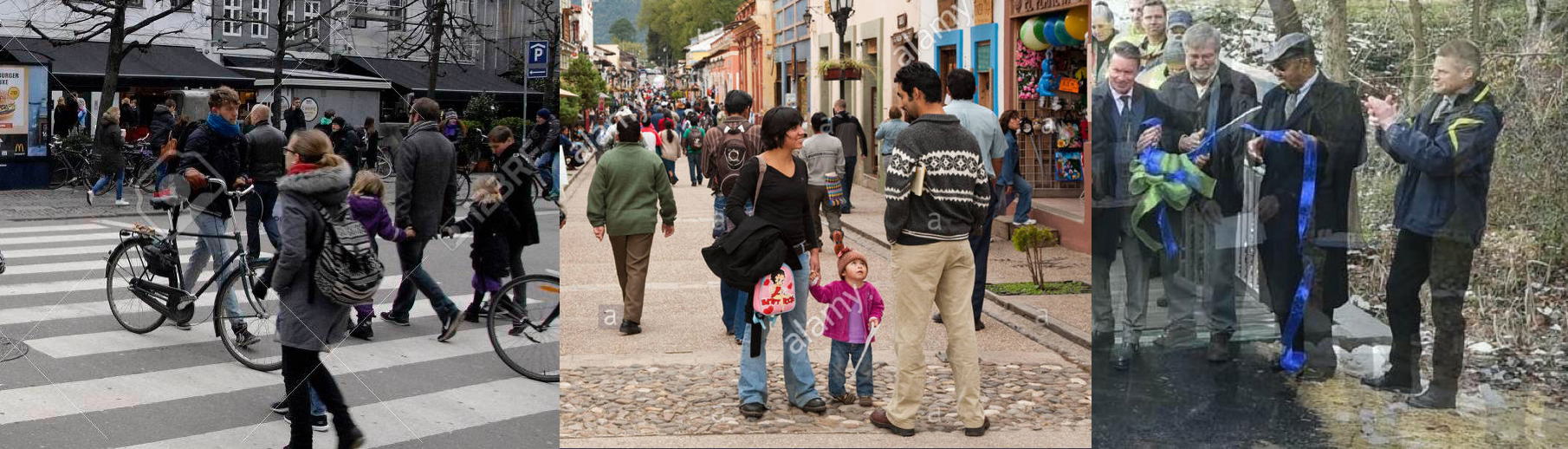}
        \caption{Humans}
        \label{fig:humanClass}
    \end{subfigure} 
    \begin{subfigure}[b]{0.48\textwidth}
        \includegraphics[width=\textwidth]{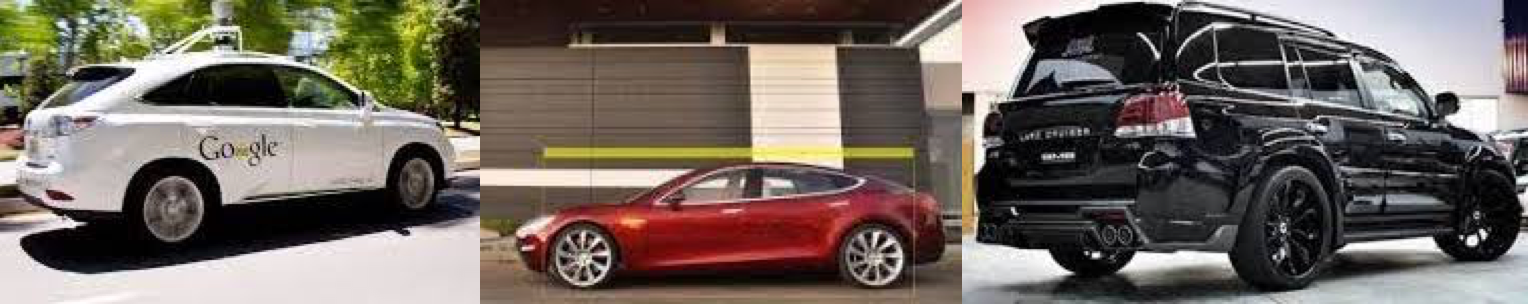}
        \caption{Cars}
        \label{fig:carClass}
    \end{subfigure} 
    \begin{subfigure}[b]{0.47\textwidth}
        \includegraphics[width=\textwidth]{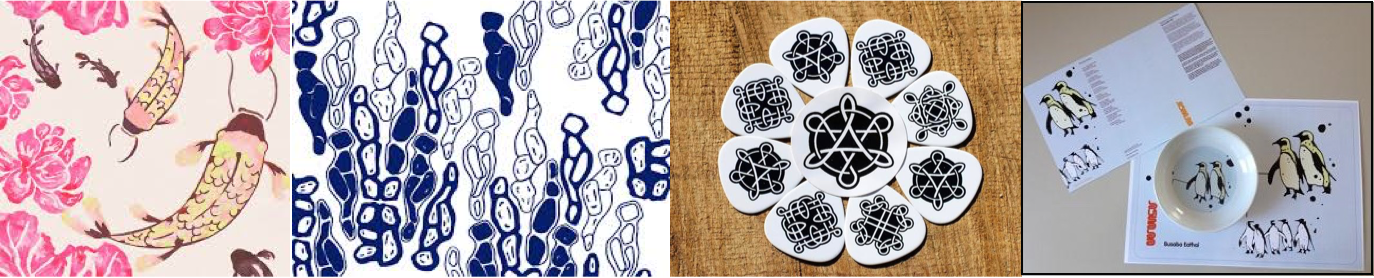}
        \caption{Artcodes}
        \label{fig:artcodeClass}
    \end{subfigure} 
    \caption{
        Example object classes (faces, cars, and humans) and Artcodes. 
        (\subref{fig:faceClass}), (\subref{fig:carClass}) and (\subref{fig:humanClass}) are common semantic object classes, which have relatively fixed shapes and share common visual features and structures. 
        In contrast, the Artcodes presented in (\subref{fig:artcodeClass}) are freeform but share some visual features that reflects topology such as high contrast areas. 
        They are grouped as a class because their topological structures follow the predefined topological rules of Artcodes.
    }
    \label{fig:objectClass}
\end{figure}

The systems build upon the SOH feature descriptor show promising performance on the task of Artcode proposal detection.
This work has multiple limitations.
First, due to the realistic constraints, the datasets collected are small and will continue to be small in the near future, especially comparing to other mainstream widely-used visual datasets such as ImageNet \citep{deng2009imagenet} and Fashion MINST \citep{xiao2017online}; small dataset increases the risk of outliers, overfitting, sampling bias, etc.
Having admitted this reality, we took measures to mitigate the impact of training on a small dataset, including using low-dimensional feature vectors, small-dataset-friendly classification models, cross-validation, and extensive experiments.
Second, despite their adequacy for a proposal detector, the performance of the proposed systems is still low, and there is a large room for improvement. 
Perhaps, the relatively low performance is due to 
1) the inherent difficulty of classifying objects via their topological structures; 
2) the low representation power of the current feature vector;
3) the small datasets. 
Further work needs to be conducted to tackle these three issues, including expanding the relevant datasets.
Third, the proposed feature vector is inspired from and specially designed for Artcode application; it might be unsuitable to be straightly applied into other scenarios due to the lack of studies on generalisation. 
Further studies on similar problems would be necessary to complement this work.

Despite the limitations, this findings of this work would contribute both human-computer interaction (HCI) and computer vision domain. 
The two properties --- smoothness and symmetry --- used for constructing the SOH draw the attention on the shape of an orientation histogram for topological representation. 
Moreover, Artcode detection proposal, the focal problem of this paper, opens up new interaction trajectories for AR applications as discussed in \citet{xu2022connecting} and would implicates a new computer vision problem --- topological object detection.
We discuss topological object detection further below.

Conventional object class detection deals with detecting instances of objects of a certain class, such as faces, humans, and cars. 
\figuresautorefname\ref{fig:faceClass}, \ref{fig:carClass}, and \ref{fig:humanClass} show the object classes of face, car, and pedestrian. 
These classes have their own {\it semantic} components that have relatively fixed geometrical shapes and appearance (such as eyes, mouths, tyres, hands and legs), and can be distinguished from other object classes by visual inspection. 
However, as shown in \figureautorefname\ref{fig:artcodeClass}, the four Artcode objects are freeform, without fixed geometric shapes, but they do actually share some visual features such as high contrast edges, which reflects the image topology to some extent. 
They are grouped in a same class because of their predefined topological structures. 
In this work, we called the detection of an object class that shares common generic topological structure as {\bf topological object detection}.  
Artcode is a special kind of topological object that created by humans. 
Topoligical object detection may be useful for situations that categorise objects based on their generic topological description rather than semantic and geometrical description, such as similar geometric shapes. 
The main differences between topological object detection and conventional object detection are described as follows: 
\begin{description}[style=unboxed, leftmargin=1cm]
    \item[\textit{Geometry vs Topology}]
    Geometry deals with shapes, relative positions, sizes of figures, and properties of space such as curvature, while topology studies the properties of space that are preserved under continuous deformations: stretching and bending, but not cutting or gluing. 
    Topological objects such as Artcodes have their own predefined topology, and offers freedom in geometrical properties such as curvature, shape or position. 
    Conventional object classes such as cars, faces or humans have their own special geometrical features. 
    Therefore, they have similar appearances and can be easily identified by humans. 
    Objects of same topological class can have completely different geometrical shapes, and would be even difficult for humans to categorise them only by visual inspection. 

    \item[\textit{Local vs Global}]
    Geometry has {\it local} structure, while topology has {\it global} structure. 
    Local geometric features or their aggregated structure (such as the BoG descriptor \citep{csurka2004visual}) can represent common object classes; they may not be able to represent global topology.
    Thus, topological object detection needs to examine global features rather than only local information.
    Carefully designed features are required to describe such topological classes as Artcodes, which follow a predefined topological structure. 

\end{description}

Due to space limitation, the topological object detection problem is briefly presented. 
Properly establishing this problem needs to do more explorations on similar problems.
We hope this work will inspire and facilitate the potential applications on describing topological structures in other scenarios {\it beyond} Artcode detection.

\section{Conclusion and Future Work}
\label{sec:conclusion}
In this paper, we have reported on a problem that recognise the presence of ``invisible'' yet structured markers in images, using Artcode as a case for study.
we have proposed a new feature-based system for detection such visually ``hidden'' markers. 
Thanks to a novel feature set and a machine learning-driven approach, the proposed system --- {\sc ArtcodePresence} --- can detect the visual marker proposals via their high-level characteristics, in contrast to the previous detect-by-matching or -decoding methods \citep{costanza2003d, costanza2009designable, Fiala2005} which need a low-level, full-detail inspection.
The new system requires relatively loose constraints on imaging conditions, which can discover visual markers with occlusions, under bad lighting condition, and at a relatively long distance. 
Therefore, it technically broadens design space and opens up new interaction opportunities in visual marker-powered scenarios.

Specifically, we have proposed the Shape of Orientation Histogram (SOH), which is a feature descriptor for representing the generic topological structures of Artcodes.
This feature descriptor is built upon the two properties of the orientation histogram: symmetry and smoothness. 
Shape studies were conducted to empirically study the feasibility of these two properties on describing topological structures.
We have conducted to compare SOH with conventional geometric feature sets, such as BoW and HoG for detecting Artcode proposals.
Experimental results showed that showed that SOH outperforms HoG and BoW in terms recall and efficiency --- the two critical measures for evaluating detection proposal methods. 
Because SOH uses topological information for representation, it would be explainable and interpretable to designers, who have little background on computer vision.

We have also conducted extensive experiments to evaluate the performance of the proposed system, showing its effectiveness of on the classification of Artcodes versus non-Artcodes using Random Forest and SVM classification methods.
In addition, two datasets were collected for above experimental studies: the true Artcode dataset (TAD) and the extended Artcode dataset (EAD, extended by adding printed-out Artcodes, which provided a larger and more balanced dataset.).
Artcodes are hand-crafted by designers, and therefore, the total number of Artcode samples is expected to be small.
Becasue of the large number of available non-Artcode samples, these datasets are imbalanced, with many more negative samples than positives.
The dataset imbalance handling technique --- SMOTE ---  was therefore used to tackle this imbalance, creating an augmented dataset with a relatively balanced class distribution. 
Experimental results showed the greatly improved performance using this SMOTE-augmented datasets.

The proposed system can be used for Artcode proposal generation, giving cues to users to guide their interactions with Artcodes before they can finally decode them. 
Artcode candidates, returned by the system may include a large number of false positives, can then be further examined by subsequent operation such as decoding. 
This work attempts to addressing the longstanding question in HCI community \citep{hinckley2000sensing}: how to interact with invisible sensing systems (the vision based AR system here) using the machine learning-driven methods. 
Meanwhile, this work implicates a vision task --- topological object detection --- which is a problem that classifies semantically and geometrically different but topologically similar objects as a same class. 
This work would enable and facilitate the follow-up works on exploring these two aspects.
Our future work includes extending the datasets using generative models and generalising the proposed system to other similar problems.

\bibliographystyle{unsrtnat}
\bibliography{references}  






\appendix
\section{Supplementary Material}
\label{app:supplementaryMaterial}

\subsection{Shape Studies}
\label{sec:shapeStudies}
We conducted so-called shape studies to study whether the two proposed properties can empirically capture the topological information.
A number of synthetic and real example images, and their corresponding edge maps, cumulative histograms, and orientation histograms are presented in \figuresautorefname \ref{fig:shapeStudies} and \ref{fig:realImageStudies} (see Appendix \ref{app:supplementaryMaterial}), respectively.
As shown in \figureautorefname \ref{fig:shapeStudies}, the left (red) part of the cumulative histograms is translationally symmetric to the right part to varing degree. 
\figureautorefname \ref{fig:shapeStudies} presents a number of synthetic shapes and their corresponding cumulative histogram curves. 
The shapes in \figureautorefname \ref{fig:shapeStudies} all have high contrast as well as thick and closed region boundaries --- the visual properties that generate a certain level of symmetry and smoothness of the orientation histogram. 
Therefore, the orientation around the boundaries can be smoothly distributed over the range from $-180^\circ$ to $180^\circ$ in most instances. 
This can be illustrated by the comparison between the circle (\figureautorefname\ref{fig:circle_im}), and the ellipse (\figureautorefname\ref{fig:ellipse_im}), and the rectangle (\figureautorefname\ref{fig:rectangle_im}) and triangle (\figureautorefname\ref{fig:triangle_im}): the cumulative histograms of the circle and ellipse are much smoother (smoothly varying) than that of the rectangle and triangle. 
For the circle and ellipse, the orientations of the gradients around the boundaries are continuously changing and are relatively uniformly distributed over the range from  $-180^\circ$ to $180^\circ$, whereas the rectangle and the triangle have four and three dominant orientations, respectively, which results in these sharp peaks in their cumulative histograms. 
In the terminology of this thesis, they are not ``Artcode-like'' (refer to \sectionautorefname \ref{sec:detectionProposal} for a detailed description of Artcode-like).

Another example is the two d-touch codes as shown in \figureautorefname\ref{fig:dtouch-1_im} and \figureautorefname\ref{fig:dtouch-2_im}. 
The d-touch marker in \figureautorefname\ref{fig:dtouch-1_im} is mainly composed of smooth and closed lines, while rectangles dominate the primary structure of the marker in \figureautorefname\ref{fig:dtouch-2_im}. 
Because of this distinction, although their cumulative histograms are both symmetrical, the cumulative histogram of the marker in \figureautorefname\ref{fig:dtouch-2_im} is not smooth, but has sharp changes. Hence, \figureautorefname\ref{fig:dtouch-1_im} is more ``Artcode-like'' than \figureautorefname \ref{fig:dtouch-2_im}. 
The cumulative histograms of the shapes of \figuresautorefname \ref{fig:occlusion-1_im} and \ref{fig:occlusion-2_im} are both smooth and symmetrical. It is worthy to note that although part of the main structure of \figuresautorefname\ref{fig:occlusion-2_im}) is occluded, its cumulative histogram is still relatively smooth and symmetrical to some extent. Theoretically, \figuresautorefname\ref{fig:dtouch-1_im}, \ref{fig:occlusion-1_im} and \ref{fig:occlusion-2_im} should be detected as ``Artcode-like'' becasue they follow the generic topological structure as defined in \definitionautorefname \ref{def:genericTopologicalStructure}. 
Alerting the user to their presence and thus guiding the user to shift viewpoint or even remove the occlusion is the purpose of this work.

\figureautorefname\ref{fig:shapeStudies} shows the shape of orientation histograms of synthetic images, while \figureautorefname\ref{fig:realImageStudies} shows the shape of orientation histograms of real images. 
The first six images in \figureautorefname\ref{fig:realImageStudies} are not Artcodes. 
Except in \figuresautorefname\ref{fig:realImage-4_im} and \ref{fig:realImage-5_im}, the cumulative histograms of the non-Artcodes are clearly not symmetrical. 
However, note that the cumulative histograms of \figuresautorefname\ref{fig:realImage-4_im} and \ref{fig:realImage-5_im} are not smooth and so do not meet both criteria. 
This is because the dominant structure of these two images are some parallel lines (see their edge maps in \figuresautorefname \ref{fig:realImage-4_im} and \ref{fig:realImage-5_im}, which are not closed, and the points along this straight line only have the same directions.

The last six images (\figuresautorefname \ref{fig:realImage-7_im} -- \ref{fig:realImage-12_im}), which are Artcode-like images) of \figureautorefname \ref{fig:realImageStudies} contain Artcodes or partial Artcodes. 
The cumulative histograms of all images are reasonably symmetrical and smooth; and the cumulative histograms of \figuresautorefname \ref{fig:realImage-7_im} and \ref{fig:realImage-8_im}, which contain partial Artcodes (fish and penguin), are less smooth than those of the last four images  (\figuresautorefname \ref{fig:realImage-9_im} -- \ref{fig:realImage-12_im}) but smoother than the first six images (\figuresautorefname\ref{fig:realImage-1_im} -- \ref{fig:realImage-6_im}), which are non-Artcode images). 
Therefore, it is desirable that the last four images (\figuresautorefname\ref{fig:realImage-9_im} -- \ref{fig:realImage-12_im}) should be detected as ``Artcodes'' with greater likelihood than those of the 6th (\figureautorefname\ref{fig:realImage-7_im}) and 7th (\figureautorefname\ref{fig:realImage-7_im}) images.

These observations inspired me to employ both the symmetry and smoothness of the orientation histogram for representing the general topological structure of Artcodes. Based on these two properties, a feature vector known as the Shape of Orientation Histogram (SOH) was constructed. The two properties --- symmetry and smoothness --- cannot {\em perfectly} describe all Artcodes, as Artcodes can take a wide variety of appearances. But it provides additional evidence to propose the potential Artcode-like candidates, which would then be further examined by subsequent decoding.

\begin{figure}[!t]
    \centering
    \begin{subfigure}[t]{0.16\textwidth}
    \stackinset{c}{}{b}{1.10in}{Example shape}{%
        \includegraphics[width=\textwidth]{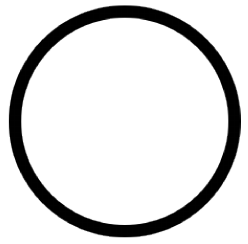}}
        \caption{}
        \label{fig:circle_im}
    \end{subfigure}
    \begin{subfigure}[t]{0.16\textwidth}
    \stackinset{c}{}{b}{1.10in}{Edge map}{%
        \includegraphics[width=\textwidth]{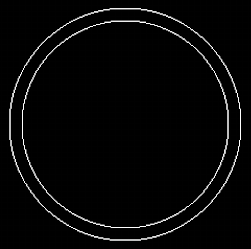}}
        \label{fig:circle_cum}
    \end{subfigure}   
    \begin{subfigure}[t]{0.30\textwidth}
    \stackinset{c}{}{b}{1.10in}{Orientation histogram}{%
        \includegraphics[width=\textwidth]{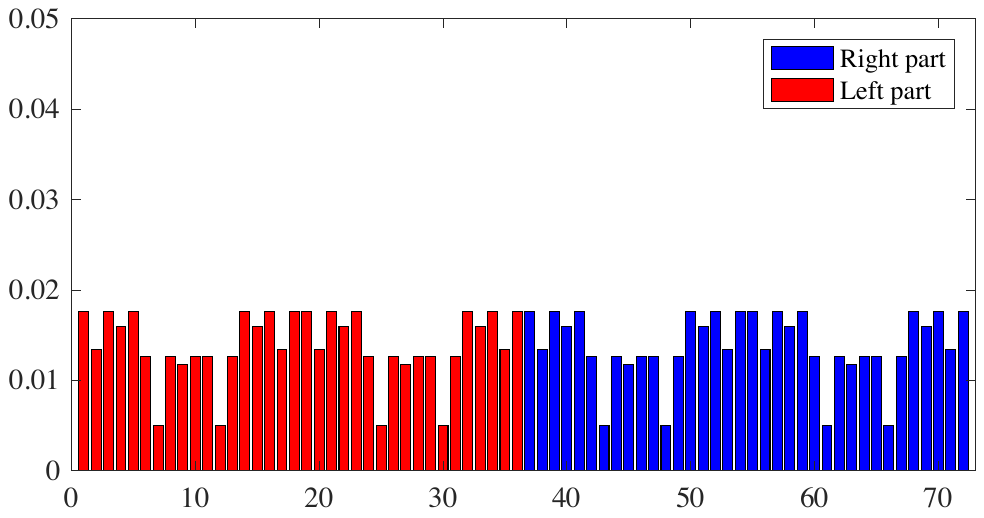}}
        \label{fig:circle_oh}
    \end{subfigure}
    \begin{subfigure}[t]{0.30\textwidth}
    \stackinset{c}{}{b}{1.10in}{Cumulative histogram curve}{%
       \includegraphics[width=\textwidth]{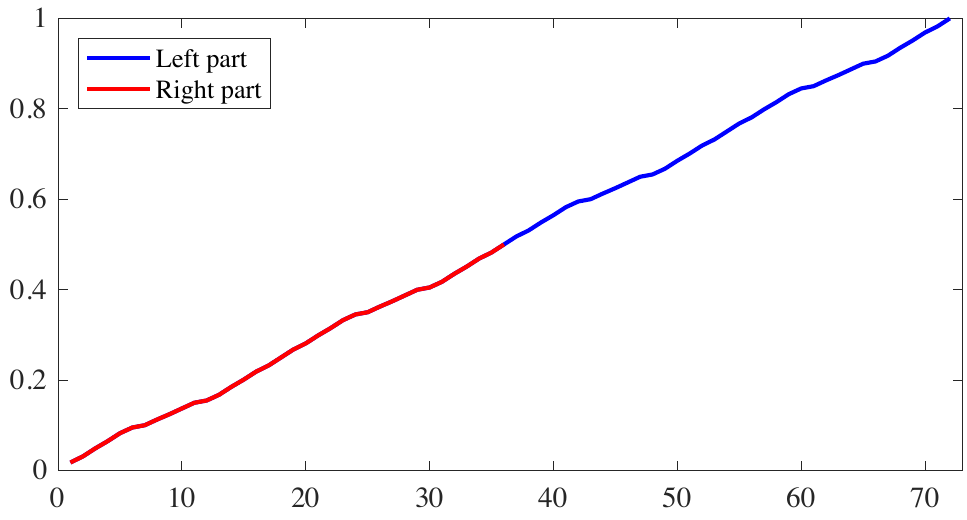}}
       \label{fig:circle_cum}
    \end{subfigure}   
    
    \begin{subfigure}[t]{0.16\textwidth}
        \includegraphics[width=\textwidth]{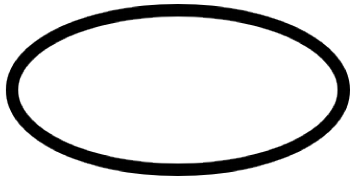}
        \caption{}
        \label{fig:ellipse_im}
    \end{subfigure}
    \begin{subfigure}[t]{0.16\textwidth}
        \includegraphics[width=\textwidth]{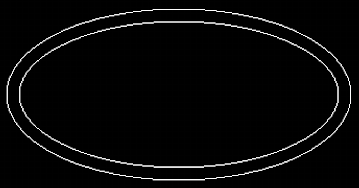}
        \label{fig:ellipse_em}
    \end{subfigure}
    \begin{subfigure}[t]{0.30\textwidth}
        \includegraphics[width=\textwidth]{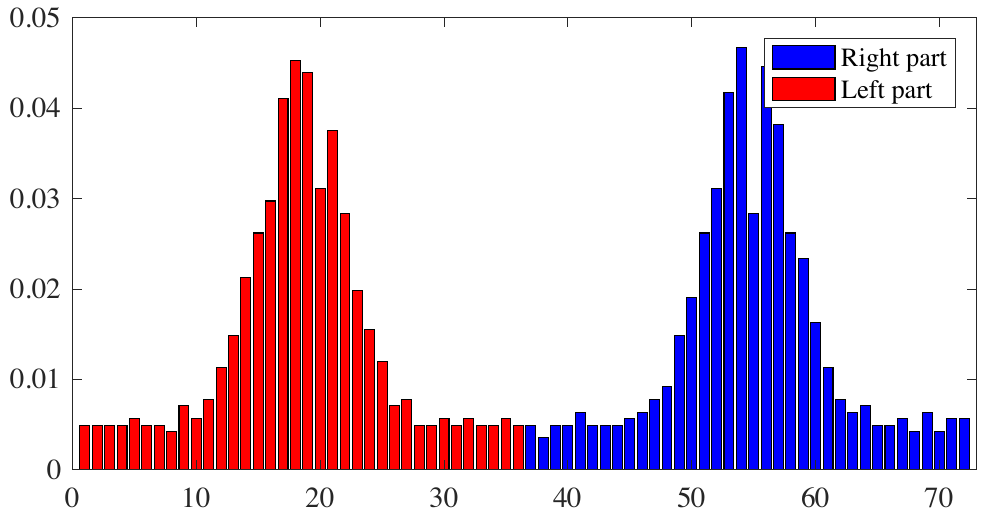}
        \label{fig:ellipse_oh}
    \end{subfigure}
    \begin{subfigure}[t]{0.30\textwidth}
        \includegraphics[width=\textwidth]{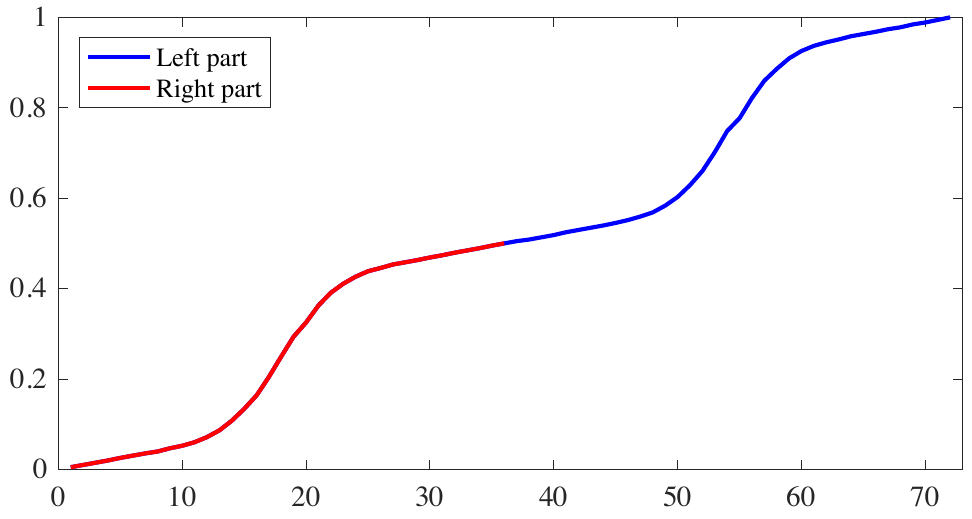}
        \label{fig:ellipse_cum}
    \end{subfigure} 
    \begin{subfigure}[t]{0.16\textwidth}
        \includegraphics[width=\textwidth]{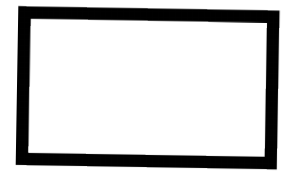}
        \caption{}
        \label{fig:rectangle_im}
    \end{subfigure}
    \begin{subfigure}[t]{0.16\textwidth}
        \includegraphics[width=\textwidth]{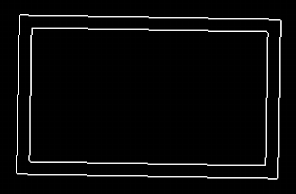}
        \label{fig:rectangle_em}
    \end{subfigure}
    \begin{subfigure}[t]{0.30\textwidth}
        \includegraphics[width=\textwidth]{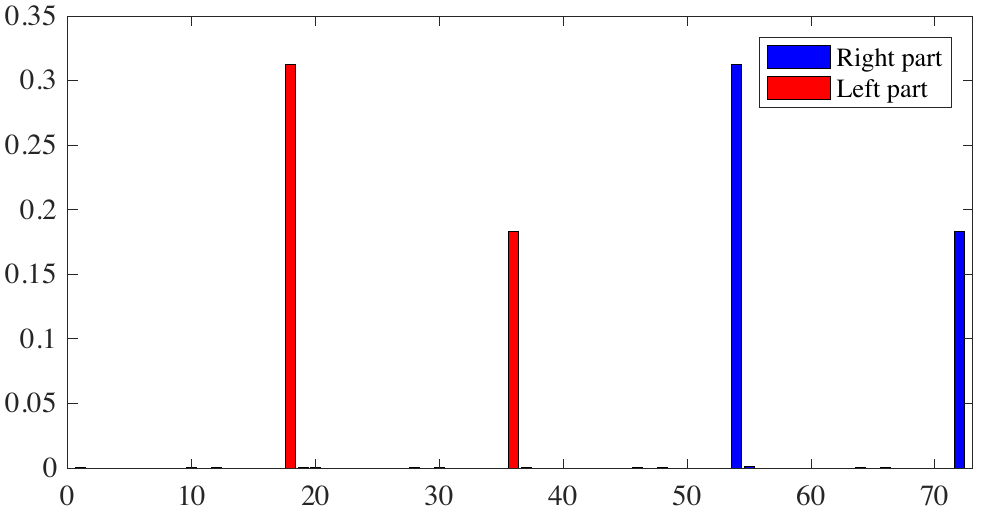}
        \label{fig:rectangle_oh}
    \end{subfigure} 
    \begin{subfigure}[t]{0.30\textwidth}
        \includegraphics[width=\textwidth]{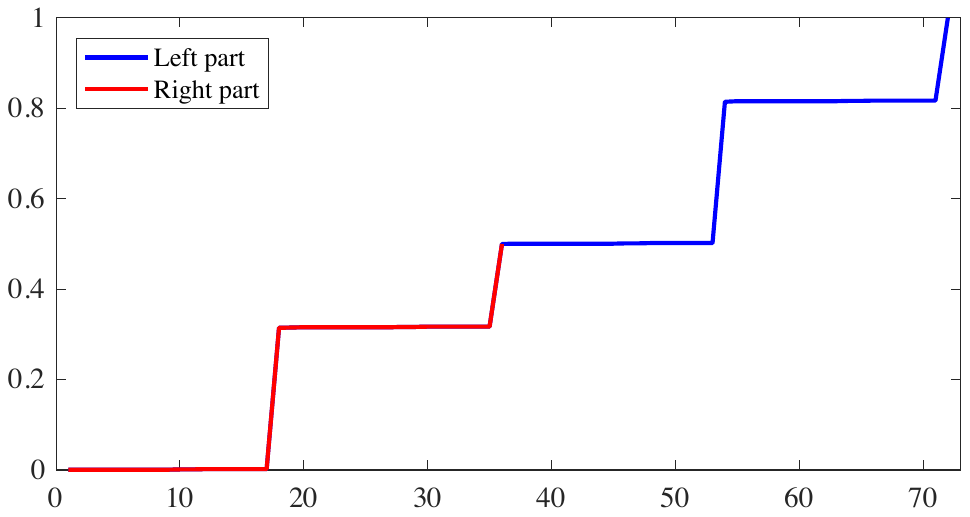}
        \label{fig:rectangle_cum}
    \end{subfigure} 
    \begin{subfigure}[t]{0.16\textwidth}
        \includegraphics[width=\textwidth]{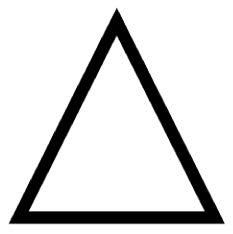}
        \caption{}
        \label{fig:triangle_im}
    \end{subfigure}
    \begin{subfigure}[t]{0.16\textwidth}
        \includegraphics[width=\textwidth]{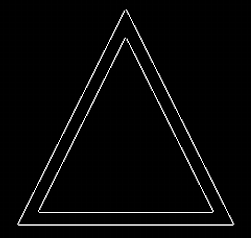}
        \label{fig:triangle_em}
    \end{subfigure}
    \begin{subfigure}[t]{0.30\textwidth}
        \includegraphics[width=\textwidth]{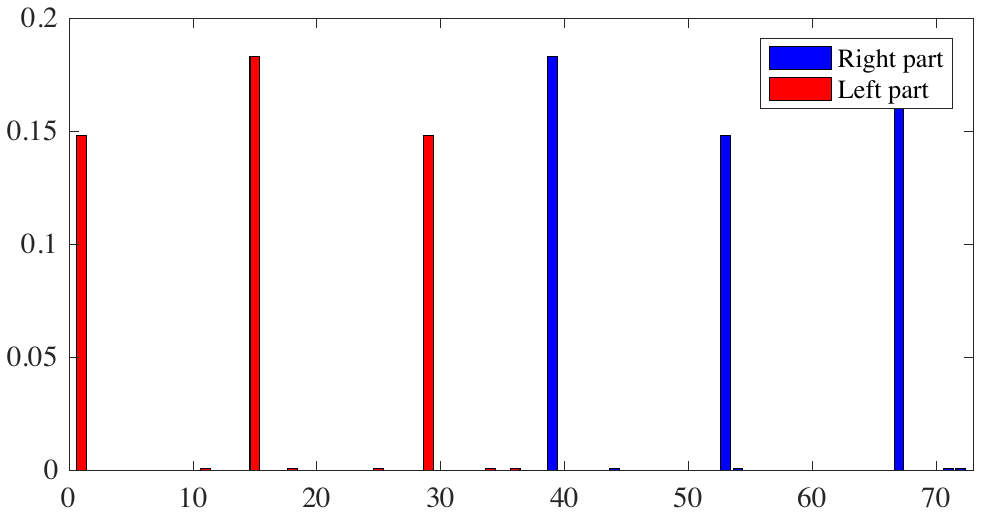}
        \label{fig:ellipse-circle_oh}
    \end{subfigure}
    \begin{subfigure}[t]{0.30\textwidth}
        \includegraphics[width=\textwidth]{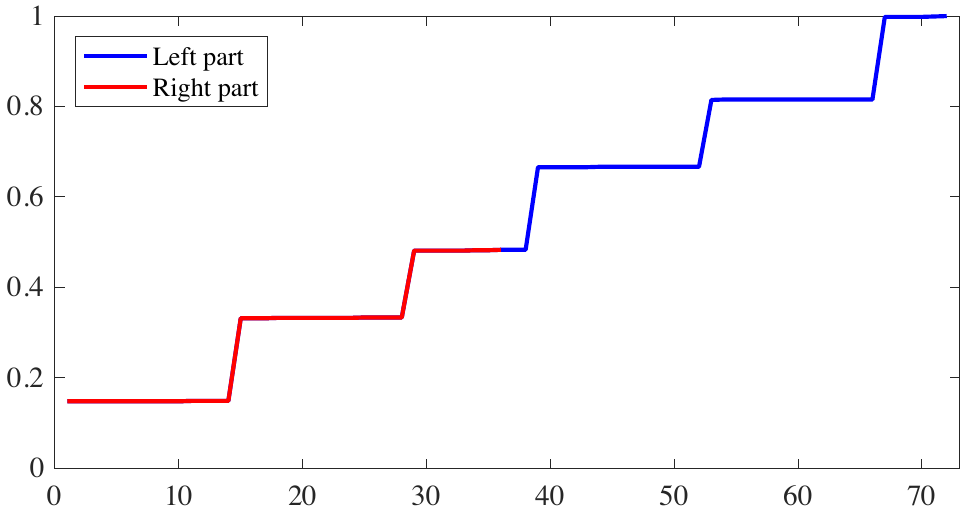}
        \label{fig:triangle_cum}
    \end{subfigure}
    
    \begin{subfigure}[t]{0.16\textwidth}
        \includegraphics[width=\textwidth]{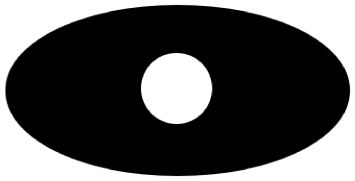}
        \caption{}
        \label{fig:ellipse-circle_im}
    \end{subfigure}
    \begin{subfigure}[t]{0.16\textwidth}
        \includegraphics[width=\textwidth]{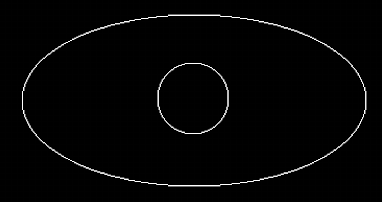}
        \label{fig:ellipse-circle_em}
    \end{subfigure}  
    \begin{subfigure}[t]{0.30\textwidth}
        \includegraphics[width=\textwidth]{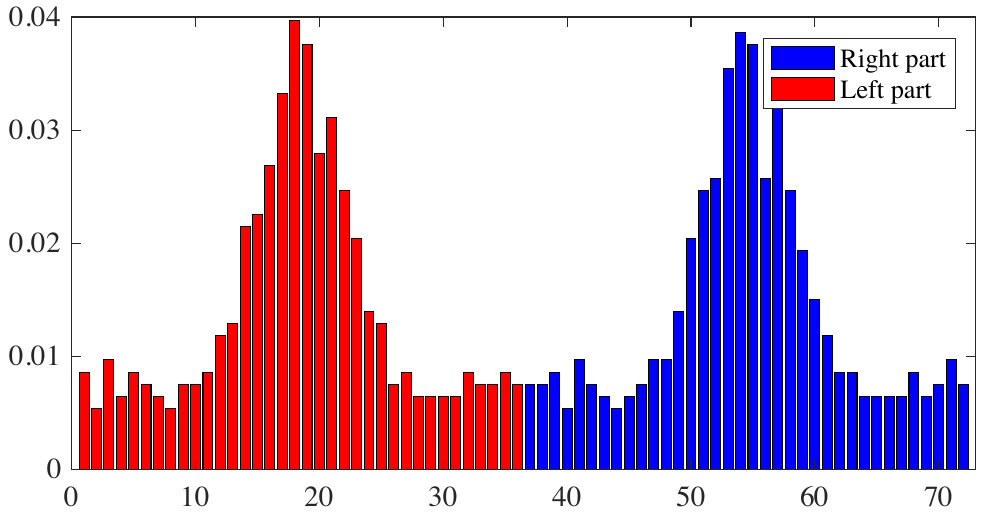}
        \label{fig:ellipse-circle_oh}
    \end{subfigure}
    \begin{subfigure}[t]{0.30\textwidth}
        \includegraphics[width=\textwidth]{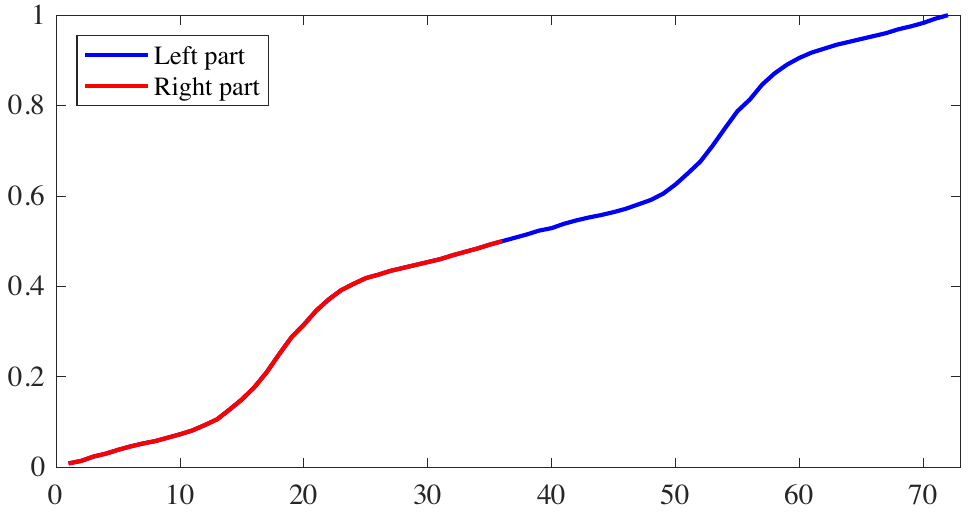}
        \label{fig:ellipse-circle_cum}
    \end{subfigure}  
    
    \begin{subfigure}[t]{0.16\textwidth}
        \includegraphics[width=\textwidth]{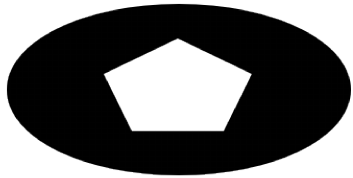}
        \caption{}
        \label{fig:ellipse-polygon_im}
    \end{subfigure}
    \begin{subfigure}[t]{0.16\textwidth}
        \includegraphics[width=\textwidth]{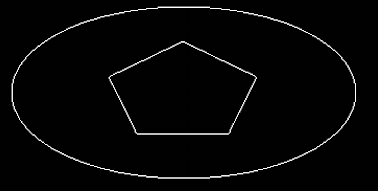}
        \label{fig:ellipse-polygon_em}
    \end{subfigure} 
    \begin{subfigure}[t]{0.30\textwidth}
        \includegraphics[width=\textwidth]{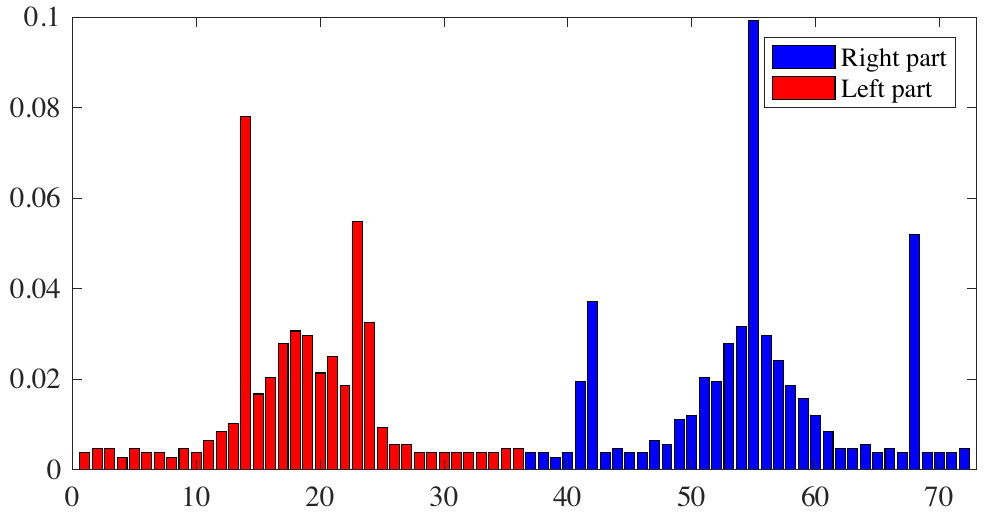}
        \label{fig:ellipse-polygon_oh}
    \end{subfigure}
    \begin{subfigure}[t]{0.30\textwidth}
        \includegraphics[width=\textwidth]{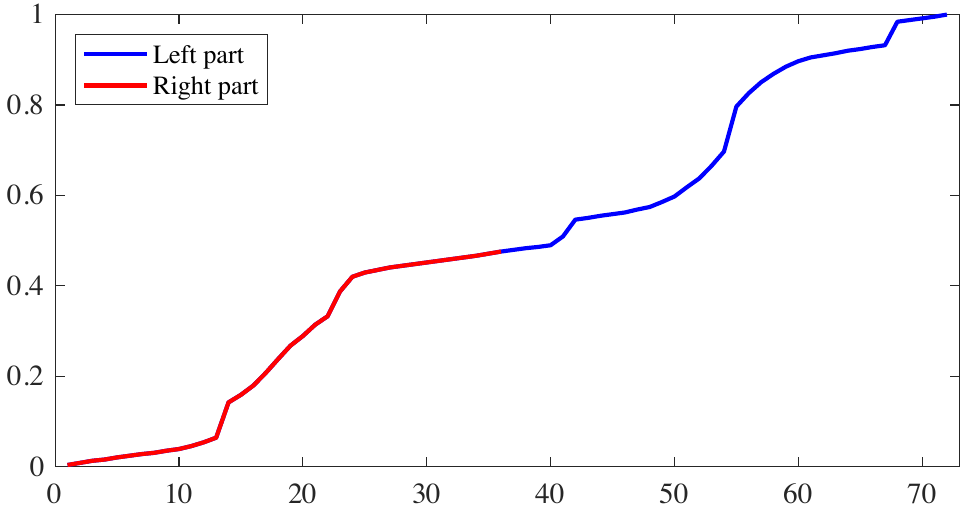}
        \label{fig:ellipse-polygon_cum}
    \end{subfigure} 
    
    \begin{subfigure}[t]{0.16\textwidth}
        \includegraphics[width=\textwidth]{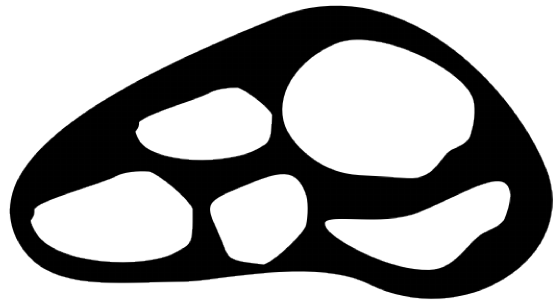}
        \caption{}
        \label{fig:curves-2_im}
    \end{subfigure}
    \begin{subfigure}[t]{0.16\textwidth}
        \includegraphics[width=\textwidth]{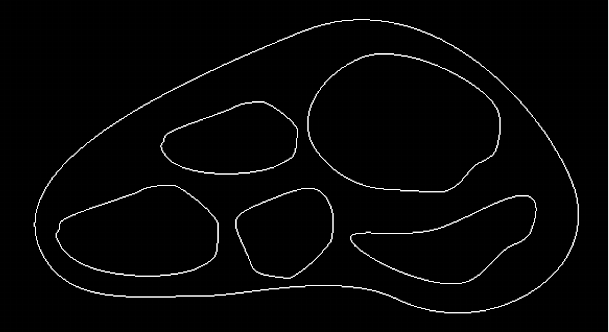}
        \label{fig:curves-2_em}
    \end{subfigure}  
    \begin{subfigure}[t]{0.30\textwidth}
        \includegraphics[width=\textwidth]{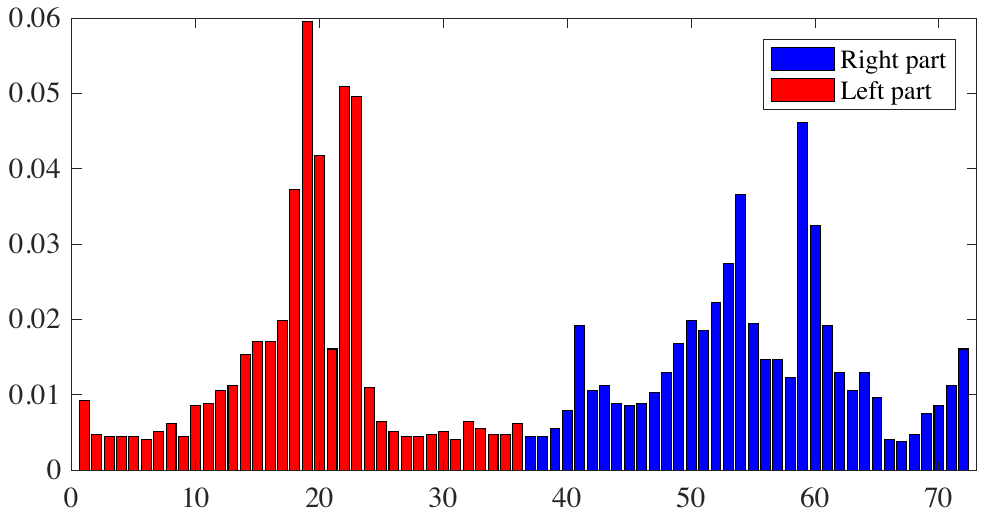}
        \label{fig:curves-2_oh}
    \end{subfigure}
    \begin{subfigure}[t]{0.30\textwidth}
        \includegraphics[width=\textwidth]{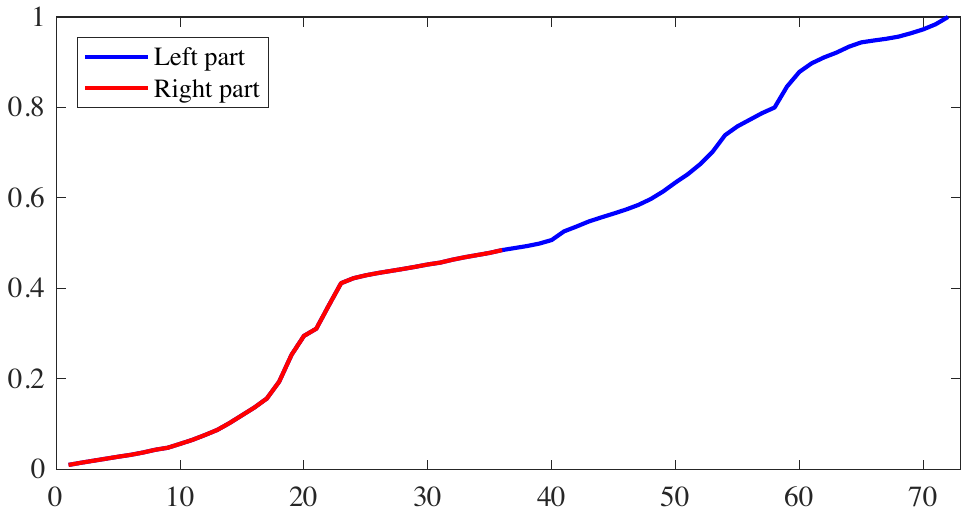}
        \label{fig:curves-2_cum}
    \end{subfigure}   
\end{figure}

\begin{figure}[t!]
          \ContinuedFloat
          \centering
            \begin{subfigure}[t]{0.16\textwidth}
            \stackinset{c}{}{b}{1.10in}{Example shape}{%
                  \includegraphics[width=\textwidth]{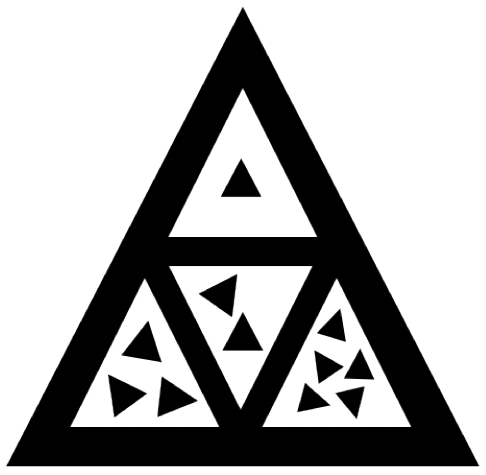}}
                  \caption{}
                  \label{fig:triangles_im}
         \end{subfigure}
         \begin{subfigure}[t]{0.16\textwidth}
          \stackinset{c}{}{b}{1.10in}{Edge map}{%
                   \includegraphics[width=\textwidth]{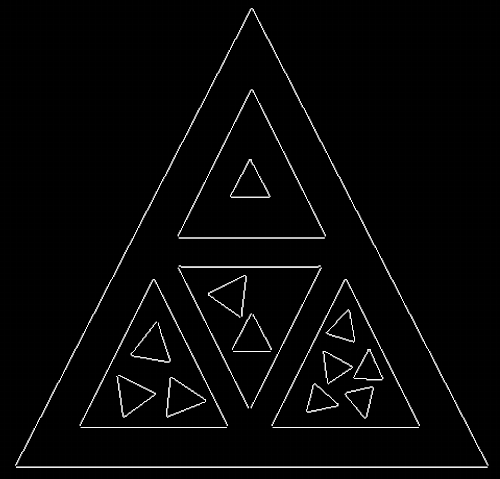}}
                   \label{fig:triangles_em}
          \end{subfigure}
           \begin{subfigure}[t]{0.30\textwidth}
             \stackinset{c}{}{b}{1.10in}{Orientation histogram}{%
                  \includegraphics[width=\textwidth]{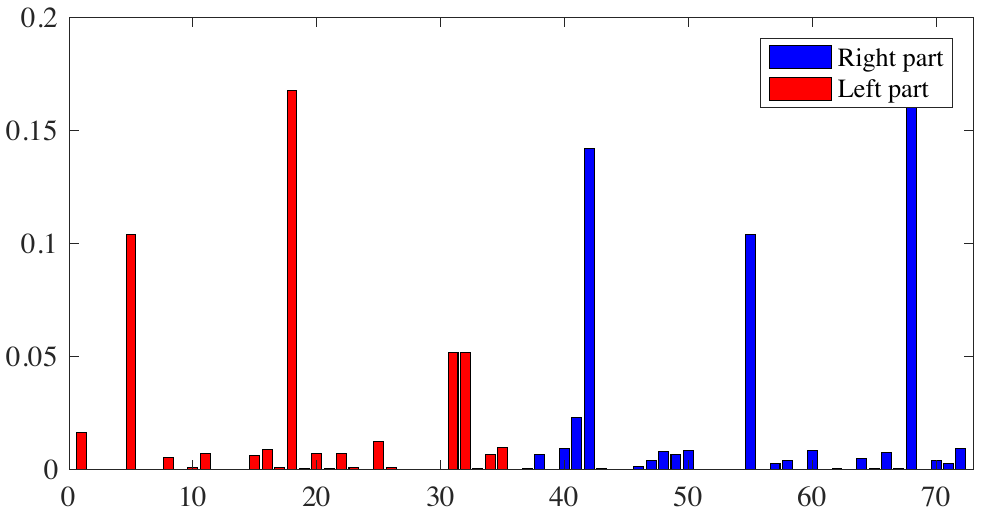}}
                  \label{fig:triangles_oh}
         \end{subfigure}
         \begin{subfigure}[t]{0.30\textwidth}
         \stackinset{c}{}{b}{1.10in}{Cumulative histogram curve}{%
               \includegraphics[width=\textwidth]{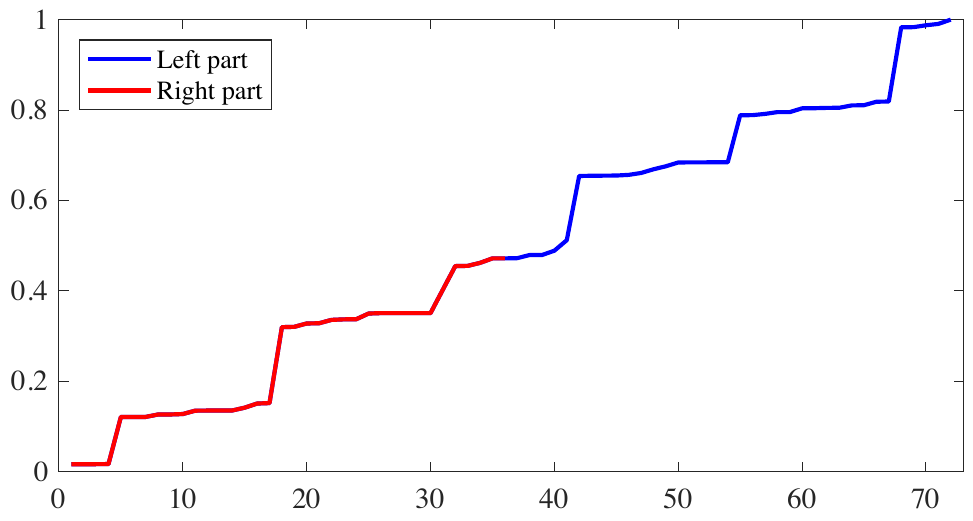}}
               \label{fig:triangles_cum}
          \end{subfigure}
          
          \begin{subfigure}[t]{0.16\textwidth}
                  \includegraphics[width=\textwidth]{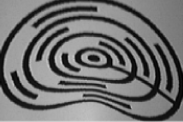}
                  \caption{}
                  \label{fig:dtouch-1_im}
         \end{subfigure}
         \begin{subfigure}[t]{0.16\textwidth}
                   \includegraphics[width=\textwidth]{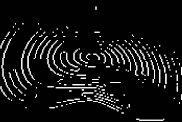}
                   \label{fig:dtouch-1_em}
          \end{subfigure}  
          \begin{subfigure}[t]{0.30\textwidth}
                  \includegraphics[width=\textwidth]{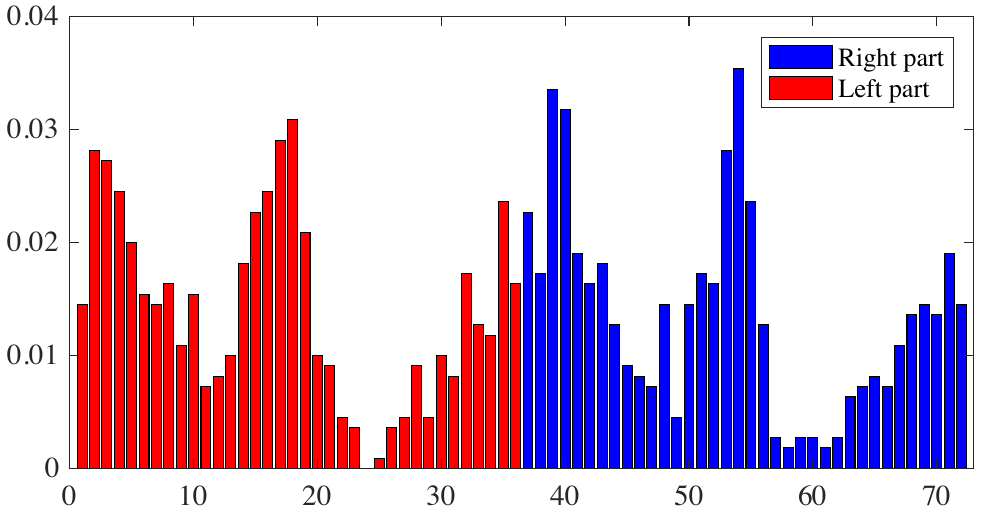}
                  \label{fig:dtouch-1_oh}
         \end{subfigure}
         \begin{subfigure}[t]{0.30\textwidth}
                   \includegraphics[width=\textwidth]{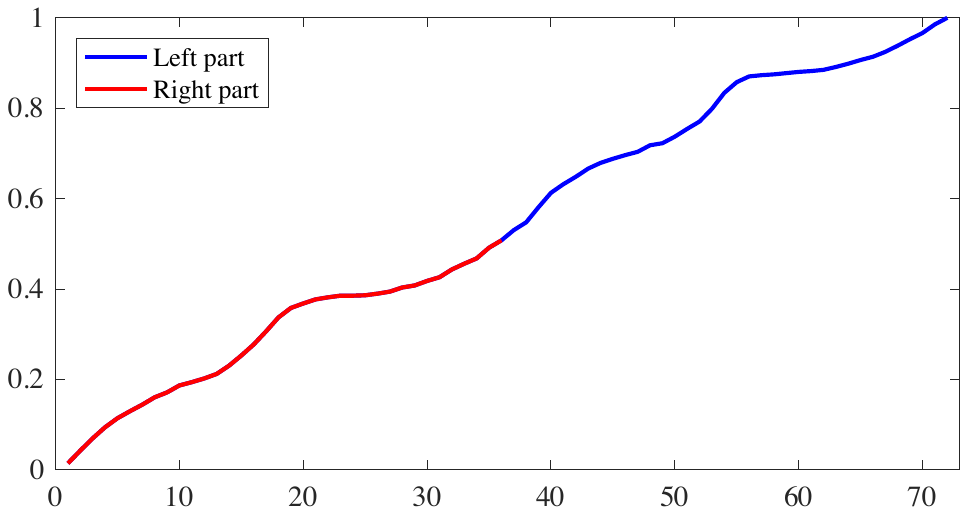}
                   \label{fig:dtouch-1_cum}
          \end{subfigure}  
          
           \begin{subfigure}[t]{0.16\textwidth}
                  \includegraphics[width=\textwidth]{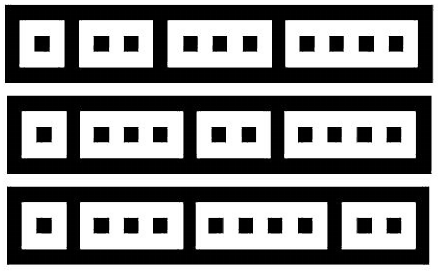}
                  \caption{}
                  \label{fig:dtouch-2_im}
         \end{subfigure}
         \begin{subfigure}[t]{0.16\textwidth}
                   \includegraphics[width=\textwidth]{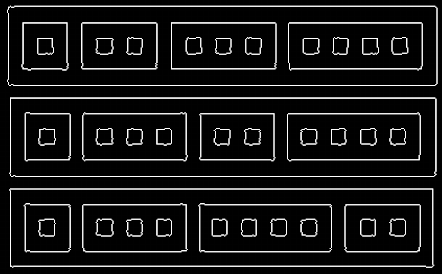}
                   \label{fig:dtouch-2_em}
          \end{subfigure} 
          \begin{subfigure}[t]{0.30\textwidth}
                  \includegraphics[width=\textwidth]{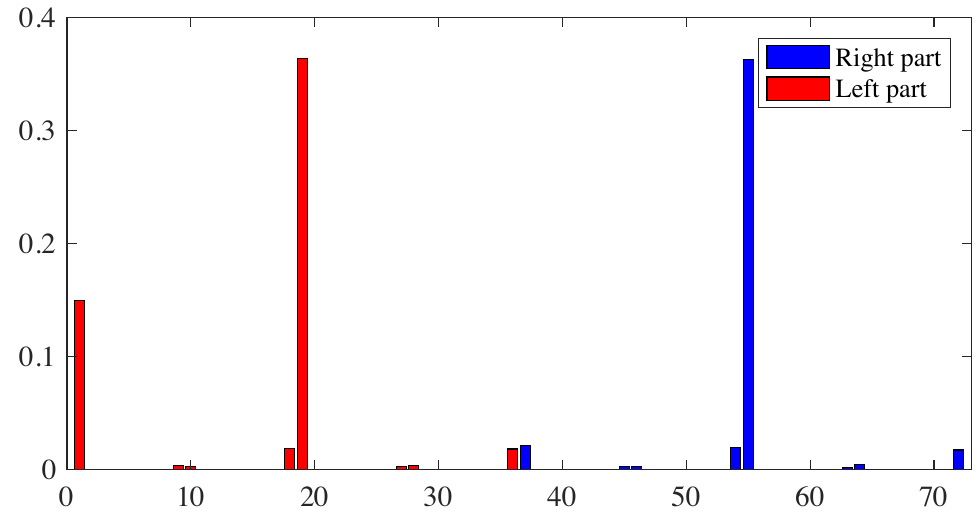}
                  \label{fig:dtouch-2_oh}
         \end{subfigure}
         \begin{subfigure}[t]{0.30\textwidth}
                   \includegraphics[width=\textwidth]{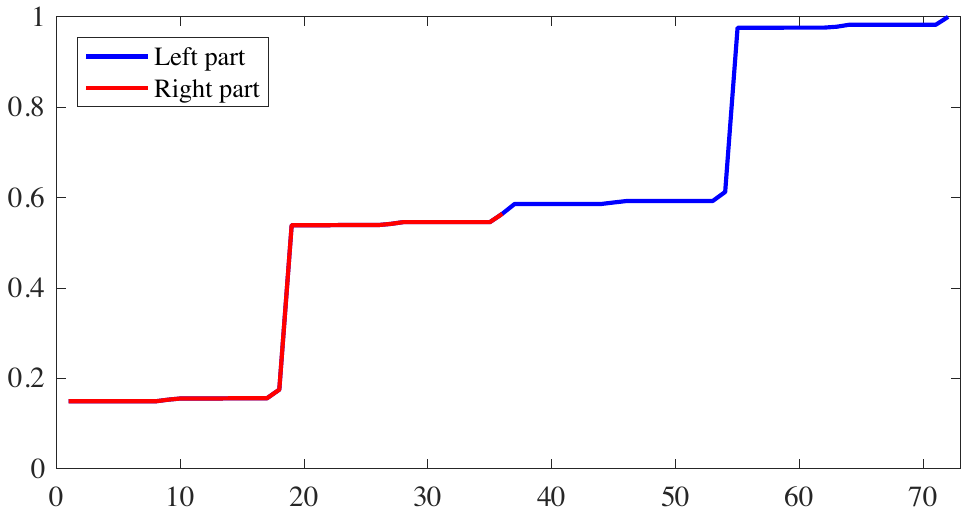}
                   \label{fig:dtouch-2_cum}
          \end{subfigure} 
          
          \begin{subfigure}[t]{0.16\textwidth}
                  \includegraphics[width=\textwidth]{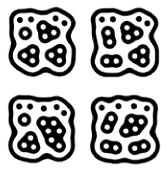}
                  \caption{}
                  \label{fig:occlusion-1_im}
         \end{subfigure}
         \begin{subfigure}[t]{0.16\textwidth}
                   \includegraphics[width=\textwidth]{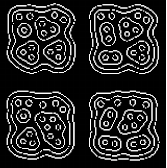}
                   \label{fig:occlusion-1_em}
          \end{subfigure}
           \begin{subfigure}[t]{0.30\textwidth}
                  \includegraphics[width=\textwidth]{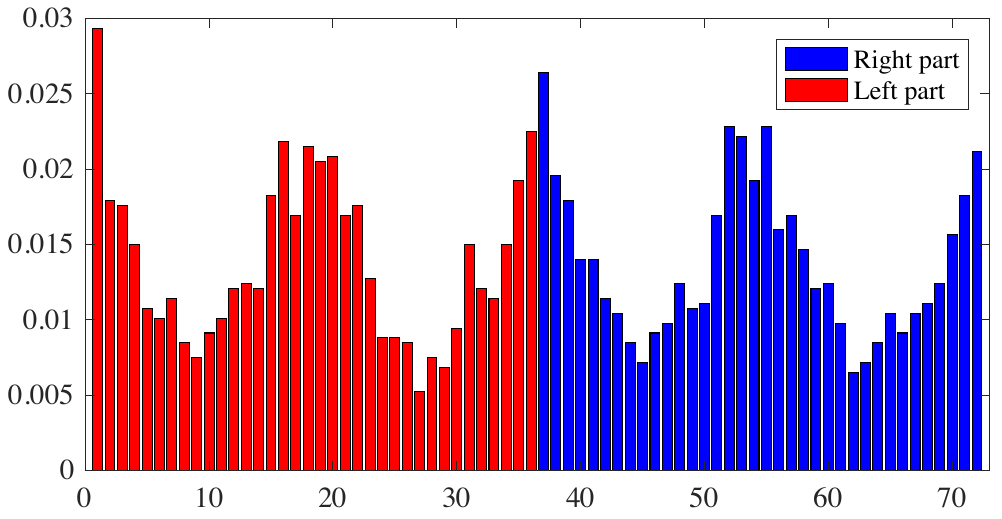}
                  \label{fig:occlusion-1_oh}
         \end{subfigure}
         \begin{subfigure}[t]{0.30\textwidth}
                   \includegraphics[width=\textwidth]{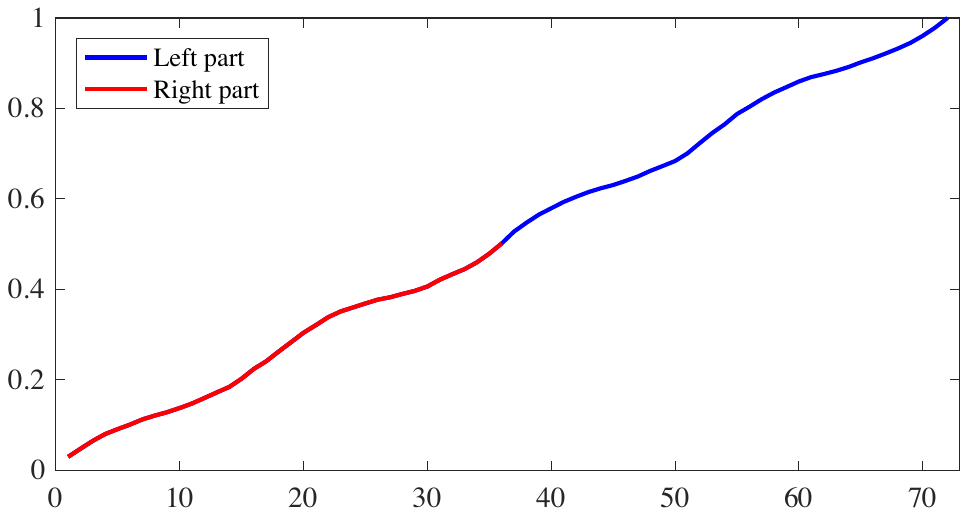}
                   \label{fig:occlusion-1_cum}
          \end{subfigure}
          
           \begin{subfigure}[t]{0.16\textwidth}
                  \includegraphics[width=\textwidth]{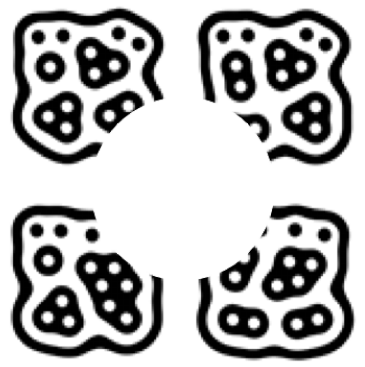}
                  \caption{}
                  \label{fig:occlusion-2_im}
         \end{subfigure}
         \begin{subfigure}[t]{0.16\textwidth}
                   \includegraphics[width=\textwidth]{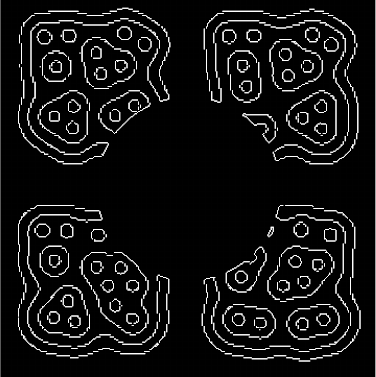}
                   \label{fig:occlusion-2_em}
          \end{subfigure}
          \begin{subfigure}[t]{0.30\textwidth}
                  \includegraphics[width=\textwidth]{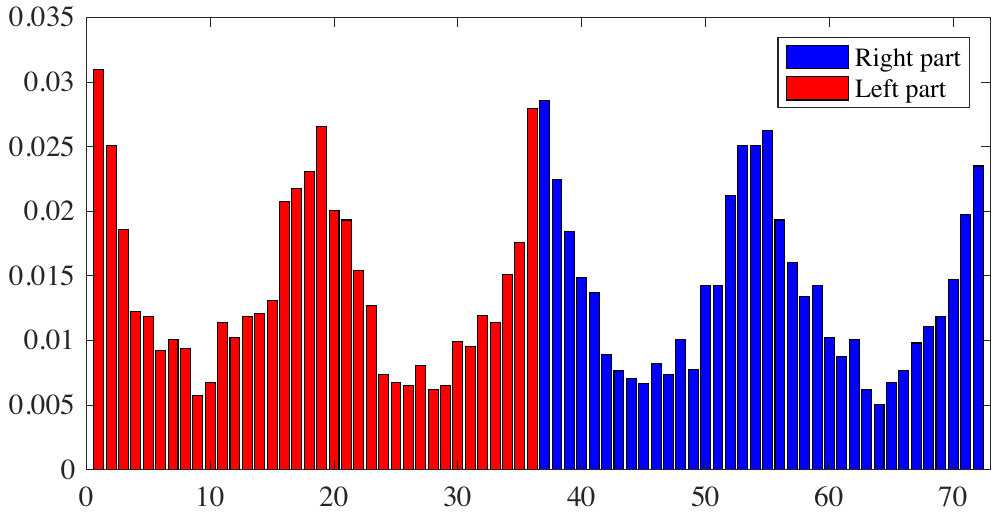}
                  \label{fig:occlusion-2_oh}
         \end{subfigure}
         \begin{subfigure}[t]{0.30\textwidth}
                   \includegraphics[width=\textwidth]{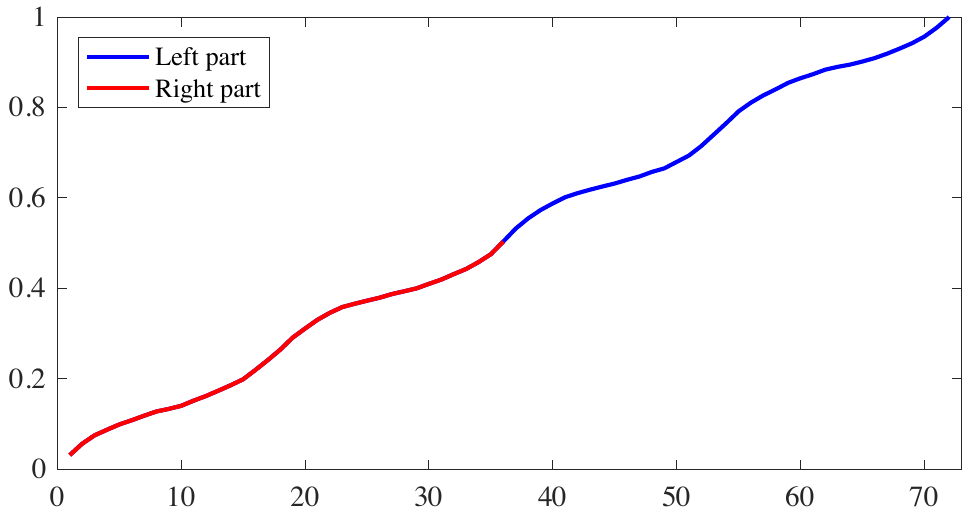}
                   \label{fig:occlusion-2_cum}
          \end{subfigure}
         \caption{
            Example synthetic shapes and their corresponding edge maps, orientation histograms, and cumulative histogram curves. 
            These histograms include 72 bins, uniformly splitting from $-180^\circ$ to $180^\circ$, with $5^\circ$ for each bin. 
            Red shows the left part (bin 1 to 36, i.e., $-180^\circ$ to $0^\circ$) and blue shows the right part (bin 37 to 72, $0^\circ$ to $180^\circ$) of the orientation and cumulative histograms. 
            {\bf Note}: the {\it x}-axes of orientation histograms and cumulative histograms show the number of the histogram bin.}
          \label{fig:shapeStudies}
\end{figure}

\begin{figure}[!t]
    \centering
    \rotatebox{90}{Non-Artcode}\hspace{0.05in}
    \begin{subfigure}[t]{0.16\textwidth}
    \stackinset{c}{}{b}{1.10in}{Real image}{%
        \includegraphics[width=\textwidth]{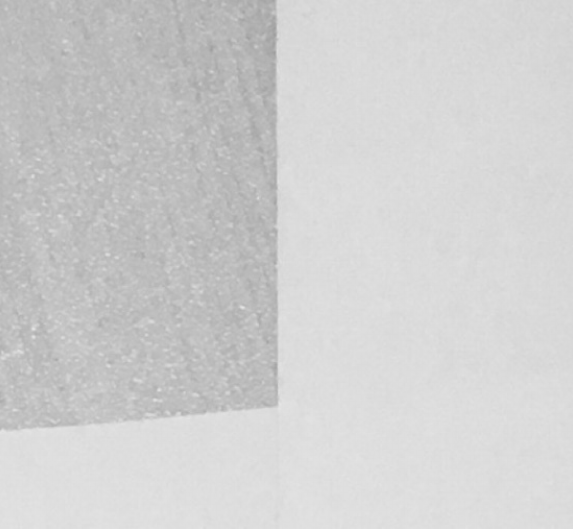}}
        \caption{}
        \label{fig:realImage-1_im}
    \end{subfigure}
    \begin{subfigure}[t]{0.16\textwidth}
    \stackinset{c}{}{b}{1.10in}{Edge map}{%
        \includegraphics[width=\textwidth]{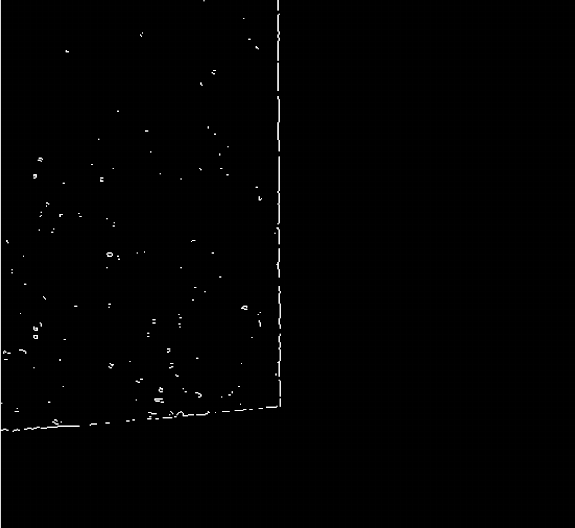}}
        \label{fig:realImage-2_em}
    \end{subfigure} 
    \begin{subfigure}[t]{0.30\textwidth}
    \stackinset{c}{}{b}{1.10in}{Orientation histogram}{%
        \includegraphics[width=\textwidth]{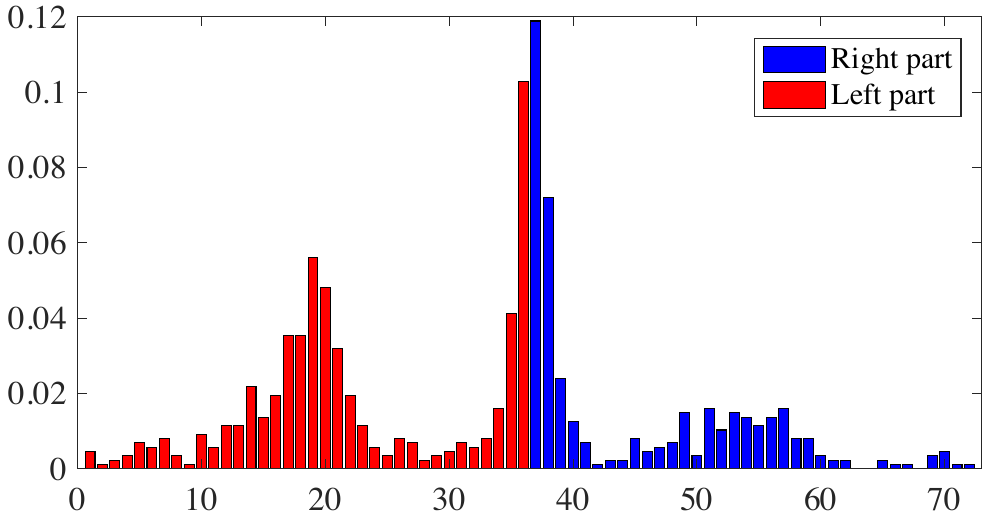}}
        \label{fig:realImage-2_oh}
    \end{subfigure} 
    \begin{subfigure}[t]{0.30\textwidth}
    \stackinset{c}{}{b}{1.10in}{Cumulative histogram curve}{%
        \includegraphics[width=\textwidth]{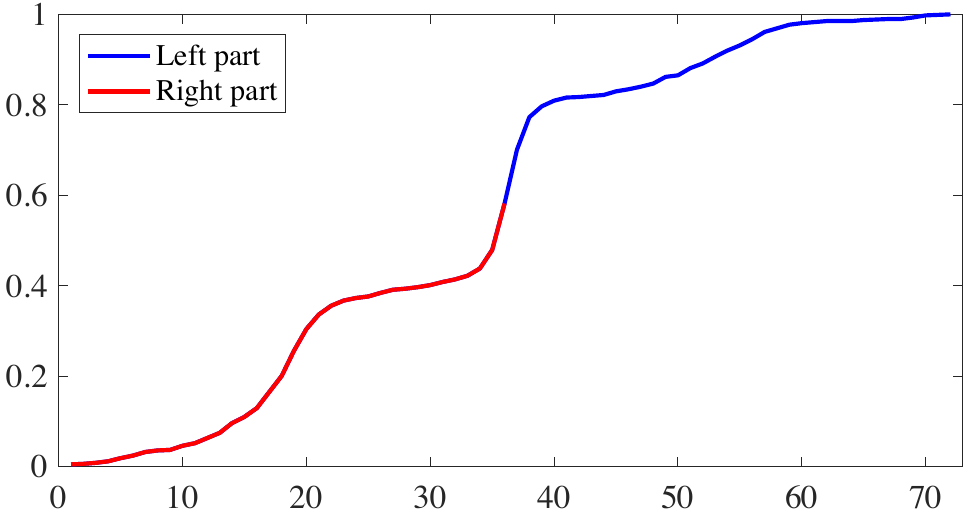}}
        \label{fig:realImage-2_cum}
    \end{subfigure} 
    
    \rotatebox{90}{Non-Artcode}\hspace{0.05in} 
    \begin{subfigure}[t]{0.16\textwidth}
        \includegraphics[width=\textwidth]{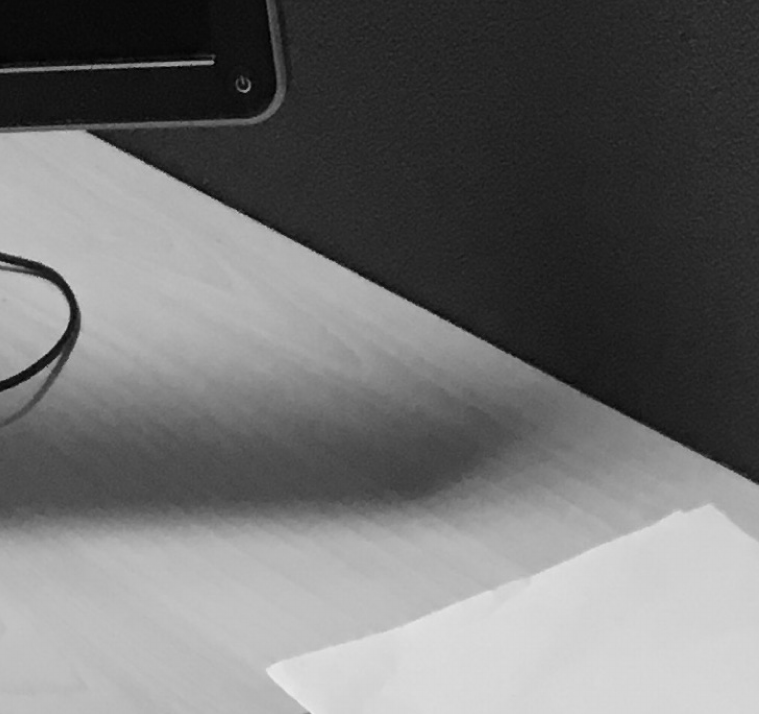}
        \caption{}
        \label{fig:realImage-2_im}
    \end{subfigure}
    \begin{subfigure}[t]{0.16\textwidth}
        \includegraphics[width=\textwidth]{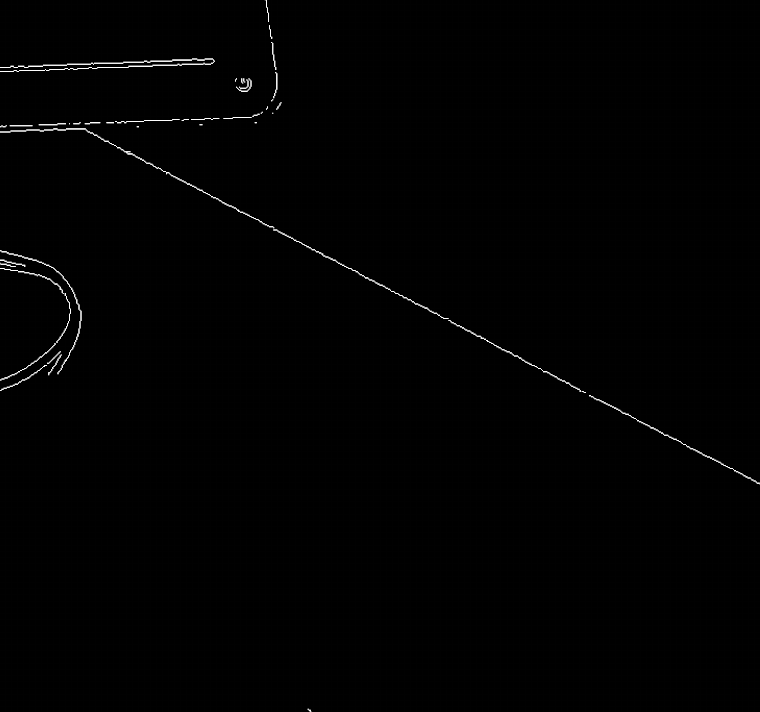}
        \label{fig:realImage-2_em}
    \end{subfigure}
    \begin{subfigure}[t]{0.30\textwidth}
        \includegraphics[width=\textwidth]{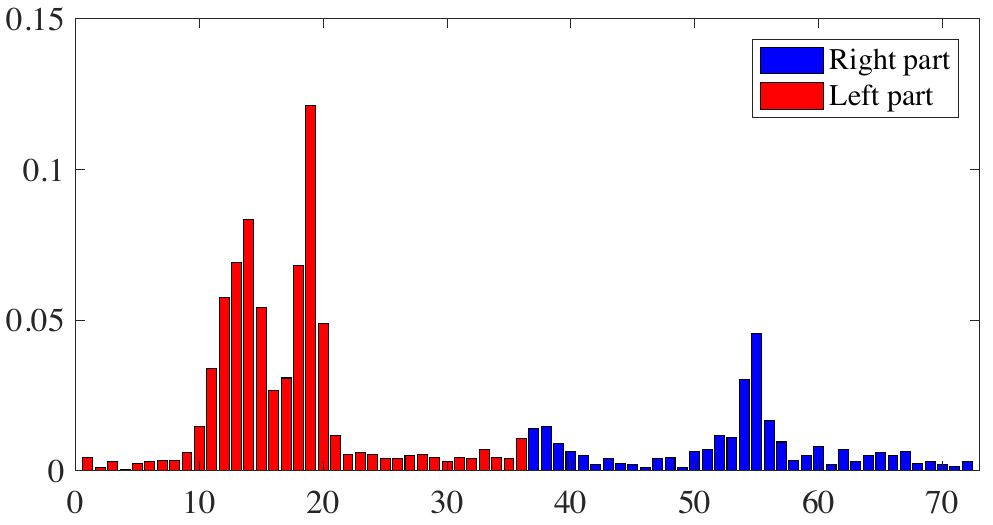}
        \label{fig:realImage-2_oh}
    \end{subfigure}
    \begin{subfigure}[t]{0.30\textwidth}
        \includegraphics[width=\textwidth]{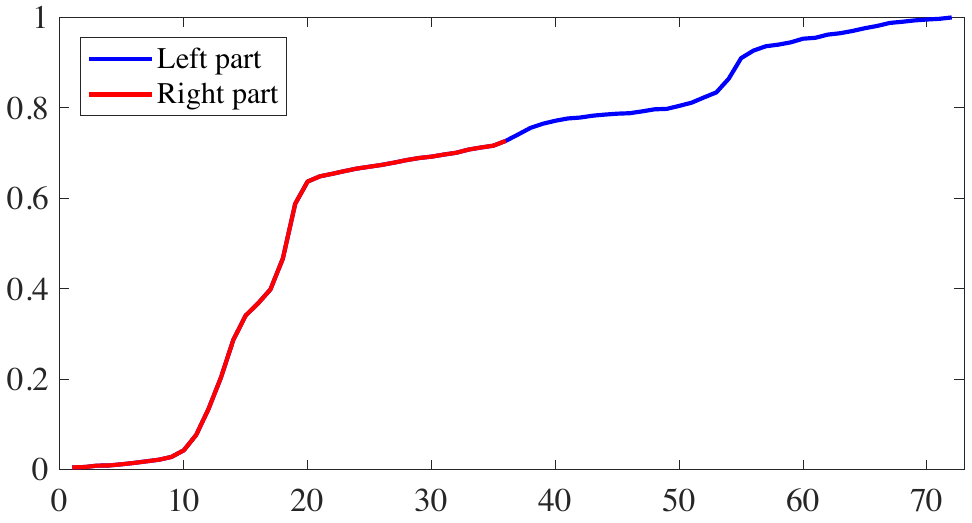}
        \label{fig:realImage-2_cum}
    \end{subfigure}
    
    \rotatebox{90}{Non-Artcode}\hspace{0.05in}
    \begin{subfigure}[t]{0.16\textwidth}
        \includegraphics[width=\textwidth]{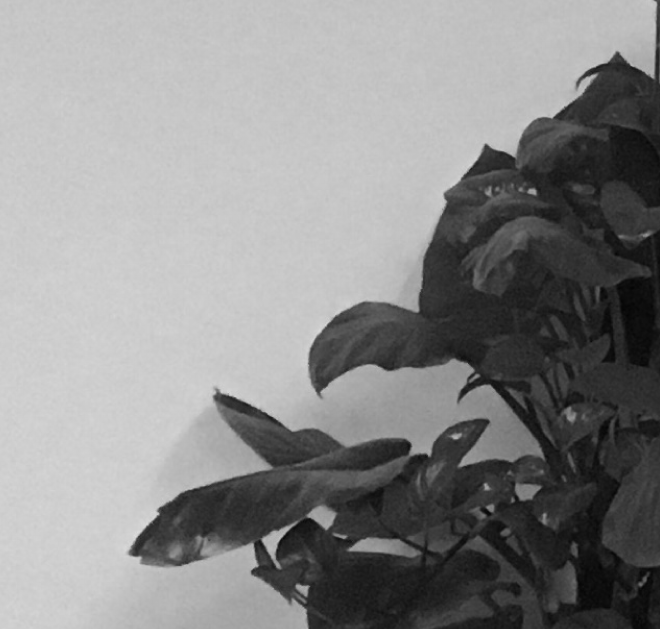}
        \caption{}
        \label{fig:realImage-3_im}
    \end{subfigure}
    \begin{subfigure}[t]{0.16\textwidth}
        \includegraphics[width=\textwidth]{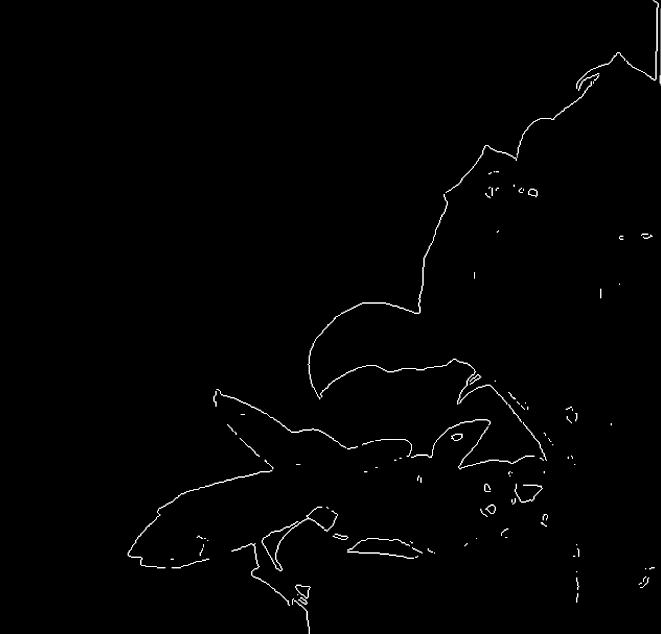}
        \label{fig:realImage-3_em}
    \end{subfigure}
    \begin{subfigure}[t]{0.30\textwidth}
        \includegraphics[width=\textwidth]{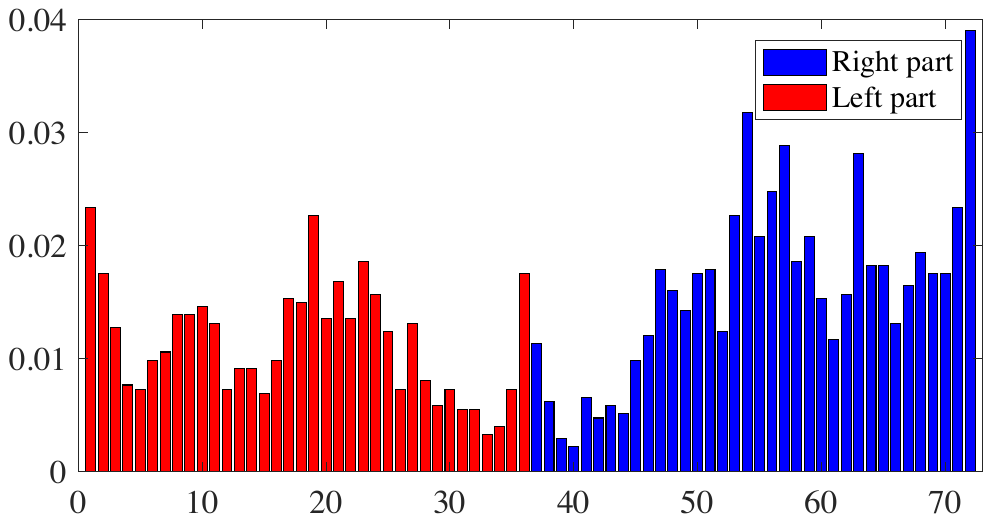}
        \label{fig:realImage-3_oh}
    \end{subfigure}
    \begin{subfigure}[t]{0.30\textwidth}
        \includegraphics[width=\textwidth]{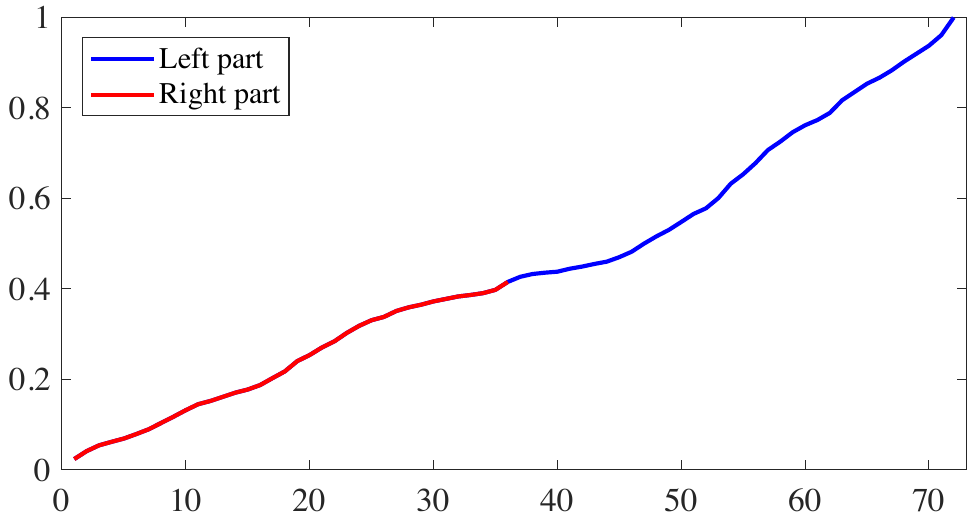}
        \label{fig:realImage-3_cum}
    \end{subfigure}
    
    \rotatebox{90}{Non-Artcode}\hspace{0.05in}
    \begin{subfigure}[t]{0.16\textwidth}
        \includegraphics[width=\textwidth]{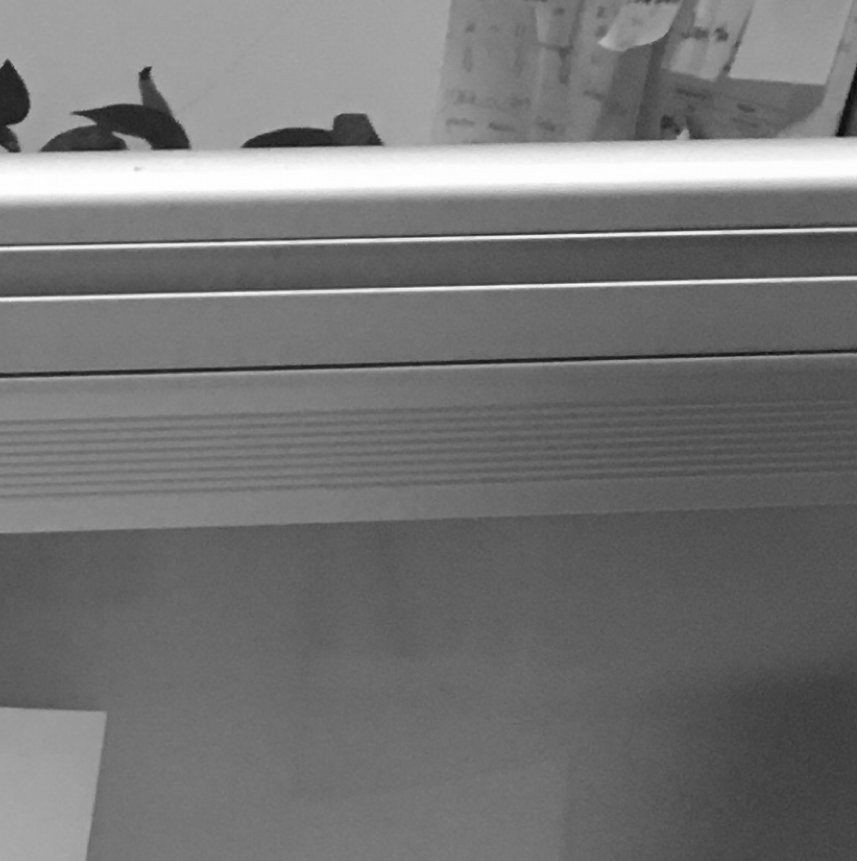}
        \caption{}
        \label{fig:realImage-4_im}
    \end{subfigure}
    \begin{subfigure}[t]{0.16\textwidth}
        \includegraphics[width=\textwidth]{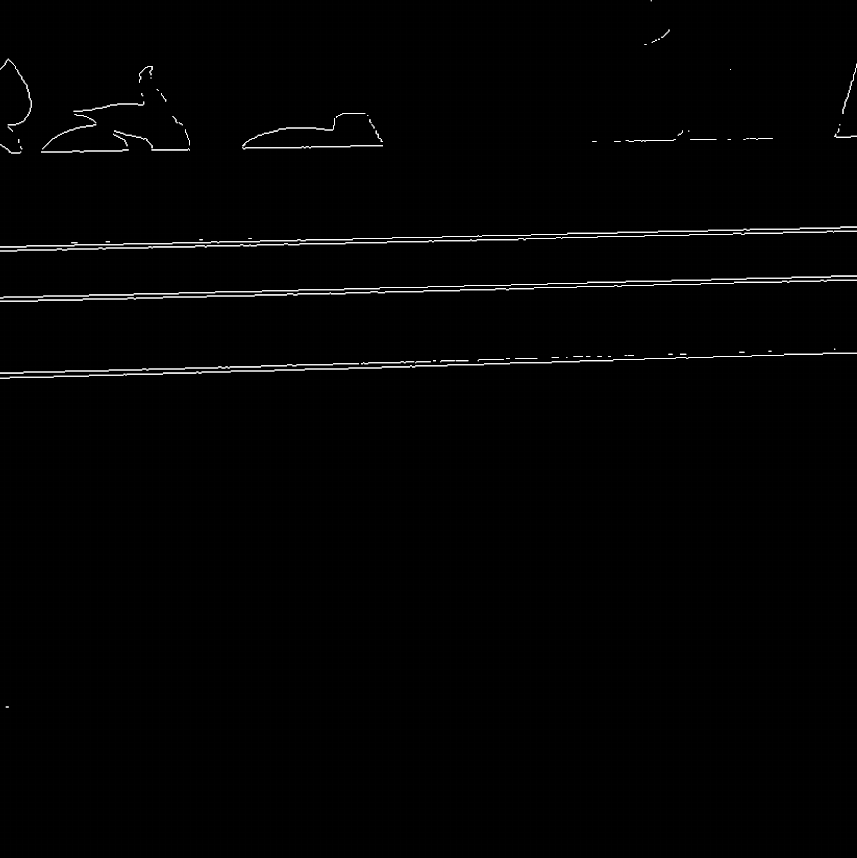}
        \label{fig:realImage-4_em}
    \end{subfigure} 
    \begin{subfigure}[t]{0.30\textwidth}
        \includegraphics[width=\textwidth]{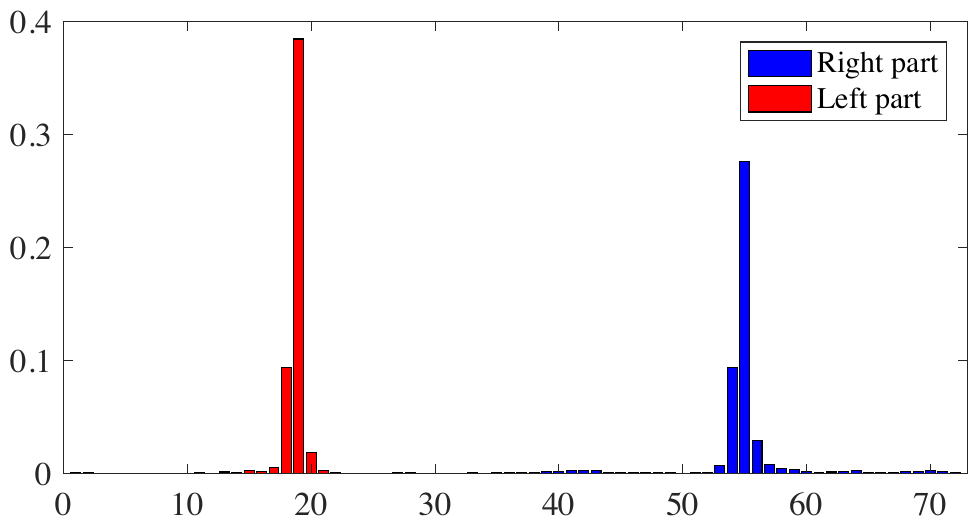}
        \label{fig:realImage-4_oh}
    \end{subfigure}
    \begin{subfigure}[t]{0.30\textwidth}
        \includegraphics[width=\textwidth]{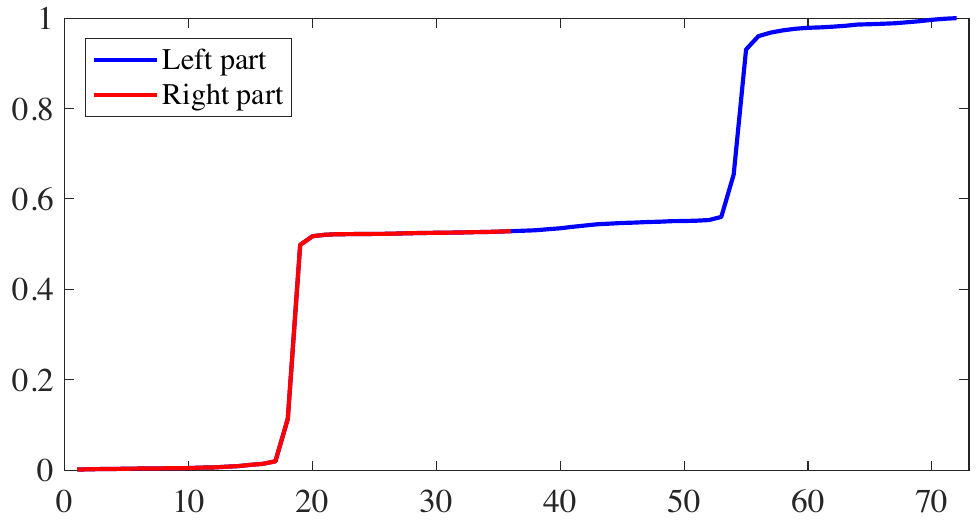}
        \label{fig:realImage-4_cum}
    \end{subfigure} 
    
    \rotatebox{90}{Non-Artcode}\hspace{0.05in}
    \begin{subfigure}[t]{0.16\textwidth}
        \includegraphics[width=\textwidth]{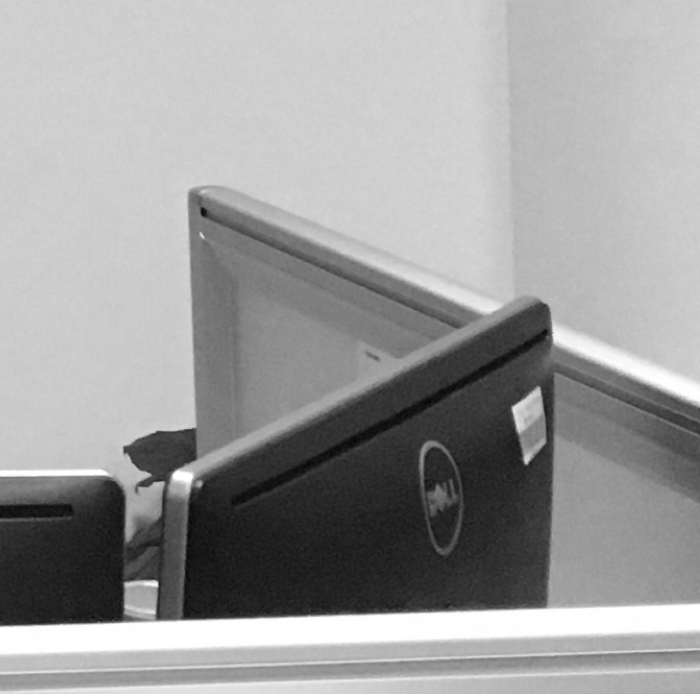}
        \caption{}
        \label{fig:realImage-5_im}
    \end{subfigure}
    \begin{subfigure}[t]{0.16\textwidth}
        \includegraphics[width=\textwidth]{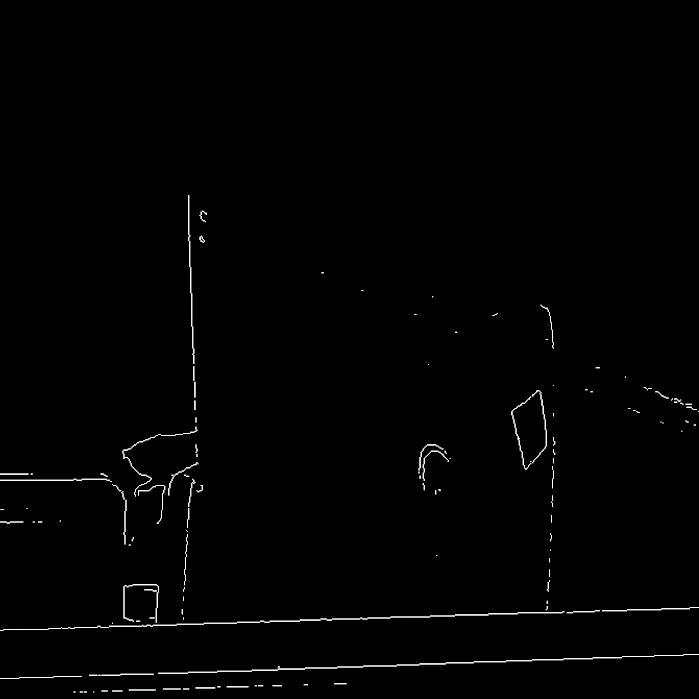}
        \label{fig:realImage-5_em}
    \end{subfigure}
    \begin{subfigure}[t]{0.30\textwidth}
        \includegraphics[width=\textwidth]{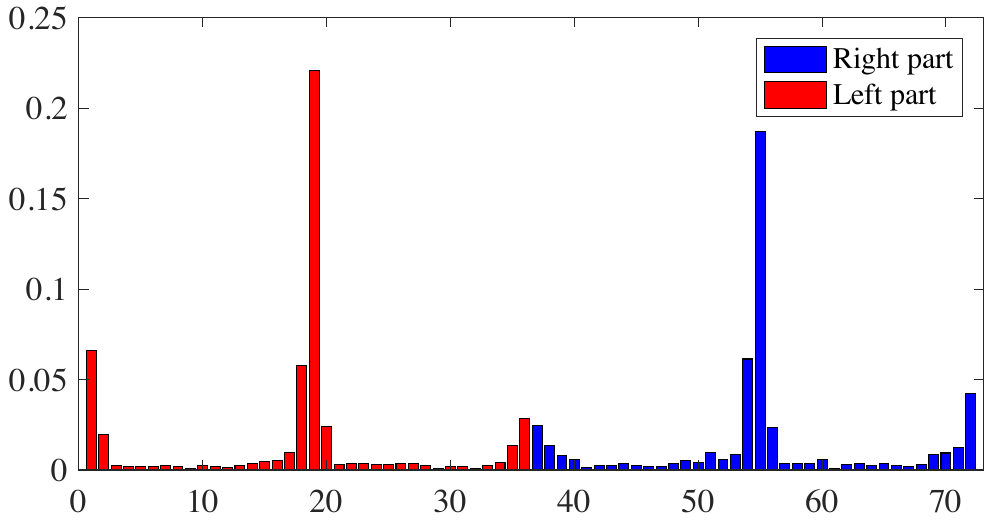}
        \label{fig:realImage-5_oh}
    \end{subfigure}
    \begin{subfigure}[t]{0.30\textwidth}
        \includegraphics[width=\textwidth]{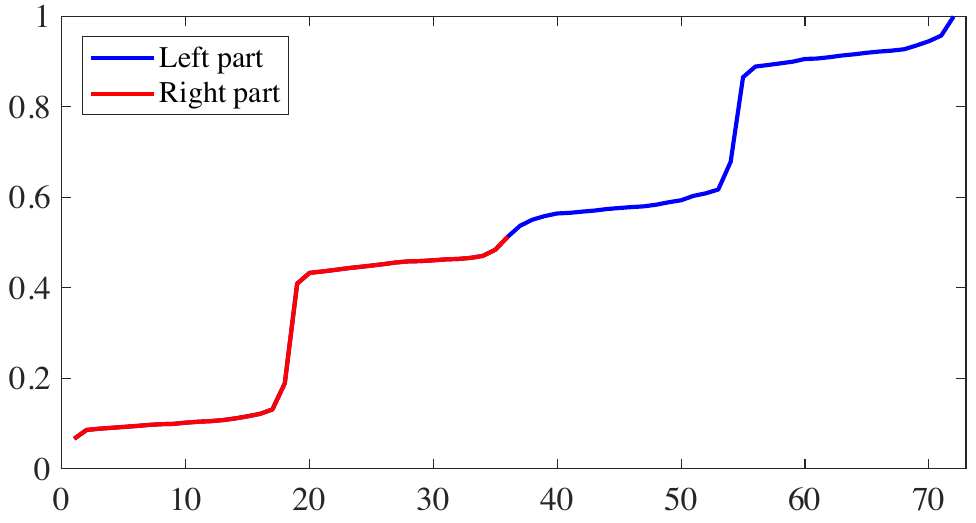}
        \label{fig:realImage-5_cum}
    \end{subfigure}
    
    \rotatebox{90}{Non-Artcode}\hspace{0.05in}
    \begin{subfigure}[t]{0.16\textwidth}
        \includegraphics[width=\textwidth]{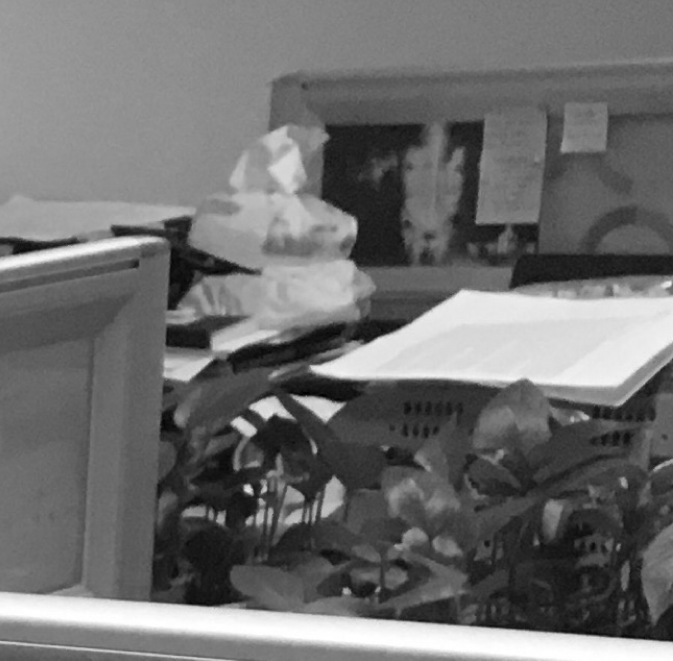}
        \caption{}
        \label{fig:realImage-6_im}
    \end{subfigure}
    \begin{subfigure}[t]{0.16\textwidth}
        \includegraphics[width=\textwidth]{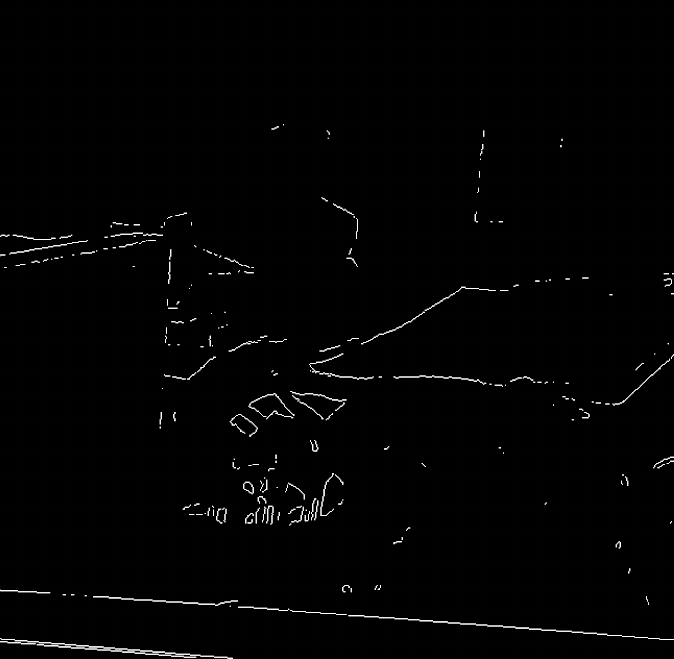}
        \label{fig:realImage-6_em}
    \end{subfigure}
    \begin{subfigure}[t]{0.30\textwidth}
        \includegraphics[width=\textwidth]{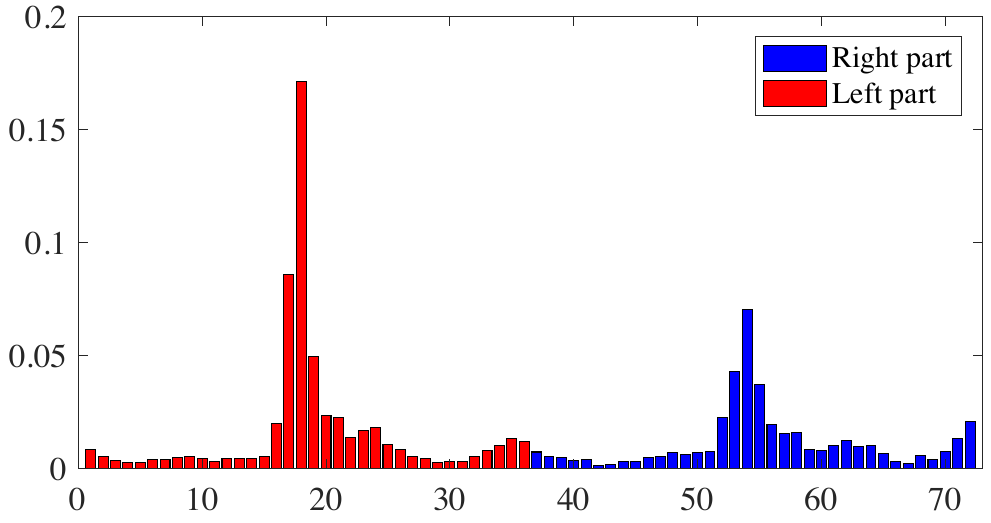}
        \label{fig:realImage-6_oh}
    \end{subfigure}
    \begin{subfigure}[t]{0.30\textwidth}
        \includegraphics[width=\textwidth]{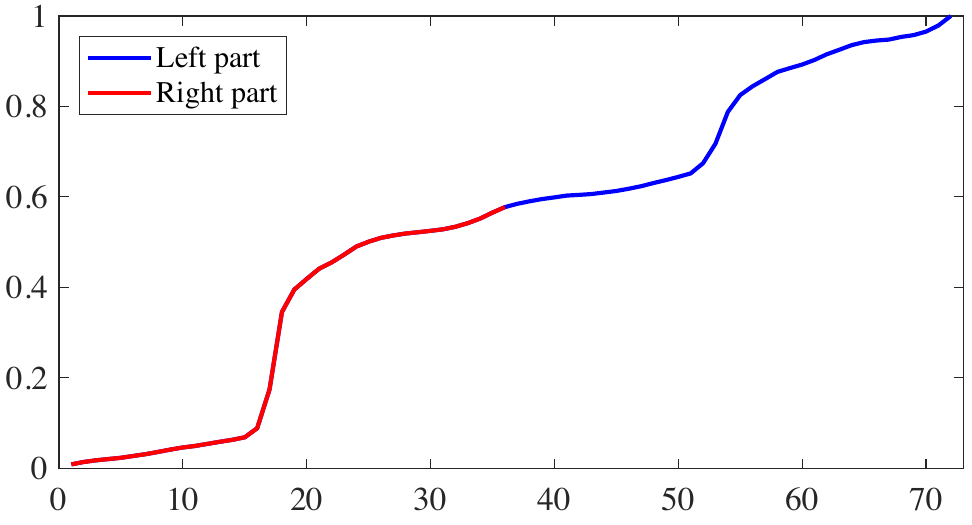}
        \label{fig:realImage-6_cum}
    \end{subfigure}
\end{figure}

\begin{figure}[!t]
    \ContinuedFloat
    \centering
    \rotatebox{90}{Artcode-like}\hspace{0.05in} 
    \begin{subfigure}[t]{0.16\textwidth}
    \stackinset{c}{}{b}{1.10in}{Real image}{%
        \includegraphics[width=\textwidth]{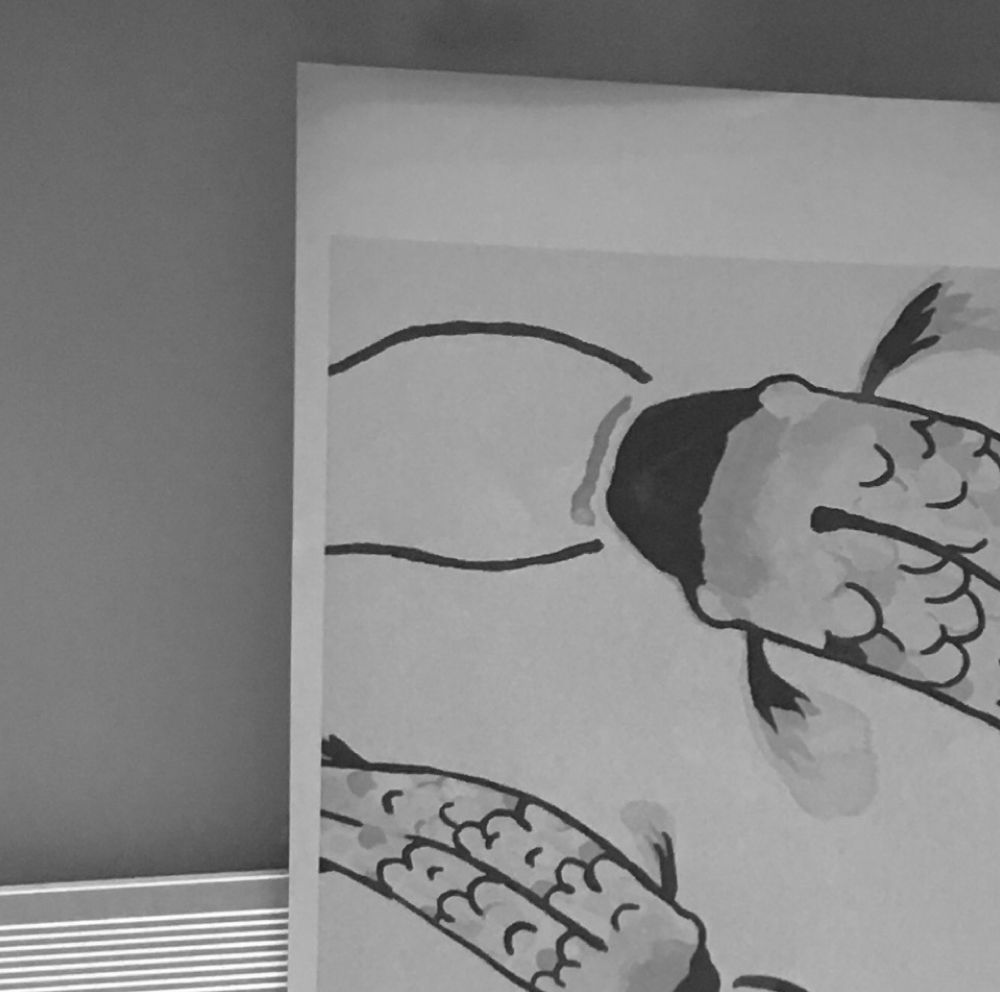}}
        \caption{}
        \label{fig:realImage-7_im}
    \end{subfigure}
    \begin{subfigure}[t]{0.16\textwidth}
    \stackinset{c}{}{b}{1.10in}{Edge map}{%
        \includegraphics[width=\textwidth]{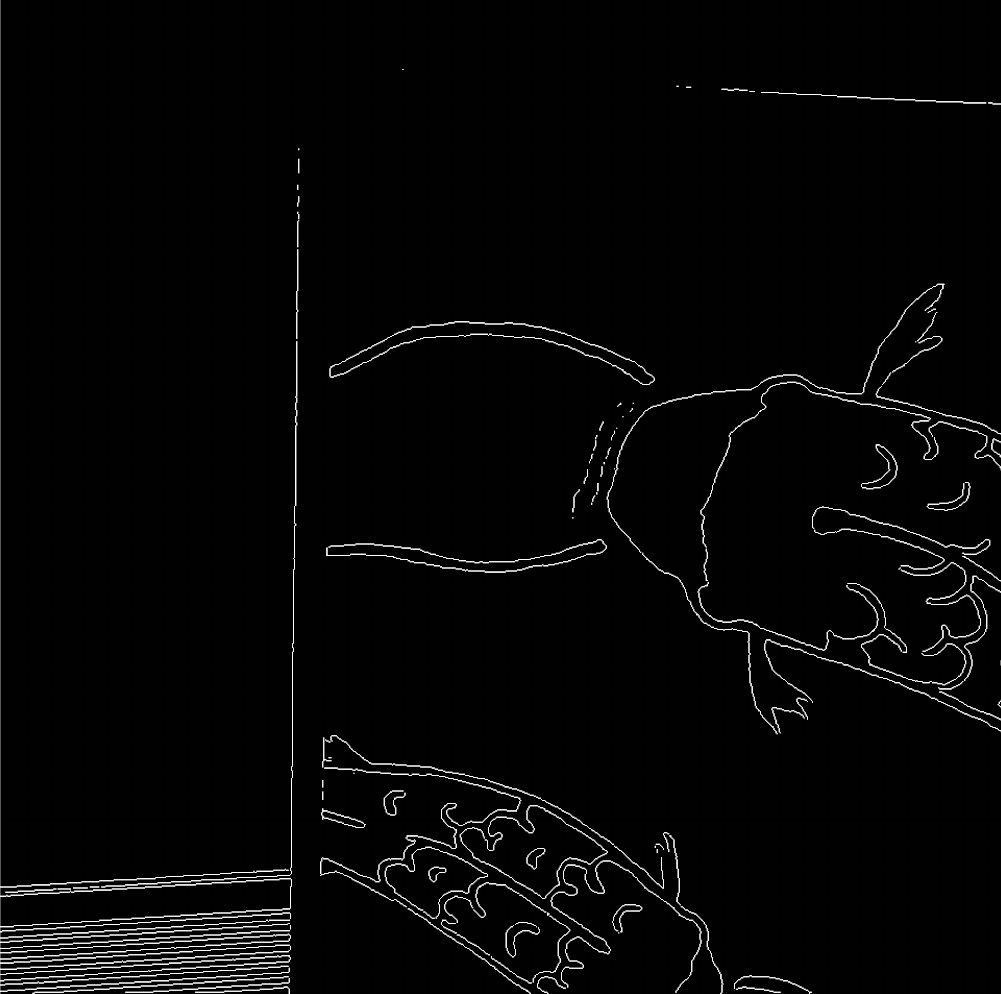}}
        \label{fig:realImage-7_em}
    \end{subfigure}
    \begin{subfigure}[t]{0.30\textwidth}
    \stackinset{c}{}{b}{1.10in}{Orientation histogram}{%
        \includegraphics[width=\textwidth]{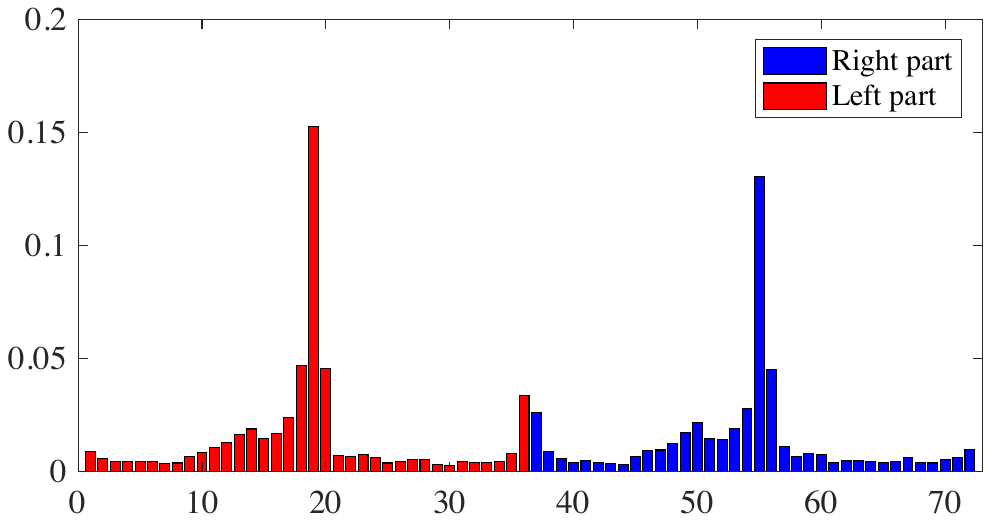}}
        \label{fig:realImage-7_oh}
    \end{subfigure}
    \begin{subfigure}[t]{0.30\textwidth}
    \stackinset{c}{}{b}{1.10in}{Cumulative histogram curve}{%
        \includegraphics[width=\textwidth]{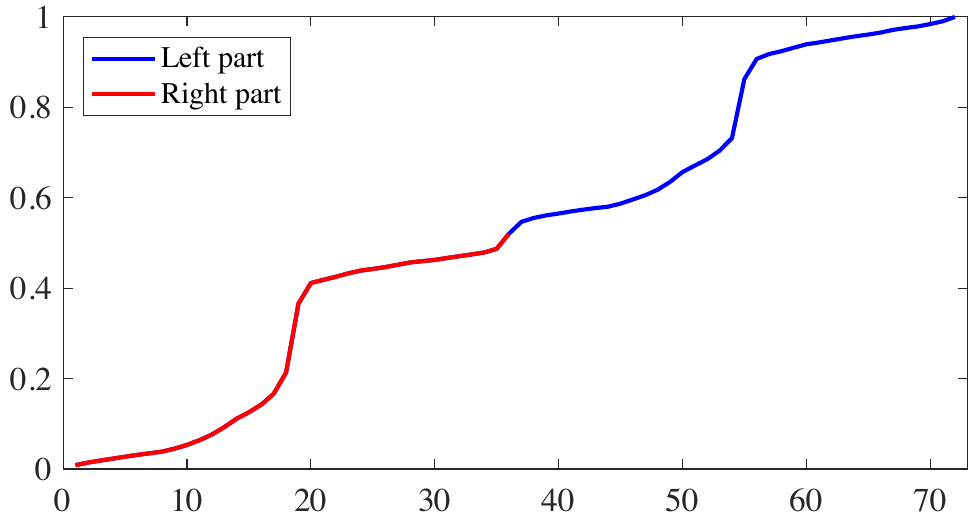}}
        \label{fig:realImage-7_cum}
    \end{subfigure}
    
    \rotatebox{90}{Artcode-like}\hspace{0.05in} 
    \begin{subfigure}[t]{0.16\textwidth}
        \includegraphics[width=\textwidth]{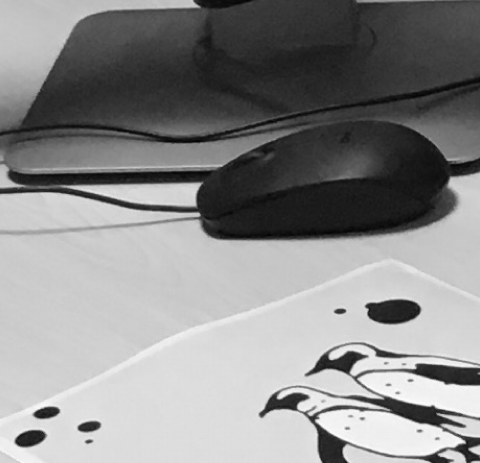}
        \caption{}
        \label{fig:realImage-8_im}
    \end{subfigure}
    \begin{subfigure}[t]{0.16\textwidth}
        \includegraphics[width=\textwidth]{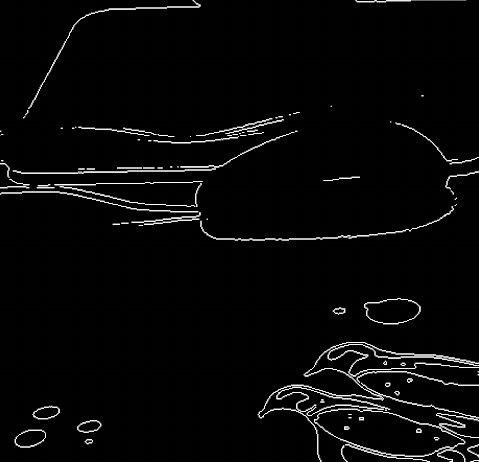}
        \label{fig:realImage-9_em}
    \end{subfigure}
    \begin{subfigure}[t]{0.30\textwidth}
        \includegraphics[width=\textwidth]{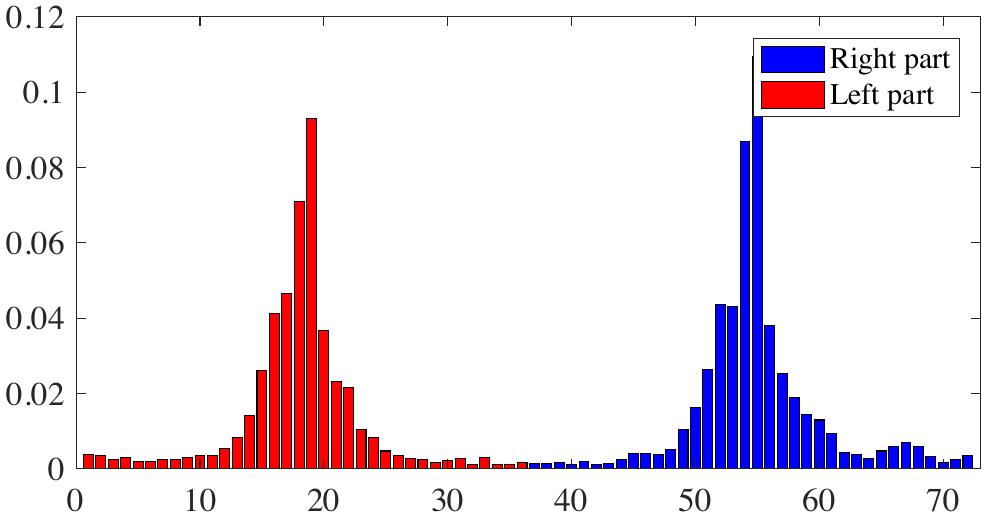}
        \label{fig:realImage-8_oh}
    \end{subfigure}
    \begin{subfigure}[t]{0.30\textwidth}
        \includegraphics[width=\textwidth]{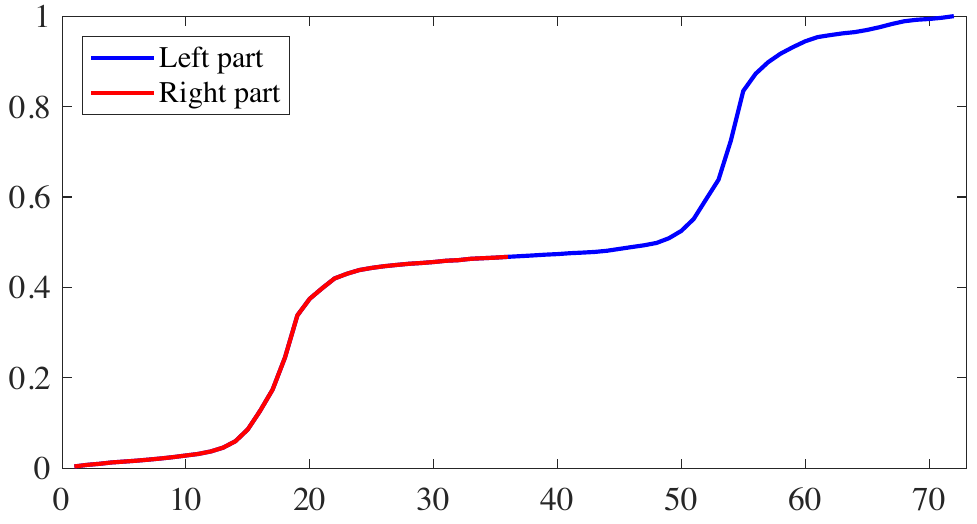}
        \label{fig:realImage-9_cum}
    \end{subfigure}
    
    \rotatebox{90}{Artcode-like}\hspace{0.05in} 
    \begin{subfigure}[t]{0.16\textwidth}
        \includegraphics[width=\textwidth]{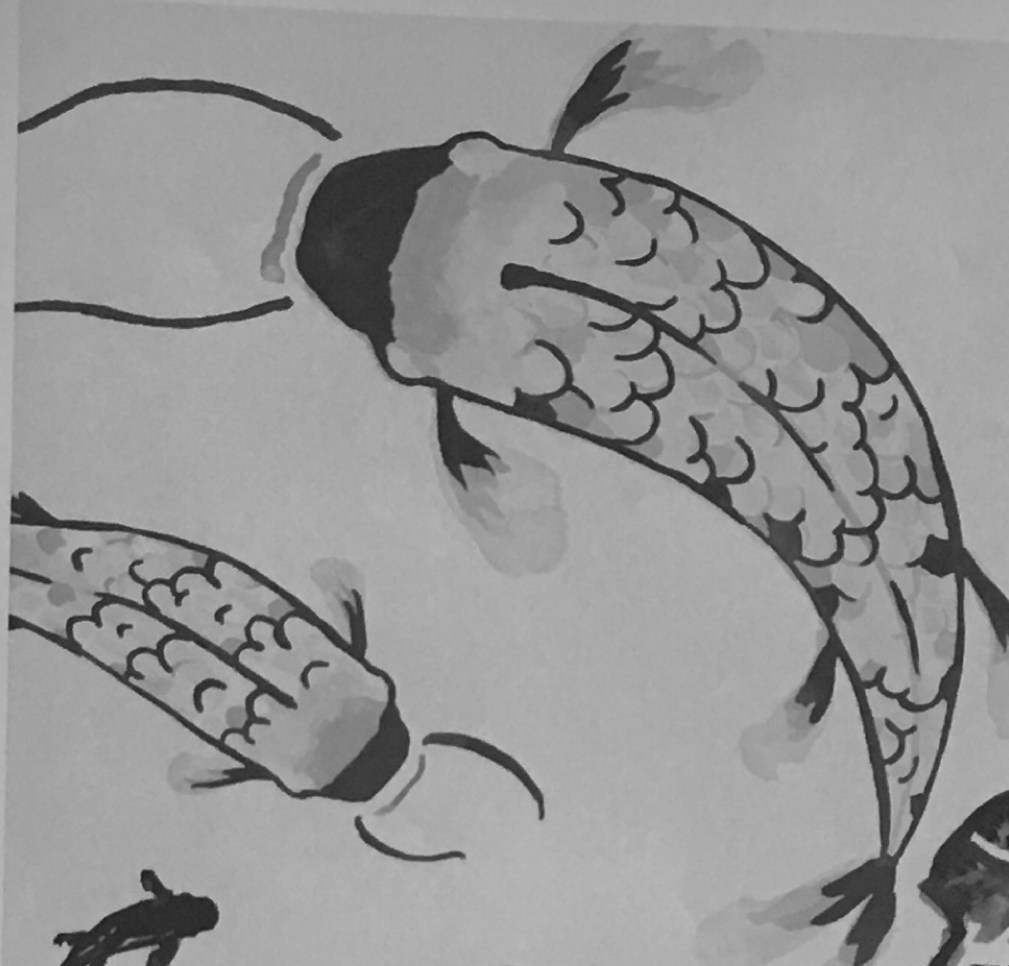}
        \caption{}
        \label{fig:realImage-9_im}
    \end{subfigure}
    \begin{subfigure}[t]{0.16\textwidth}
        \includegraphics[width=\textwidth]{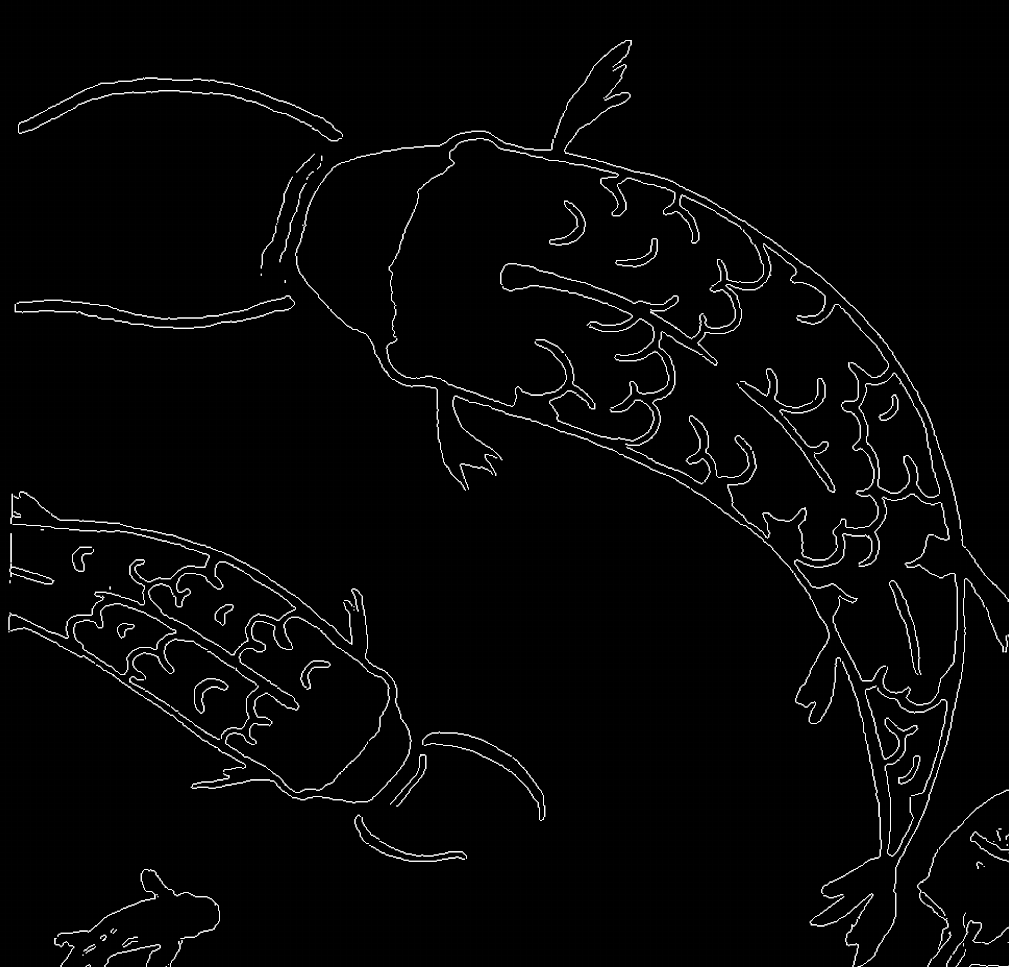}
        \label{fig:realImage-9_em}
    \end{subfigure}
    \begin{subfigure}[t]{0.30\textwidth}
        \includegraphics[width=\textwidth]{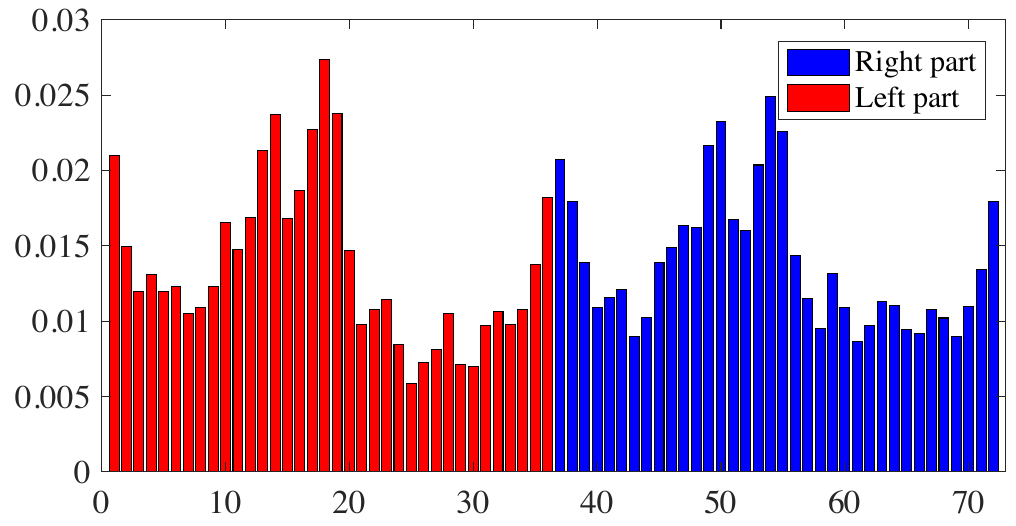}
        \label{fig:realImage-9_oh}
    \end{subfigure}
    \begin{subfigure}[t]{0.30\textwidth}
        \includegraphics[width=\textwidth]{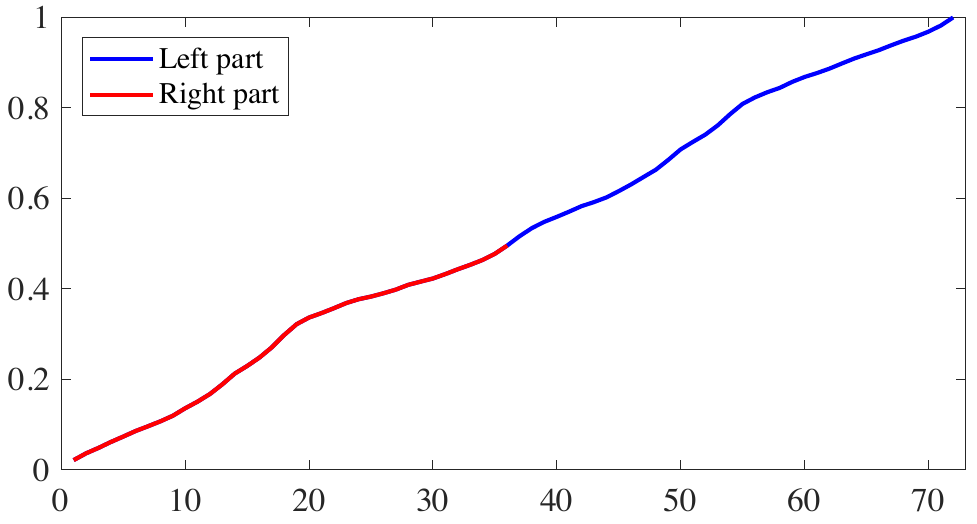}
        \label{fig:realImage-9_cum}
    \end{subfigure}
    
    \rotatebox{90}{Artcode-like}\hspace{0.05in} 
    \begin{subfigure}[t]{0.16\textwidth}
        \includegraphics[width=\textwidth]{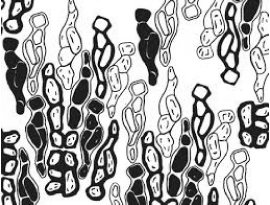}
        \caption{}
        \label{fig:realImage-10_im}
    \end{subfigure}
    \begin{subfigure}[t]{0.16\textwidth}
        \includegraphics[width=\textwidth]{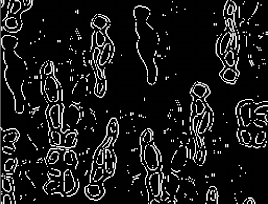}
        \label{fig:realImage-10_em}
    \end{subfigure}
    \begin{subfigure}[t]{0.30\textwidth}
        \includegraphics[width=\textwidth]{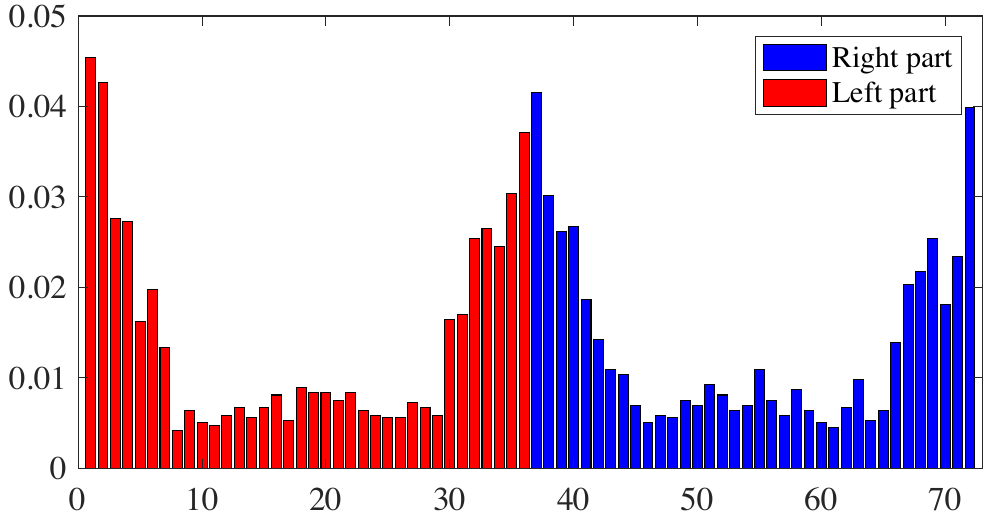}
        \label{fig:realImage-10_oh}
    \end{subfigure}
    \begin{subfigure}[t]{0.30\textwidth}
        \includegraphics[width=\textwidth]{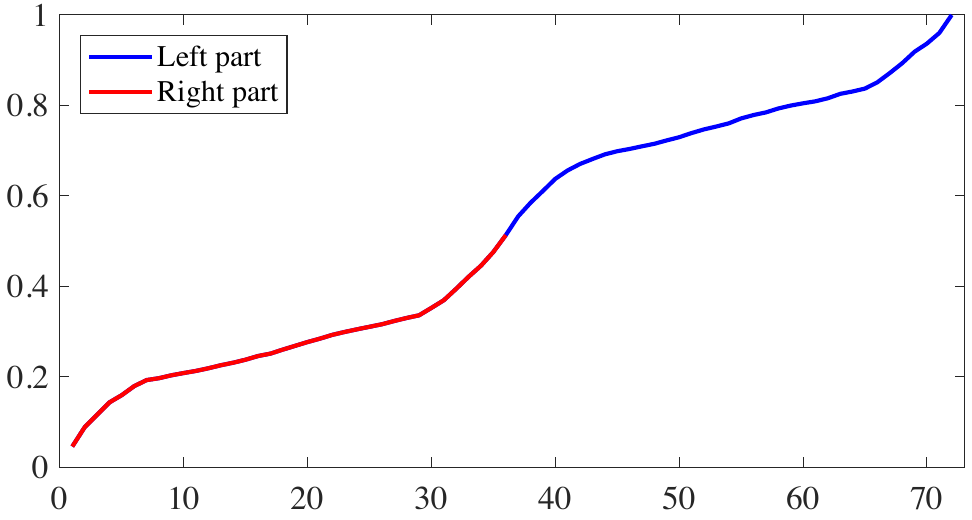}
        \label{fig:realImage-10_cum}
    \end{subfigure}
    
    \rotatebox{90}{Artcode-like}\hspace{0.05in} 
    \begin{subfigure}[t]{0.16\textwidth}
        \includegraphics[width=\textwidth]{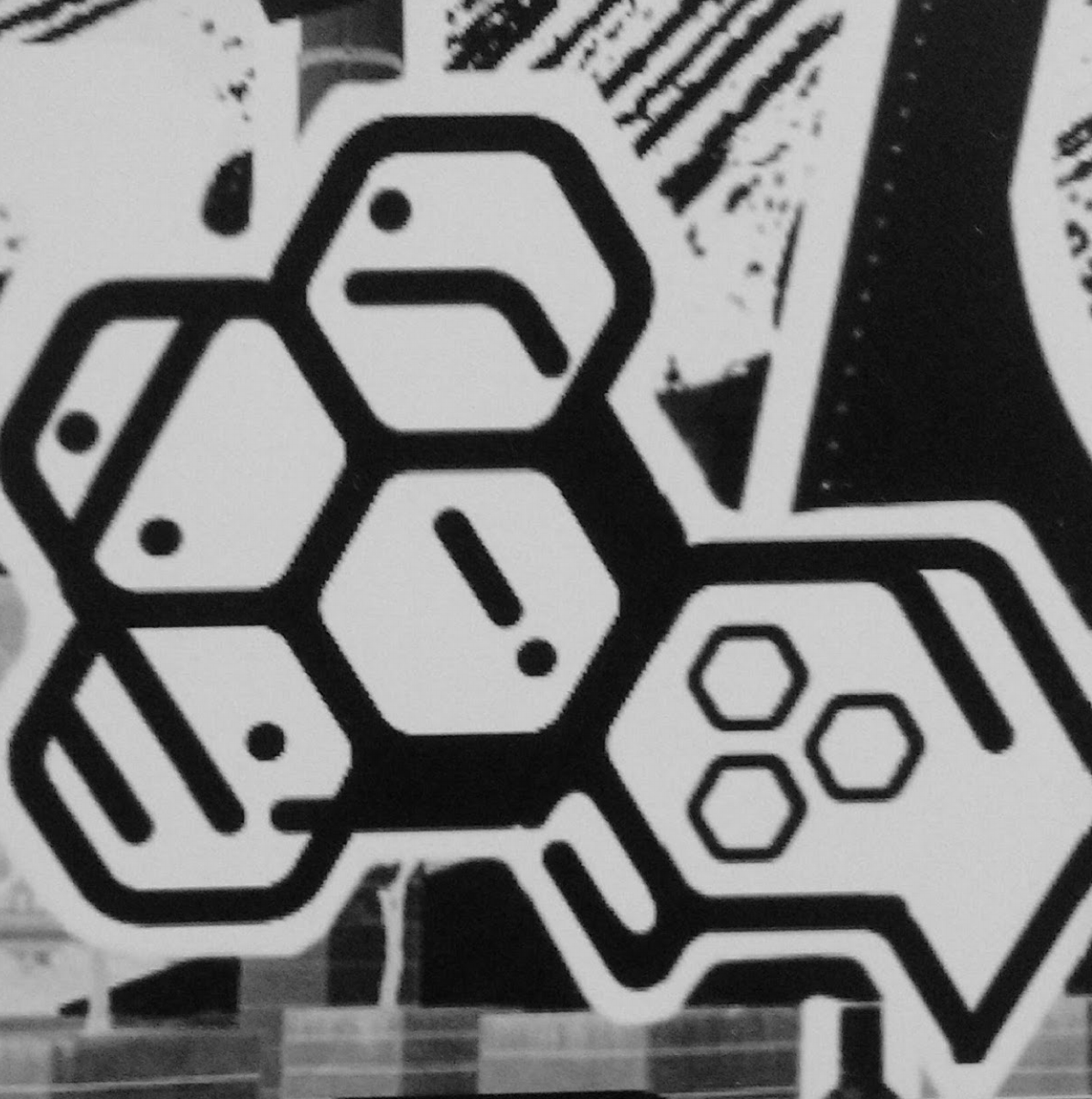}
        \caption{}
        \label{fig:realImage-11_im}
    \end{subfigure}
    \begin{subfigure}[t]{0.16\textwidth}
        \includegraphics[width=\textwidth]{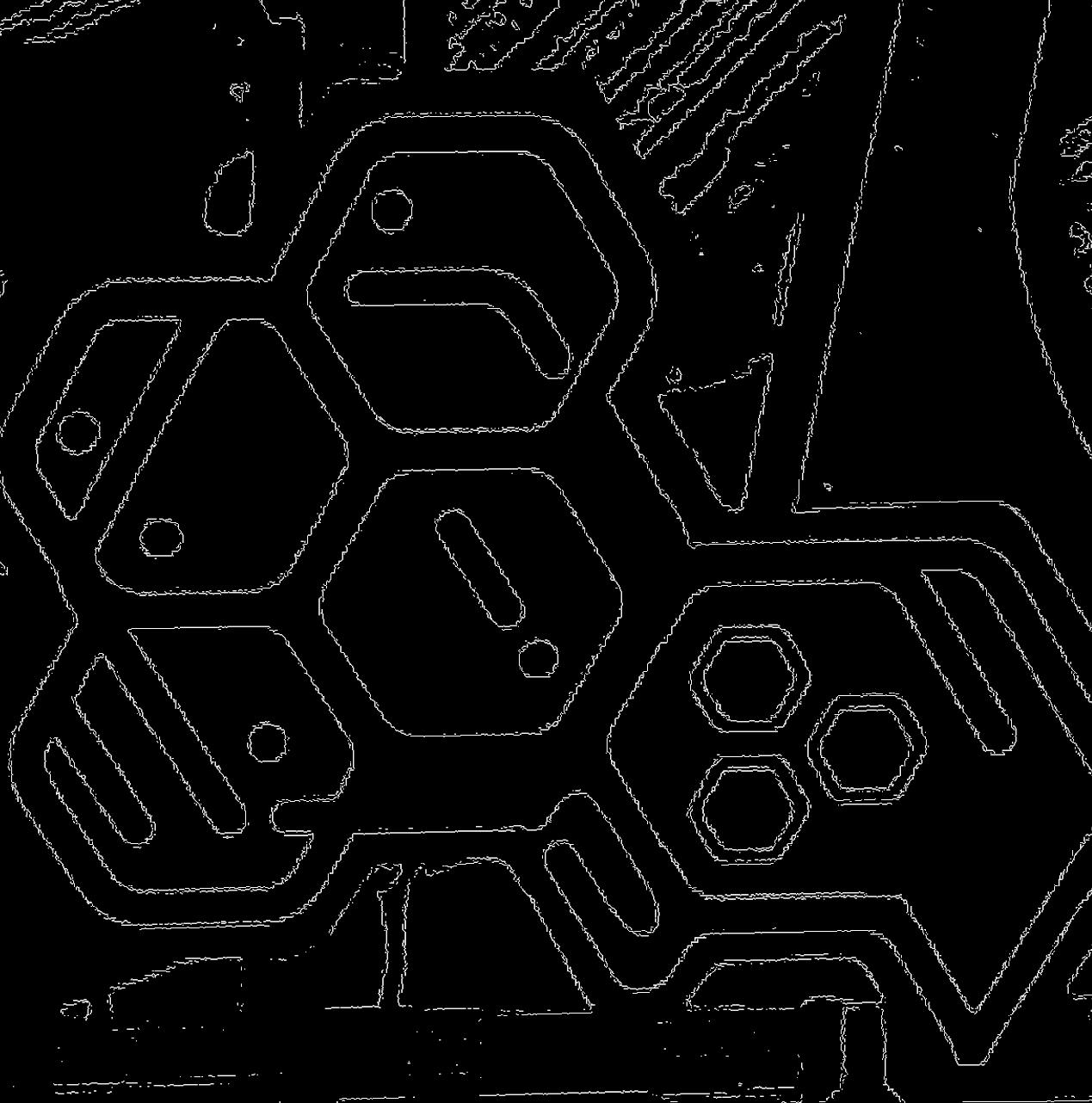}
        \label{fig:realImage-11_em}
    \end{subfigure} 
    \begin{subfigure}[t]{0.30\textwidth}
        \includegraphics[width=\textwidth]{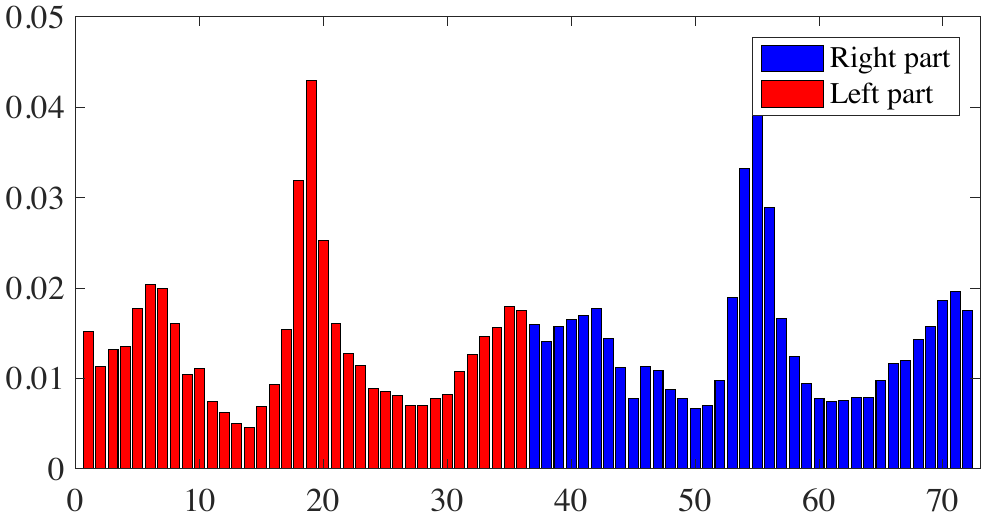}
        \label{fig:realImage-11_oh}
    \end{subfigure}
    \begin{subfigure}[t]{0.30\textwidth}
        \includegraphics[width=\textwidth]{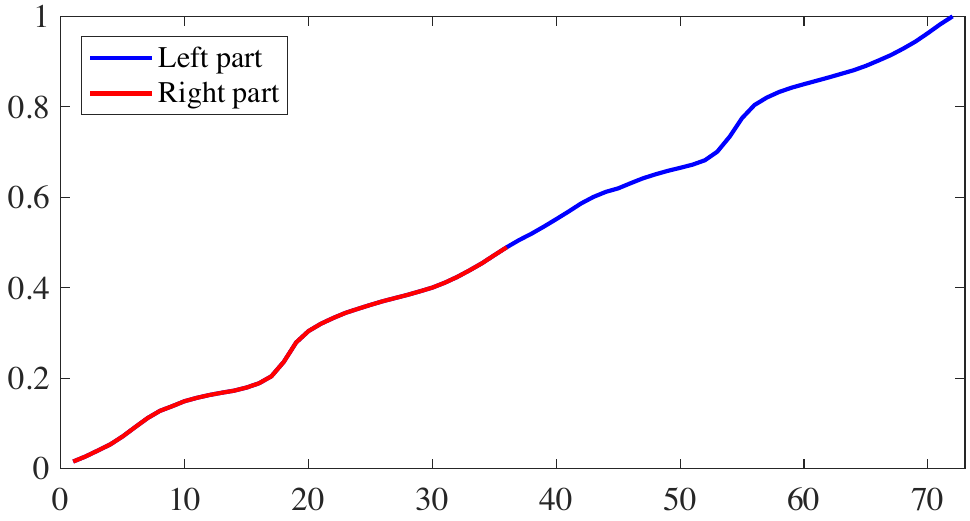}
        \label{fig:realImage-11_cum}
    \end{subfigure} 
     
    \rotatebox{90}{Artcode-like}\hspace{0.05in} 
    \begin{subfigure}[t]{0.16\textwidth}
        \includegraphics[width=\textwidth]{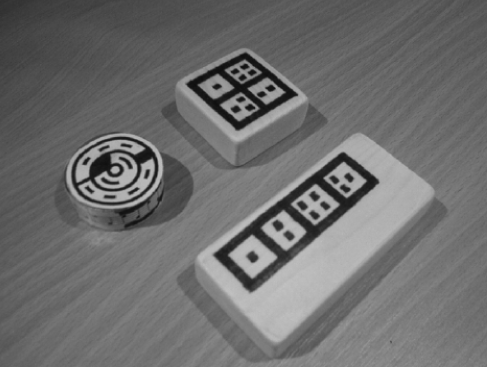}
        \caption{}
        \label{fig:realImage-12_im}
    \end{subfigure}
    \begin{subfigure}[t]{0.16\textwidth}
        \includegraphics[width=\textwidth]{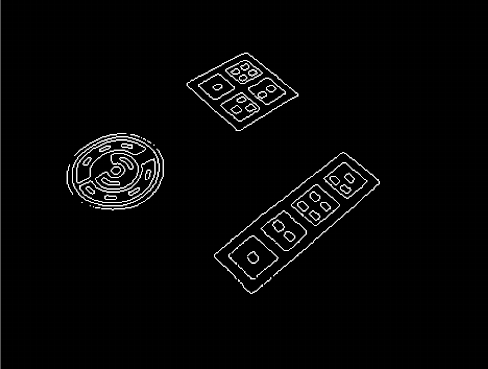}
        \label{fig:realImage-12_em}
    \end{subfigure}   
    \begin{subfigure}[t]{0.30\textwidth}
        \includegraphics[width=\textwidth]{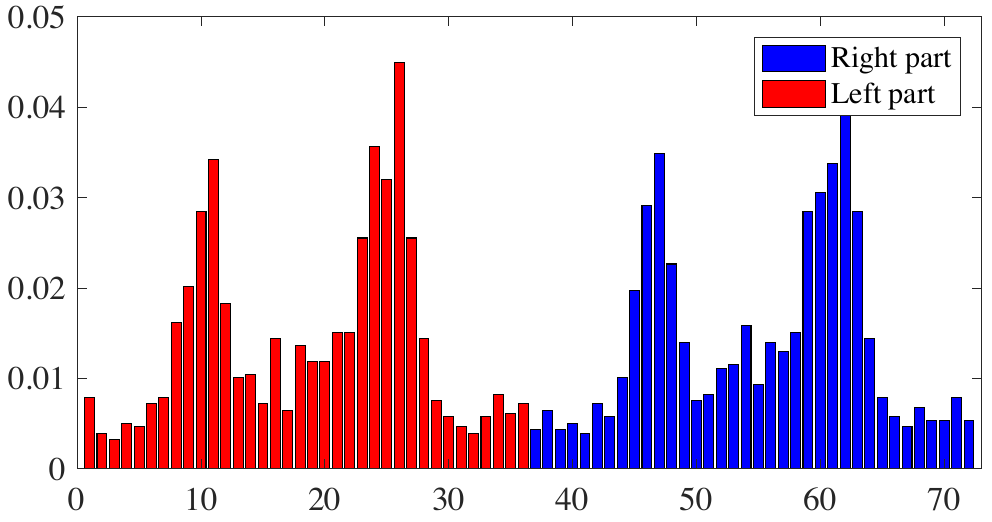}
        \label{fig:realImage-12_oh}
    \end{subfigure}
    \begin{subfigure}[t]{0.30\textwidth}
        \includegraphics[width=\textwidth]{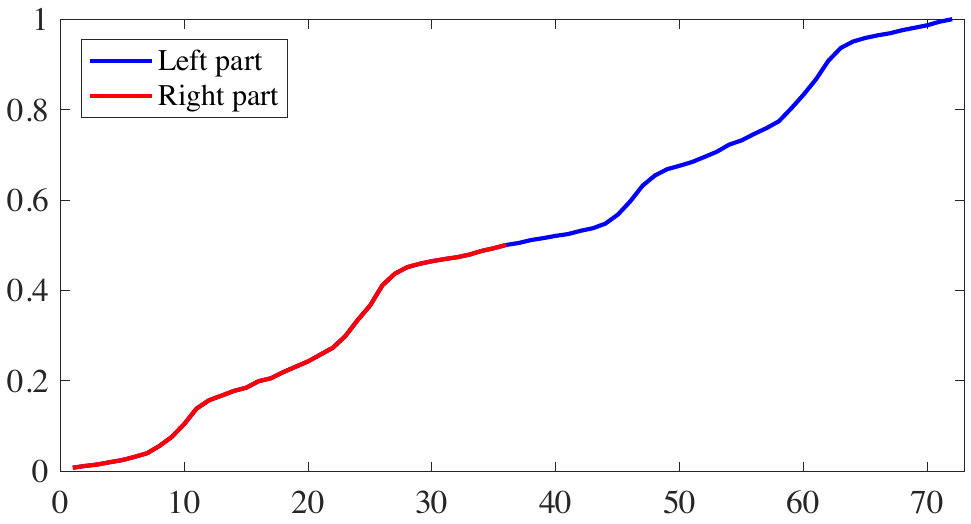}
        \label{fig:realImage-12_cum}
    \end{subfigure}           
    \caption{
        Example real images and their corresponding edge maps, orientation histograms, and cumulative histogram curves. 
        These histograms have 72 bins in total, uniformly ranging from $-180^\circ$ to $180^\circ$, with $5^\circ$ for each bin. 
        Red: the left part (bin 1 to 36, i.e., $-180^\circ$ to $0^\circ$) and blue shows the right part (bin 37 to 72, $0^\circ$ to $180^\circ$) of the orientation and cumulative histograms. 
        {\bf Note}: the {\it x}-axes of orientation histograms and cumulative histograms show the number of the histogram bin.
    }
    \label{fig:realImageStudies}
\end{figure}

\subsection{Effect of Rotations on Orientation Histogram}
\label{sec:effectOrientationHistogram}
\begin{figure}
    \centering
    \begin{subfigure}[b]{0.10\textwidth}
        \includegraphics[width=\textwidth]{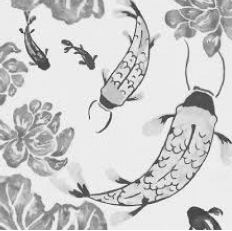}
        \caption{$0^\circ$}
     \end{subfigure}
     \begin{subfigure}[b]{0.115\textwidth}
        \includegraphics[width=\textwidth]{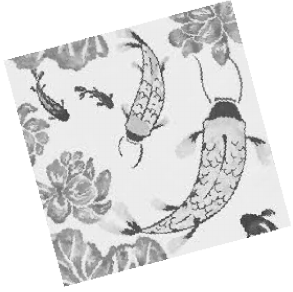}
        \caption{$15^\circ$}
     \end{subfigure}
     \begin{subfigure}[b]{0.125\textwidth}
        \includegraphics[width=\textwidth]{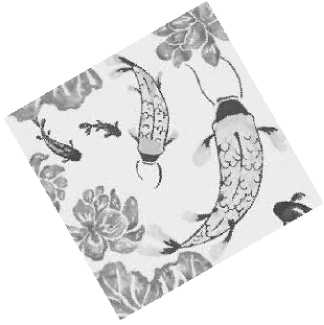}
        \caption{$30^\circ$}
     \end{subfigure}
     \begin{subfigure}[b]{0.13\textwidth}
        \includegraphics[width=\textwidth]{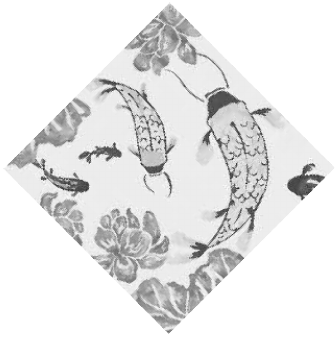}
        \caption{$45^\circ$}
     \end{subfigure} 
     \begin{subfigure}[b]{0.13\textwidth}
        \includegraphics[width=\textwidth]{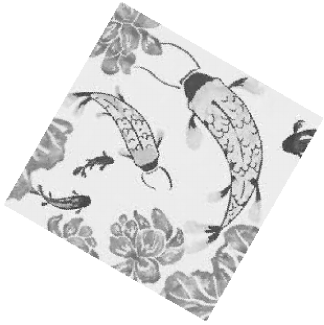}
        \caption{$60^\circ$}
     \end{subfigure} 
     \begin{subfigure}[b]{0.09\textwidth}
        \includegraphics[width=\textwidth]{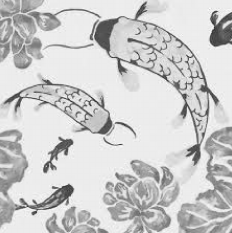}
        \caption{$90^\circ$}
     \end{subfigure} 
     \begin{subfigure}[b]{0.13\textwidth}
        \includegraphics[width=\textwidth]{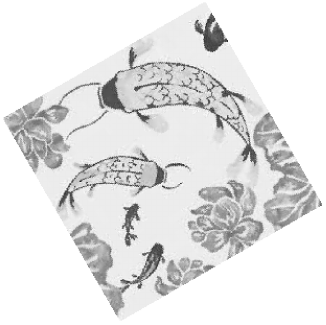}
        \caption{$120^\circ$}
     \end{subfigure} 
     \begin{subfigure}[b]{0.13\textwidth}
        \includegraphics[width=\textwidth]{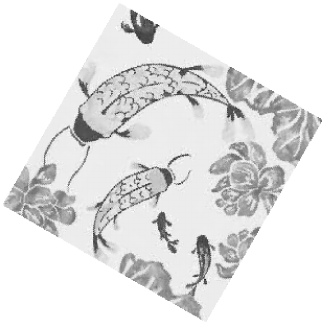}
        \caption{$150^\circ$}
     \end{subfigure} 
     \caption{
        Different versions of an example image under rotations: $0^\circ$, $15^\circ$, $30^\circ$, $45^\circ$, $60^\circ$, $90^\circ$, $120^\circ$ and $150^\circ$.
     }
     \label{fig:imageRotations}
\end{figure}

\begin{figure}[t]
    \centering
    \begin{subfigure}[b]{0.245\textwidth}
        \includegraphics[width=\textwidth]{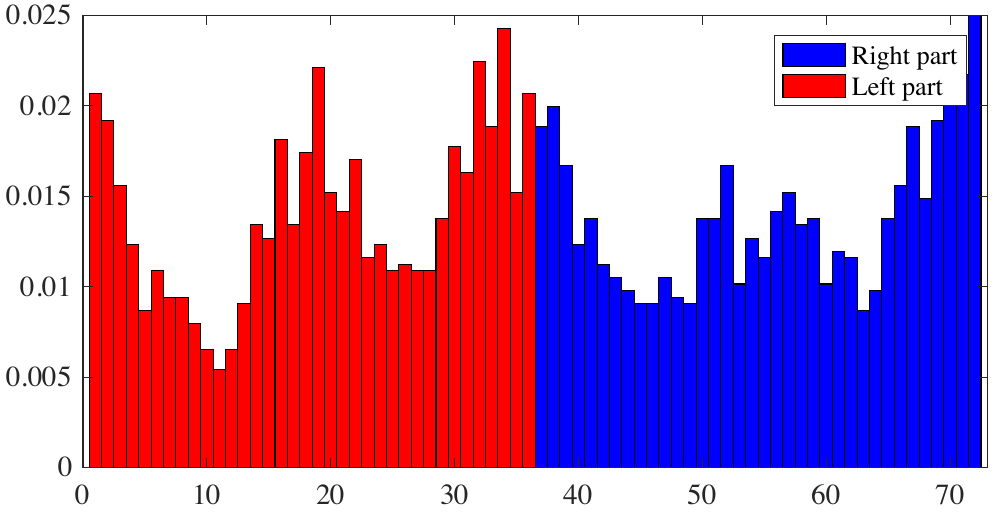}
        \caption{$0^\circ$}
    \end{subfigure}
    \begin{subfigure}[b]{0.245\textwidth}
        \includegraphics[width=\textwidth]{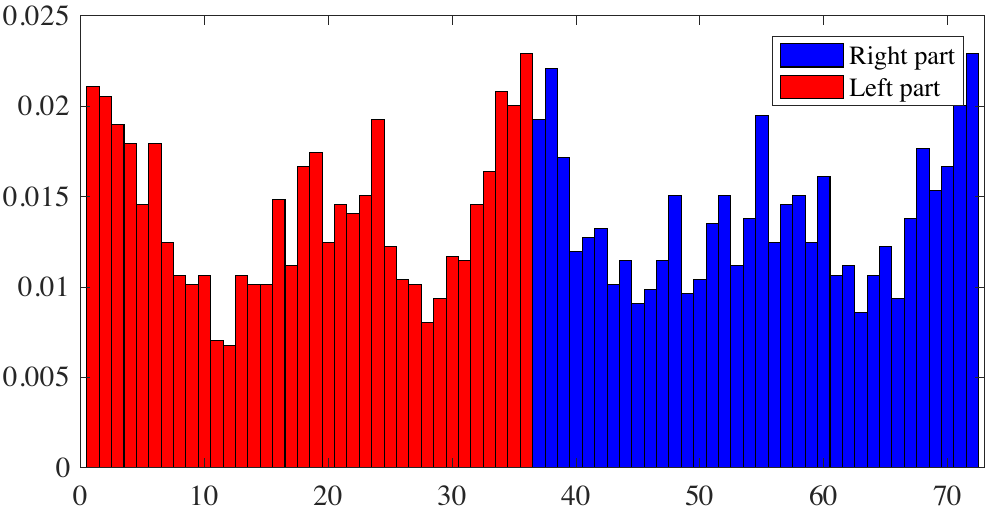}
        \caption{$15^\circ$}
    \end{subfigure}
    \begin{subfigure}[b]{0.245\textwidth}
        \includegraphics[width=\textwidth]{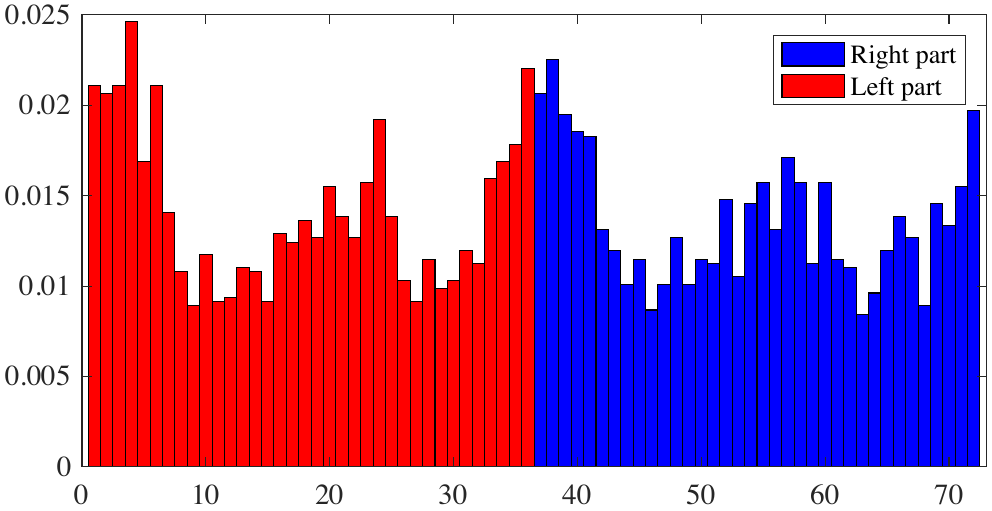}
        \caption{$30^\circ$}
    \end{subfigure}
    \begin{subfigure}[b]{0.245\textwidth}
        \includegraphics[width=\textwidth]{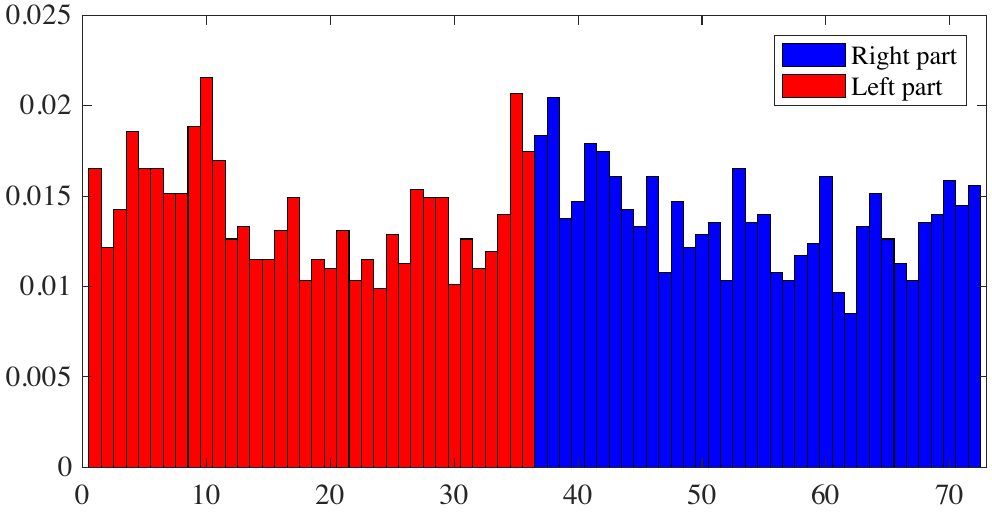}
        \caption{$45^\circ$}
    \end{subfigure} 
    \begin{subfigure}[b]{0.245\textwidth}
        \includegraphics[width=\textwidth]{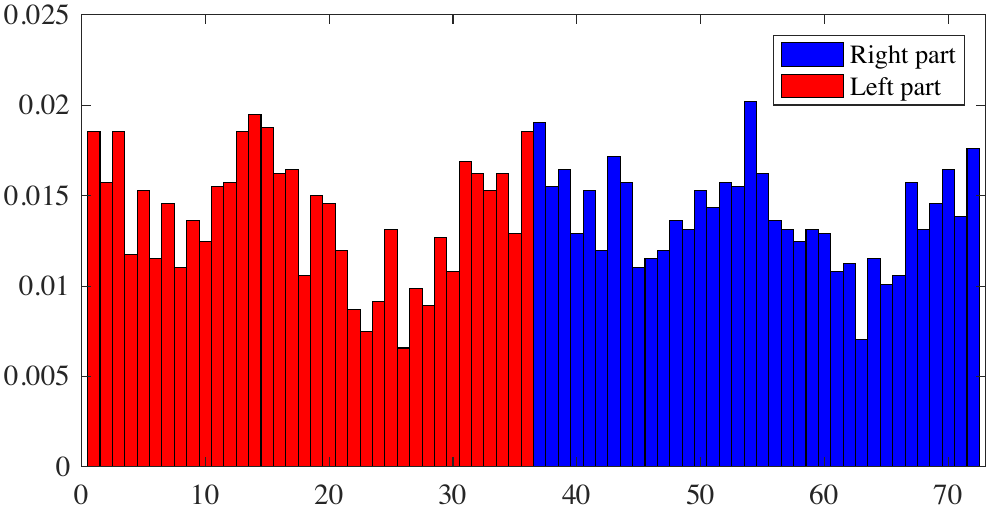}
        \caption{$60^\circ$}
    \end{subfigure} 
    \begin{subfigure}[b]{0.245\textwidth}
        \includegraphics[width=\textwidth]{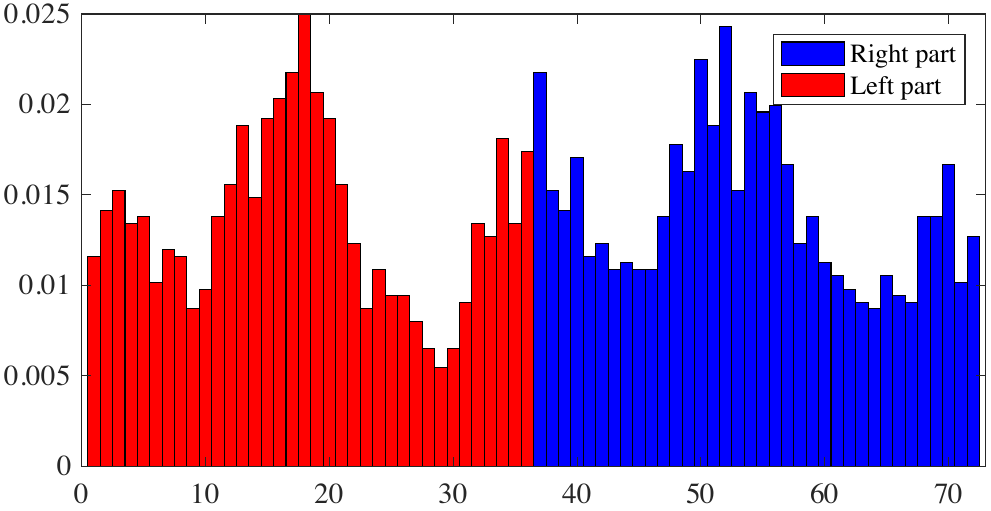}
        \caption{$90^\circ$}
    \end{subfigure} 
    \begin{subfigure}[b]{0.245\textwidth}
        \includegraphics[width=\textwidth]{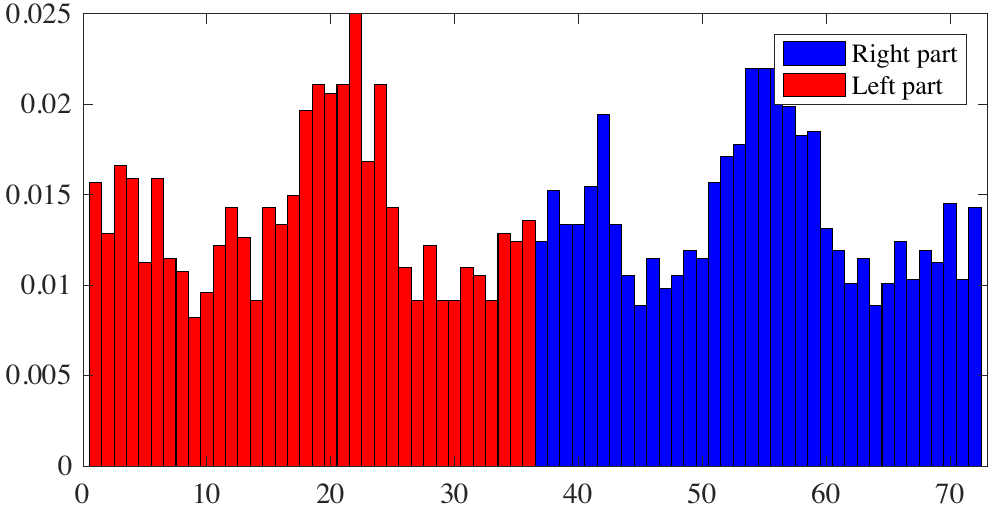}
        \caption{$120^\circ$}
     \end{subfigure} 
     \begin{subfigure}[b]{0.245\textwidth}
        \includegraphics[width=\textwidth]{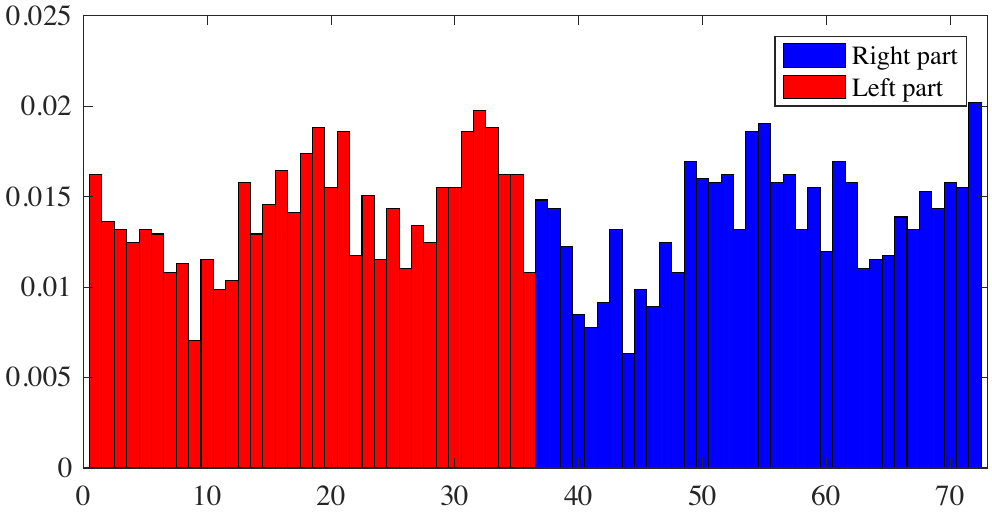}
        \caption{$150^\circ$}
     \end{subfigure} 
     \caption{
        Orientation histograms of the corresponding images in \figureautorefname \ref{fig:imageRotations}. 
        The levels of symmetry and smoothness of the orientation histogram of rotated images are preserved after rotations, showing that the SOH is invariant to rotation to some extent. 
        Their corresponding cumulative histogram curves are also presented in \figureautorefname\ref{fig:orientationRotations_cum}. 
        (Note: the x-axes shows the number of histogram bins; while the y-axes shows the bin value.)
     }
     \label{fig:orientationRotations}
\end{figure}

\begin{figure}[!t]
    \centering
    \begin{subfigure}[b]{0.245\textwidth}
        \includegraphics[width=\textwidth]{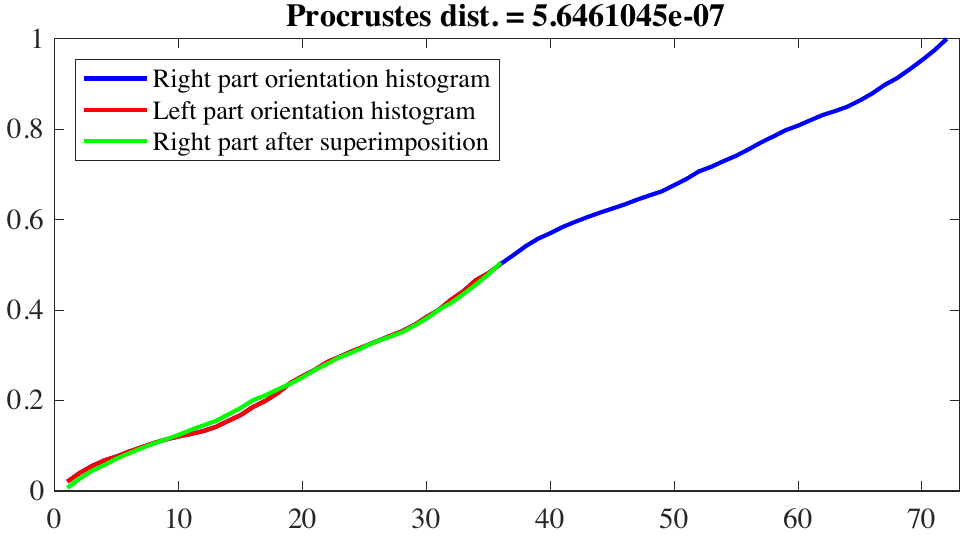}
        \caption{$0^\circ$}
    \end{subfigure}
    \begin{subfigure}[b]{0.245\textwidth}
        \includegraphics[width=\textwidth]{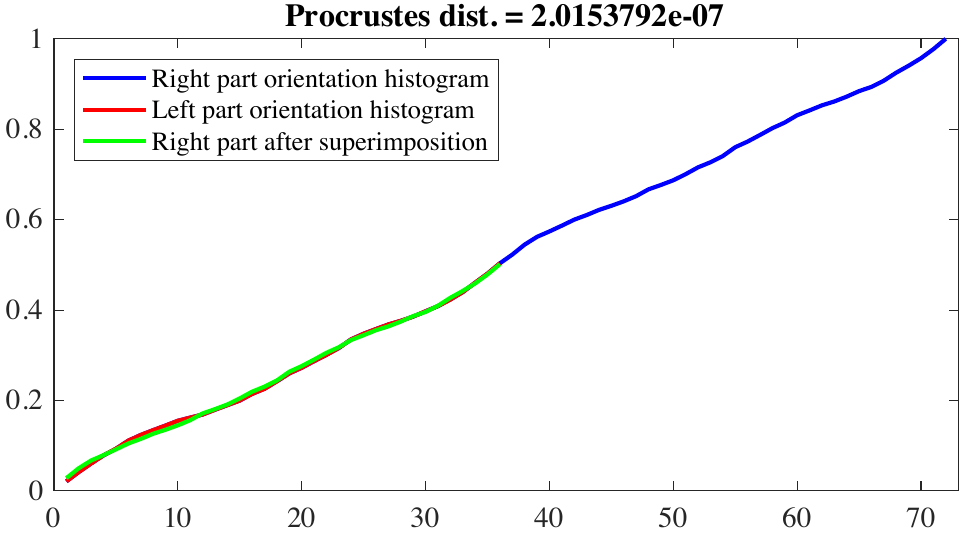}
        \caption{$15^\circ$}
    \end{subfigure}
    \begin{subfigure}[b]{0.245\textwidth}
        \includegraphics[width=\textwidth]{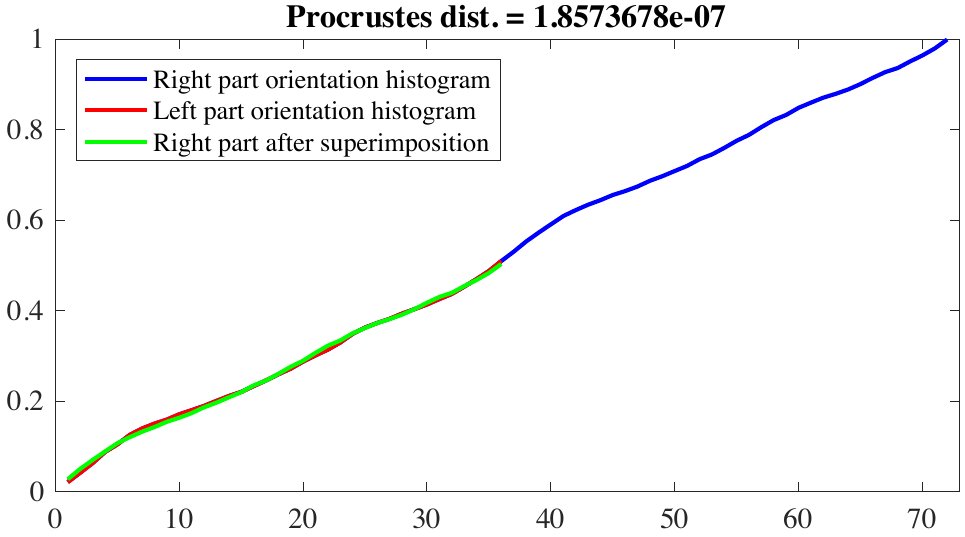}
        \caption{$30^\circ$}
    \end{subfigure}
    \begin{subfigure}[b]{0.245\textwidth}
        \includegraphics[width=\textwidth]{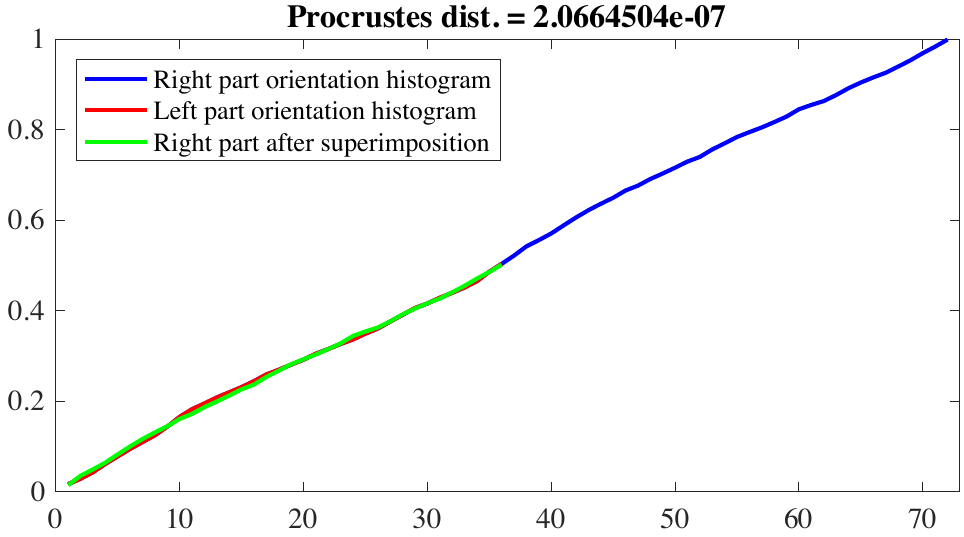}
        \caption{$45^\circ$}
    \end{subfigure} 
    \begin{subfigure}[b]{0.245\textwidth}
        \includegraphics[width=\textwidth]{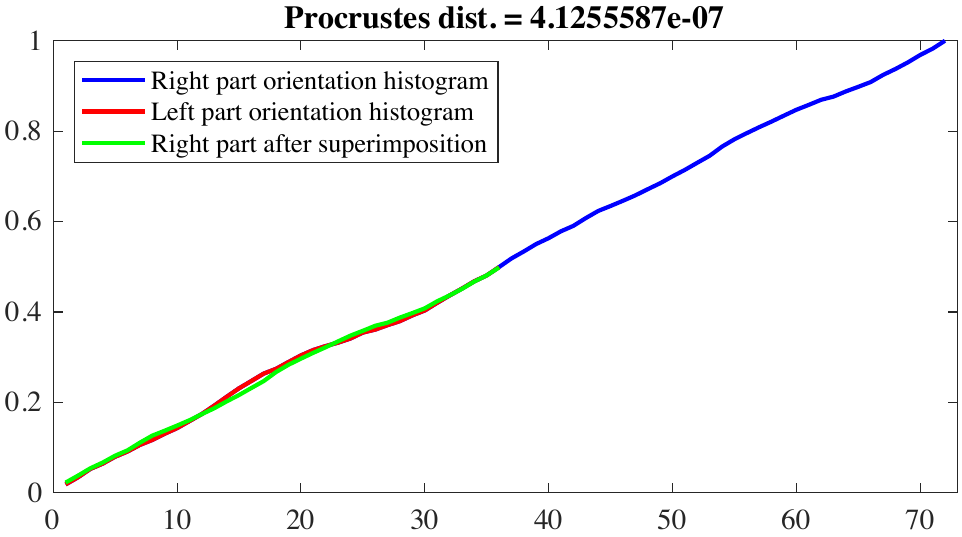}
        \caption{$60^\circ$}
    \end{subfigure} 
    \begin{subfigure}[b]{0.245\textwidth}
        \includegraphics[width=\textwidth]{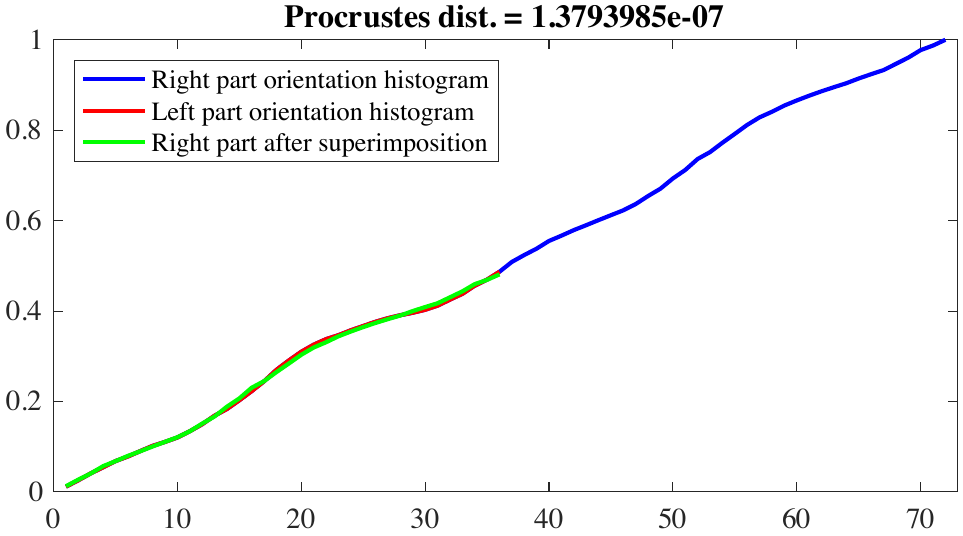}
        \caption{$90^\circ$}
    \end{subfigure} 
    \begin{subfigure}[b]{0.245\textwidth}
        \includegraphics[width=\textwidth]{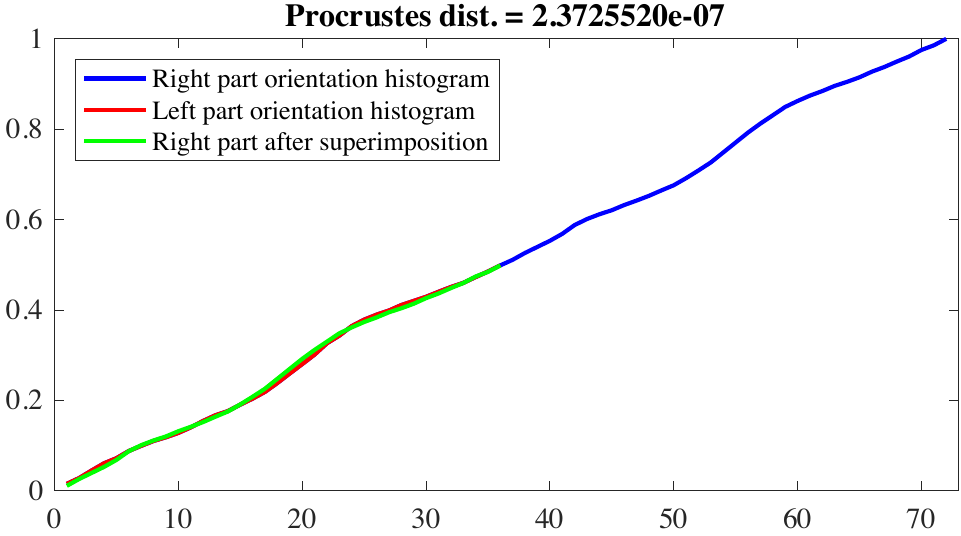}
        \caption{$120^\circ$}
    \end{subfigure} 
    \begin{subfigure}[b]{0.245\textwidth}
        \includegraphics[width=\textwidth]{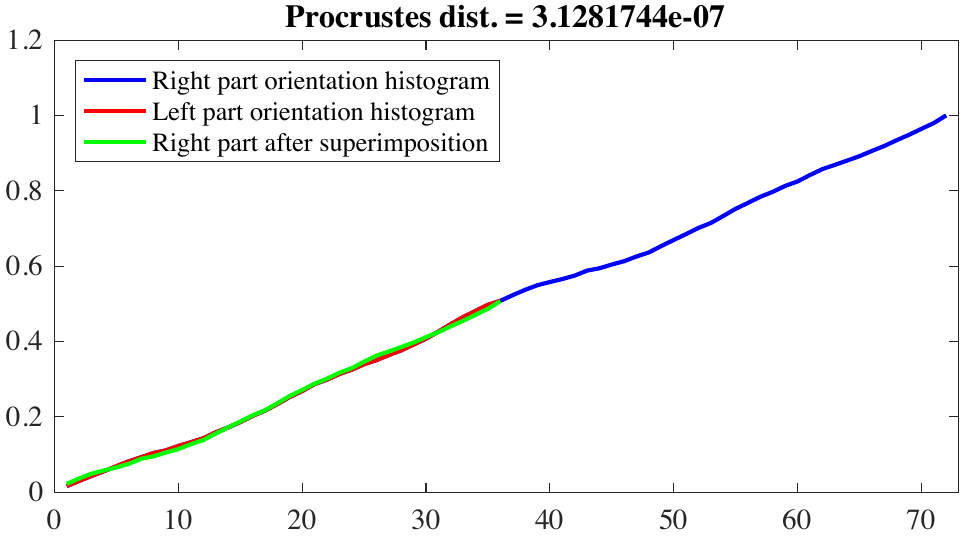}
        \caption{$150^\circ$}
    \end{subfigure} 
    \caption{
        Cumulative histogram curves of the corresponding images in \figureautorefname \ref{fig:imageRotations}. 
        Procrustes analysis is conducted on the left and right part of the curves, and the calculated Procrustes distance between them is shown in the figure title area. 
        The levels of symmetry and smoothness of the orientation histogram of rotated images are preserved after rotations, showing that the SOH is invariant to rotation to some extent. 
        As illustrated in \figuresautorefname\ref{fig:orientationRotations} and \ref{fig:orientationRotations_cum}, the cumulative histogram representation is more robust to rotations than that of orientation histogram, which is also easy to observe. 
        Note: the {\it x}-axes shows the number of histogram bins; while the {\it y}-axes indicates bin value.
    }
    \label{fig:orientationRotations_cum}
\end{figure}

Since the SOH feature vector is mainly based on symmetry and smoothness of the orientation histogram, it is invariant to rotations to some extent. 
To study this, a set of rotated versions (rotation angles: $0^\circ$, $15^\circ$, $30^\circ$, $45^\circ$, $60^\circ$, $90^\circ$, $120^\circ$, and $150^\circ$) of an image (\figureautorefname \ref{fig:imageRotations}) and their corresponding orientation histograms (\figureautorefname \ref{fig:orientationRotations}) and cumulative histogram curves (\figureautorefname \ref{fig:orientationRotations_cum}) are presented. 
The orientation histograms and their corresponding cumulative histograms contain 72 bins, with each bin $5^\circ$ spanning from $-180^\circ$ to $180^\circ$. 
Hence, the left part (red), which is the number of bins from the 1st to 36th, ranges from $-180^\circ$ to $0^\circ$; and the blue part (the 37th to the end, i.e., the 72th bin) ranges from $0^\circ$ to $180^\circ$. 
As shown in \figuresautorefname \ref{fig:orientationRotations} and \ref{fig:orientationRotations_cum}, the two properties of symmetry and smoothness of the orientation histograms of all rotated images are preserved. 
The blue parts of the orientation histograms are {\it translationally symmetrical} to the corresponding red parts (this is also reflected on their cumulative histogram curves), where the bin edge between the 36th and 37th bin (i.e., at $0^\circ$) is the {\it symmetry axis}. 
In addition, the smoothness of the cumulative histograms indicates the smoothness (smoothly varying, no sharp changes) of the corresponding orientation histograms. 
Therefore, the SOH is invariant to rotations.

\end{document}